\documentclass{article}
\pdfoutput=1
\usepackage{microtype}
\usepackage{graphicx}
\usepackage{booktabs} % for professional tables

\usepackage{hyperref}

\usepackage[accepted]{icml2021}

\icmltitlerunning{Robust Asymmetric Learning in POMDPs}

\usepackage{caption}
\usepackage{subcaption}
\usepackage{amsthm}
\usepackage{multirow}
\usepackage{amsmath}
\usepackage{amsfonts}
\usepackage{pgfplots}
\usepackage{lipsum}
\usepackage{natbib}
\usepackage{listings}
\usepackage{amssymb}
\usepackage{algorithm}
\usepackage{algorithmic,eqparbox,array}
\usepackage{balance}
\usepackage{chngpage}
\usepackage{setspace}
\usepackage{etoolbox}
\usepackage{tikz}
\usetikzlibrary{shapes.geometric}
\usetikzlibrary{calc}
\usetikzlibrary{shapes,arrows}

\pgfplotsset{compat=1.16}

\newtheorem{assumption}{Assumption}

\newtheorem{theorem}{Theorem}%[section]
\newtheorem{lemma}{Lemma}

\newtheorem{definition}{Definition}

\usepackage[colorinlistoftodos,prependcaption, textsize=tiny,textwidth=2cm]{todonotes}

\newcommand{\tends}{\rightarrow}
\DeclareMathOperator{\argmax}{arg\,max}
\DeclareMathOperator{\argmin}{arg\,min}
\newcommand{\where}{\quad \text{where }\ \ }

\newcounter{sthe}

\usepackage{enumitem}
\setlist[enumerate]{itemsep=0mm}

\usepackage{enumitem}
\setlist{leftmargin=5.5mm}
\usepackage{multicol}

\begin{document}

\twocolumn[
\icmltitle{Robust Asymmetric Learning in POMDPs}

\icmlsetsymbol{equal}{*}
\begin{icmlauthorlist}
\icmlauthor{Andrew Warrington}{equal,oxf}
\icmlauthor{J. Wilder Lavington}{equal,ubc,iai}
\icmlauthor{Adam \'Scibior}{ubc,iai}
\icmlauthor{Mark Schmidt}{ubc,ami}
\icmlauthor{Frank Wood}{ubc,iai,mil}
\end{icmlauthorlist}
\icmlaffiliation{oxf}{Department of Engineering Science, University of Oxford}
\icmlaffiliation{ubc}{Department of Computer Science, University of British Columbia}
\icmlaffiliation{iai}{Inverted AI}
\icmlaffiliation{mil}{Montr\'eal Institute for Learning Algorithms (MILA)}
\icmlaffiliation{ami}{Alberta Machine Learning Intelligence Institute (AMII)}
\icmlcorrespondingauthor{Andrew Warrington}{andreww@robots.ox.ac.uk}
\icmlkeywords{Machine Learning, ICML, Asymmetry, Reinforcement Learning, A2D, Imitation Learning}
\vskip 0.3in
]

\printAffiliationsAndNotice{\icmlEqualContribution} % otherwise use the standard text.

\renewcommand{\d}{\mathrm{d}}

\begin{abstract}
Policies for partially observed Markov decision processes can be efficiently learned by imitating expert policies learned using asymmetric information.  Unfortunately, existing approaches for this kind of imitation learning have a serious flaw: the expert does not know what the trainee cannot see, and may therefore encourage actions that are sub-optimal or unsafe under partial information.  To address this flaw, we derive an update that, when applied iteratively to an expert, maximizes the expected reward of the trainee's policy. Using this update, we construct a computationally efficient algorithm, adaptive asymmetric DAgger (A2D), that jointly trains the expert and trainee policies. We then show that A2D allows the trainee to safely imitate the modified expert, and outperforms policies learned either by imitating a fixed expert or direct reinforcement learning.
\end{abstract}

\section{Introduction}
\label{sec:intro}
Consider the stochastic shortest path problem~\citep{bertsekas1991analysis} where an agent learns to cross a frozen lake while avoiding patches of weak ice. The agent can either cross the ice directly, or take the longer, safer route circumnavigating the lake.  The agent is provided with aerial images of the lake, which include color variations at patches of weak ice.  To cross the lake, the agent must learn to identify its own position, goal position, and the location of weak ice from the images.  Even for this simple environment, high-dimensional inputs and sparse rewards can make learning a suitable policy computationally expensive and sample inefficient.  Therefore one might instead efficiently learn, in simulation, an omniscient \emph{expert}, conditioned on a low-dimensional vector which fully describes the state of the world, to complete the task. A \emph{trainee}, observing only images, can then learn to mimic the actions of the expert using sample-efficient online imitation learning~\citep{Ross2011}. This yields a high-performing trainee, conditioned on images, learned with fewer environment interactions overall compared to direct reinforcement learning (RL). 

While appealing, this approach can fail in environments where the expert has access to information unavailable to the agent, referred to as \emph{asymmetric information}.  Consider instead that the image of the lake does not indicate the location of the weak ice.  The trainee now operates under increased uncertainty.  This results in a different optimal partially observing policy, as the agent should now circumnavigate the lake.  However, imitating the expert forces the trainee to always cross the lake, despite being unable to locate and avoid the weak ice. Even though the expert is optimal under full information, the supervision provided to the trainee through imitation learning is poor and yields a policy that is not optimal under partial information. The key insight is that \emph{the expert has no knowledge of what the trainee does not know}.  Therefore, the expert cannot provide suitable supervision, and proposes actions that are not robust to the increased uncertainty under partial information. The main algorithmic contribution we present follows from this insight: the \emph{expert} must be refined based on the behavior of the \emph{trainee} imitating it.

\begin{figure*}
    \centering
    \includegraphics[width=\textwidth]{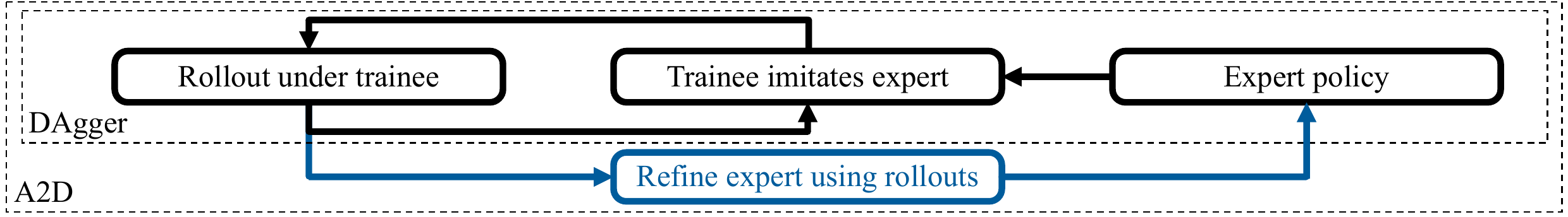}
    \vspace*{-0.3cm}
    \caption{Flow chart describing adaptive asymmetric DAgger (A2D), introduced in this work, which builds on DAgger~\citep{Ross2011} by further refining the expert conditioned on the trainee's policy. }
    \label{fig:a2d}
\end{figure*}

Building on this insight, we present a new algorithm: adaptive asymmetric DAgger (A2D), illustrated in Figure \ref{fig:a2d}.  A2D extends imitation learning by refining the expert policy, such that the resulting supervision moves the trainee policy closer to the optimal \emph{partially observed} policy.  This allows us to safely take advantage of asymmetric information in imitation learning.  Crucially, A2D can be easily integrated with a variety of different RL algorithms, does not require any pretrained artifacts, policies or example trajectories, and does not take computationally expensive and high-variance RL steps in the trainee policy network.  

We first introduce asymmetric imitation learning (AIL).  AIL uses an expert, conditioned on full state information, to supervise learning a trainee, conditioned on partial information.  We show that the solution to the AIL objective is a posterior inference over the true state; and provide sufficient conditions for when the expert is guaranteed to provide correct supervision.  Using these insights, we then derive the theoretical A2D update to the expert policy parameters in terms of Q functions.  This update maximizes the reward of the trainee implicitly defined through AIL.  We then modify this update to use Monte Carlo rollouts and GAE~\citep{schulman2015high} in place of Q functions, thereby reducing the dependence on function approximators. 

We apply A2D to two pedagogical gridworld environments, and an autonomous vehicle scenario, where AIL fails. We show A2D recovers the optimal partially observed policy with fewer samples, lower computational cost, and less variance compared to similar methods.  These experiments demonstrate the efficacy of A2D, which makes learning via imitation and reinforcement safer and more efficient, even in difficult high dimensional control problems such as autonomous driving. Code and additional materials are available at \url{https://github.com/plai-group/a2d}.

\section{Background}
\label{sec:background}

\subsection{Optimality \& MDPs}\label{mdp_background}
An MDP, $\mathcal{M}_{\Theta}(\mathcal{S},\mathcal{A},\mathcal{R},\mathcal{T}_0,\mathcal{T},\Pi_{\Theta})$, is defined as a random process which produces a sequence $\tau_t := \{a_t,s_t,s_{t+1},r_t\}$, for a set of states $s_t \in \mathcal{S}$, actions $a_t \in \mathcal{A}$, initial state $p(s_0) \in \mathcal{T}_0$, transition dynamics $ p(s_{t+1}|s_t,a_t) \in \mathcal{T}$, reward function $r_t : \mathcal{S} \times \mathcal{A} \times \mathcal{S} \rightarrow  \mathbb{R}$, and policy $\pi_{\theta} \in \Pi_{\Theta} : \mathcal{S} \rightarrow \mathcal{A}$ parameterized by $\theta \in \Theta$. The generative model, shown in Figure \ref{fig:dpgm}, for a finite horizon process is defined as:
\begin{align}
    q_{\pi_{\theta}}(\tau) & = p(s_0)\prod\nolimits_{t=0}^{T}p(s_{t+1}|s_t,a_t)\pi_{\theta}(a_t|s_t).\label{equ:background:mdp}
\end{align}
We denote the marginal distribution over state $s_t \in \mathcal{S}$ at time $t$ as  $q_{\pi_{\theta}}(s_t)$.  The objective of RL is to recover the policy which maximizes the expected cumulative reward over a trajectory, $\theta^* = \argmax_{\theta \in \Theta} \mathbb{E}_{q_{\pi_{\theta}}} [\sum_{t=0}^T r_t(s_t,a_t,s_{t+1})]$. We consider an extension of this, instead maximizing the non-stationary, infinite horizon discounted return:
\allowdisplaybreaks
\begin{align}
    & \theta^* = \argmax_{\theta \in \Theta} \ \mathbb{E}_{d^{\pi_{\theta}}(s) \pi_{\theta}(a|s)} [Q^{\pi_\theta}(a,s)], \label{equ:MDP:RL} \\
    & \mathrm{where} \ \ d^{\pi_{\theta}}(s) = (1-\gamma) \sum\nolimits_{t=0}^{\infty} \gamma^t q_{\pi_{\theta}}(s_t=s), \label{eq:background:occ} \\
    & Q^{\pi_{\theta}} (a, s) = \!\! \mathop{\mathbb{E}}_{p(s'|s,a)} \!\!
    \big[ r(s,a,s') + \!\!\! \mathop{\gamma \ \mathbb{E}}_{\pi_{\theta}(a'|s')} \!\! [ Q^{\pi_{\theta}} (a', s') ] \big],
\end{align}
where $d^{\pi_{\theta}}(s)$ is referred to as the \emph{state occupancy}~\cite{pmlr-v125-agarwal20a}, and the \emph{Q function},  $Q^{\pi}$, defines the expected discounted sum of rewards ahead given a state-action pair. 

\subsection{State Estimation and POMDPs}\label{pomdp_background}
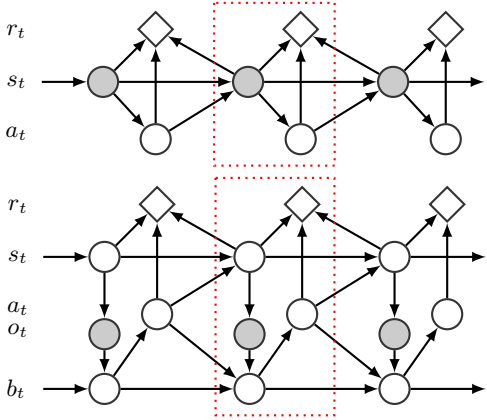
\begin{figure}
    \centering 
    \tikzstyle{main}=[circle, minimum size = 4mm, thick, draw =black!80, node distance = 6mm, fill = white!100]
\tikzstyle{obs}=[circle, minimum size = 4mm, thick, draw =black!80, node distance = 6mm, fill = black!20]
\tikzstyle{text_only}=[circle, minimum size = 4mm, thick, node distance = 6mm]
\tikzstyle{connect}=[-latex, thick]
\tikzstyle{main_sq}=[style={diamond}, minimum size = 2mm, thick, draw =black!80, node distance = 6mm]
\tikzstyle{every node}=[font=\small]%, scale=0.8]

\begin{subfigure}[t]{0.48\textwidth}
\centering
\begin{tikzpicture}[baseline={([yshift={-\ht\strutbox}]current bounding box.north)}]  % ,outer sep=0pt,inner sep=0pt]

  \node[obs] (s1) [] {};
  \node[obs, node distance = 15mm] (s2) [right = of s1] {};
  \node[obs, node distance = 15mm] (s3) [right = of s2] {};
  
  \node[main_sq] (r1) [above right = of s1] {};
  \node[main_sq] (r2) [above right = of s2] {};
  \node[main_sq] (r3) [above right = of s3] {};
  
  \node[main, node distance = 10mm] (a1) [below = of r1] {};
  \node[main, node distance = 10mm] (a2) [below = of r2] {};
  \node[main, node distance = 10mm] (a3) [below = of r3] {}; 
  
  \node[text_only] (st) [left = of s1] {$s_t$};
  \node[text_only, node distance = 0mm ] (rt) [above = of st] {$r_t$};
  \node[text_only, node distance = 0mm] (at) [below = of st] {$a_t$};
  
  \node[text_only, node distance = 10mm] (sf) [right = of s3] {};

  \path 				(s1) edge [connect] (r1) % dots.
                        (s2) edge [connect] (r2) % dots.
                        (s3) edge [connect] (r3) % dots.
                        
                        (s1) edge [connect] (s2) % dots.
                        (s2) edge [connect] (s3) % dots.
                        
                        (a1) edge [connect] (s2) % dots.
                        (a2) edge [connect] (s3) % dots.
                        
                        (s2) edge [connect] (r1) % dots.
                        (s3) edge [connect] (r2) % dots.
                        
                        (s1) edge [connect] (a1) % dots.
                        (s2) edge [connect] (a2) % dots.
                        (s3) edge [connect] (a3) % dots.
                        
                        (a1) edge [connect] (r1) % dots.
                        (a2) edge [connect] (r2) % dots.
                        (a3) edge [connect] (r3) % dots.
                        
                        (st) edge [connect] (s1) % dots.
                        (s3) edge [connect] (sf) % dots.
    ;
    
    \draw[red,thick,dotted] ($(s2.north west)+(-0.3,0.9)$)  rectangle ($(a2.south east)+(0.3,-0.2)$);
    
\end{tikzpicture}

\end{subfigure}
\hfill
\begin{subfigure}[t]{0.48\textwidth}
\vspace{0.1cm}
\centering
\begin{tikzpicture}[baseline={([yshift={-\ht\strutbox}]current bounding box.north)}]  % ,outer sep=0pt,inner sep=0pt]

  \node[main] (s1) [] {};
  \node[main, node distance = 15mm] (s2) [right = of s1] {};
  \node[main, node distance = 15mm] (s3) [right = of s2] {};
  
  \node[main_sq] (r1) [above right = of s1] {};
  \node[main_sq] (r2) [above right = of s2] {};
  \node[main_sq] (r3) [above right = of s3] {};
  
  \node[obs] (o1) [below = of s1] {};
  \node[obs] (o2) [below = of s2] {};
  \node[obs] (o3) [below = of s3] {};
   
  \node[main, node distance = 3mm] (b1) [below = of o1] {};
  \node[main, node distance = 3mm] (b2) [below = of o2] {};
  \node[main, node distance = 3mm] (b3) [below = of o3] {};
   
  \node[main, node distance = 10mm] (a1) [below = of r1] {};
  \node[main, node distance = 10mm] (a2) [below = of r2] {};
  \node[main, node distance = 10mm] (a3) [below = of r3] {}; 
  
  \node[text_only] (st) [left = of s1] {$s_t$};
  \node[text_only, node distance = 0mm ] (rt) [above = of st] {$r_t$};
  \node[text_only, node distance = 0mm] (at) [below = of st] {$a_t$};
  \node[text_only, node distance = 3mm] (ot) [below = of st] {$o_t$};
  \node[text_only] (bt) [left = of b1] {$b_t$};
  
  \node[text_only, node distance = 10mm] (sf) [right = of s3] {};
  \node[text_only, node distance = 10mm] (bf) [right = of b3] {};

  \path 				(s1) edge [connect] (r1) % dots.
                        (s2) edge [connect] (r2) % dots.
                        (s3) edge [connect] (r3) % dots.
                        
                        (s1) edge [connect] (s2) % dots.
                        (s2) edge [connect] (s3) % dots.
                        
                        (b1) edge [connect] (b2) % dots.
                        (b2) edge [connect] (b3) % dots.
                        
                        (a1) edge [connect] (s2) % dots.
                        (a2) edge [connect] (s3) % dots.
                        
                        (s2) edge [connect] (r1) % dots.
                        (s3) edge [connect] (r2) % dots.
                        
                        (s1) edge [connect] (o1) % dots.
                        (s2) edge [connect] (o2) % dots.
                        (s3) edge [connect] (o3) % dots.
                        
                        (o1) edge [connect] (b1) % dots.
                        (o2) edge [connect] (b2) % dots.
                        (o3) edge [connect] (b3) % dots.
                        
                        (b1) edge [connect] (a1) % dots.
                        (b2) edge [connect] (a2) % dots.
                        (b3) edge [connect] (a3) % dots.
                        
                        (a1) edge [connect] (r1) % dots.
                        (a2) edge [connect] (r2) % dots.
                        (a3) edge [connect] (r3) % dots.
                        
                        (st) edge [connect] (s1) % dots.
                        (s3) edge [connect] (sf) % dots.
                        
                        (bt) edge [connect] (b1) % dots.
                        (b3) edge [connect] (bf) % dots.
                        
                        (a1) edge [connect] (b2) % dots.
                        (a2) edge [connect] (b3) % dots.
    ;
    
    \draw[red,thick,dotted] ($(s2.north west)+(-0.3,0.9)$)  rectangle ($(b2.south west)+(1.28,-0.2)$);
    
\end{tikzpicture}

\end{subfigure}
    \caption{Graphical models of an MDP (top) and a POMDP (bottom) with identical initial and state transition dynamics, $p(s_t | s_{t-1}, a_{t})$, $p(s_0)$, and reward function $R(s_t, a_t,s_{t+1})$.}
    \label{fig:dpgm}
\end{figure}

A POMDP extends an MDP by observing a random variable $o_t \in \mathcal{O}$, dependent on the state, $o_t \sim p(\cdot | s_t)$, instead of the state itself. The policy then samples actions conditioned on all previous observations and actions: $\pi_{\phi} ( a_t | a_{0:t-1}, o_{0:t})$.  In practice, a \emph{belief state}, $b_t \in \mathcal{B}$, is constructed from $(a_{0:t-1}, o_{0:t})$, as an estimate of the underlying state.  The policy, $\pi_{\phi} \in \Pi_{\Phi}\!:\! \mathcal{B} \rightarrow \mathcal{A}$, is then conditioned on this belief state~\citep{doshi2013bayesian, igl2018dvrl, kaelbling1998planning}. The resulting stochastic process, denoted $\mathcal{M}_{\Phi}(\mathcal{S},\mathcal{O},\mathcal{B},\mathcal{A},\mathcal{R},\mathcal{T}_0,\mathcal{T},\Pi_{\Phi})$, generates a sequence of tuples $\tau_t\!=\!\{ a_t,b_t,o_t,s_t,s_{t+1},r_t\}$. As before, we wish to find a policy, $\pi_{\phi^*}\in\Pi_{\Phi}$, which maximizes the expected cumulative reward under the generative model: 
\begin{align}
    \begin{aligned}
       q_{\pi_{\phi}}(\tau) &= p(s_0) \prod\nolimits_{t=0}^{T} p(s_{t+1}|s_t,a_t) \times \\
     & \quad \quad p(b_t| b_{t-1}, o_{t}, a_{t-1}) p(o_t|s_t) \pi_{\phi}(a_t|b_t).\label{equ:background:pomdp_dist}
    \end{aligned}
\end{align}
It is common to instead condition the policy on the last $w$ observations and $w-1$ actions~\citep{laskin2020reinforcement, murphy2000survey}, i.e. $b_t := \left( a_{t-w:t-1}, o_{t-w:t} \right)$, rather than using the potentially infinite dimensional random variable~\citep{murphy2000survey}, defined recursively in Figure \ref{fig:dpgm}.  This ``windowed'' belief state representation is used throughout this paper.  

We also note that $q_{\pi}$ is used to denote the distribution over trajectories under the subscripted policy (\eqref{equ:background:mdp} and \eqref{equ:background:pomdp_dist} for $\pi_{\theta}(\cdot|s_t)$ and $\pi_{\phi}(\cdot|b_t)$ respectively).  The occupancies $d^{\pi_{\phi}}(s)$ and $d^{\pi_{\phi}}(b)$ define marginals of $d^{\pi_{\phi}}(s, b)$ in a partially observed processes (as in \eqref{eq:background:occ}).  Later we discuss \emph{MDP-POMDP pairs}, defined as an MDP and a POMDP with identical state transition dynamics, reward generating functions and initial state distributions. However, these process pairs can, and often do, have different optimal policies. This discrepancy is the central issue addressed in this work.

\subsection{Imitation Learning}
Imitation learning (IL) assumes access to either an expert policy capable of solving a task, or example trajectories generated by such an expert.  Given example trajectories, the \emph{trainee} is learned by regressing onto the actions of the expert.  However, this approach can perform arbitrarily poorly for states not in the training set~\citep{laskey2017dart}.  Alternatively, online IL (OIL) algorithms, such as DAgger~\citep{Ross2011}, assume access to an expert that can be queried at any state.  DAgger rolls out under a mixture of the expert $\pi_{\theta}$ and trainee $\pi_{\phi}$ policies, denoted $\pi_{\beta}$.  The trainee is then updated to replicate the experts' actions at the visited states:
\begin{align}
    & \phi^* = \argmin_{\phi \in \Phi} \mathbb{E}_{d^{\pi_{\beta}}(s)} \left[ \mathbb{KL} \left[ \pi_{\theta}(a|s) || \pi_{\phi}(a|s) \right] \right],  \\
    & \mathrm{where}\ \ \pi_{\beta}(a | s) = \beta \pi_{\theta}(a | s) + (1-\beta) \pi_{\phi}(a | s). \label{equ:background:dagger_mixture}
\end{align}
The coefficient $\beta$ is annealed to zero during training.  This provides supervision in states visited by the trainee, thereby avoiding compounding out of distribution error which grows with time horizon~\cite{Ross2011,pmlr-v70-sun17d}.  While IL provides higher sample efficiency than RL, it requires an expert or expert trajectories, and is thus not always applicable.  A trainee learned using IL from an imperfect expert can perform arbitrarily poorly~\cite{pmlr-v70-sun17d}, even in OIL.  Addition of asymmetry in OIL can cause similar failures.

\subsection{Asymmetric Information}
In many simulated environments, additional information is available during training that is not available at test time.  This additional \emph{asymmetric information} can often be exploited to accelerate learning~\citep{choudhury2018data, pinto2017asymmetric, vapnik2009new}. For example, \citet{pinto2017asymmetric} exploit asymmetry to learn a policy conditioned on noisy image-based observations which are available at test time, but where the value function (or \emph{critic}), is conditioned on a compact and noiseless state representation, only available during training.  The objective function for this \emph{asymmetric actor critic}~\citep{pinto2017asymmetric} algorithm is:
\begin{align}
    &J(\phi) = \mathbb{E}_{d^{\pi_{\phi}}(s,b)}\left[ \mathbb{E}_{\pi_{\phi}(a | b)}  \left[ A^{\pi_{\phi}}(s , a) \right] \right],  \\
    & Q^{\pi_{\phi}} (a, s) = \mathbb{E}_{p(s'|s,a)} \left[ r(s,a,s') + \gamma V^{\pi_{\phi}}(s') \right], \\ 
    &V^{\pi_{\phi}}(s) = \mathbb{E}_{\pi_{\phi}(a | b)} \left[ Q^{\pi_{\phi}} (a, s) \right], \label{equ:background:qv_b}
\end{align}
where the \emph{asymmetric advantage} is defined as $A^{\pi_{\phi}}(s , a)$ = $Q^{\pi_{\phi}}(a,s) - V^{\pi_{\phi}}(s)$, and $V^{\pi_{\phi}}(s)$ is the \emph{asymmetric value function}.  Asymmetric methods often outperform ``symmetric'' RL as $Q^{\pi_{\phi}}(a,s)$ and $V^{\pi_{\phi}}(s)$ are simpler to tune, train, and provide lower-variance gradient estimates. 

Asymmetric information has also been used in a variety of other scenarios, including policy ensembles~\citep{sasaki2021behavioral, Song2019}, imitating attention-based representations~\citep{salter2019attentionprivileged}, multi-objective RL~\citep{Schwab2019}, direct state reconstruction~\citep{nguyen2020belief}, or privileged information dropout~\citep{pierrealex2020privileged, lambert2018deep}.  Failures induced by asymmetric information have also been discussed.  \citet{arora2018hindsight} identify an environment where a particular method fails.  \citet{choudhury2018data} use asymmetric information to improve policy optimization in model predictive control, but do not solve scenarios such as ``the trapped robot problem,'' referred to later as Tiger Door~\citep{littman1995pomdp}, and solved below. Notably, asymmetric environments are naturally suited to OIL~(AIL)~\citep{pinto2017asymmetric}:
\begin{align}
     &\phi^* = \argmin_{\phi}\ \mathbb{E}_{d^{\pi_{\beta}}(s,b)}  \left[ \mathbb{KL} \left[ \pi_{\theta}(a|s) || \pi_{\phi}(a|b) \right] \right], \\ 
     &\mathrm{where}\ \ \pi_{\beta}(a | s, b) = \beta \pi_{\theta}(a | s) + (1-\beta) \pi_{\phi}(a | b). \label{equ:background:asym_dagger_mixture} %\\%
\end{align}
As the expert is not used at test time, AIL can take advantage of asymmetry to simplify learning~\citep{pinto2017asymmetric} or enable data augmentation~\citep{Chen2019}. However, naive application of AIL can yield trainees that perform arbitrarily poorly. Further work has addressed learning from imperfect experts~\citep{ross2014reinforcement, pmlr-v70-sun17d, meng2019conditional}, but does not consider issues arising from the use of asymmetric information.  We demonstrate, analyze, and then address both of these issues in the following sections.

\section{AIL as Posterior Inference}
\label{sec:prelim}

We begin by analyzing the AIL objective in \eqref{equ:background:asym_dagger_mixture}.  We first show that the optimal trainee defined by this objective can be expressed as posterior inference over state conditioned on the expert policy.  This posterior inference is defined as:
\begin{definition}[Implicit policy]
\label{def:implicit_policy}
For any state-conditional policy $\pi_{\theta} \in \Pi_{\Theta}$ and any belief-conditional policy $\pi_{\eta} \in \Pi_{\Phi}$ we define $\hat{\pi}^\eta_{\theta} \in \hat{\Pi}_{\Theta}$ as the implicit policy of $\pi_{\theta}$ under $\pi_{\eta}$ as:
\begin{equation}
    {\hat{\pi}}_{\theta}^{\eta}(a|b) := \mathbb{E}_{d^{\pi_{\eta}}(s | b)}  \left[ \pi_{\theta}(a|s) \right] , \label{equ:pi_hat} 
\end{equation}
When $\pi_{\eta} = {\hat{\pi}}_{\theta}^{\eta}$, we refer to this policy as the implicit policy of $\pi_{\theta}$, denoted as just ${\hat{\pi}}_{\theta}$.
\end{definition}
Note that a policy, or policy set, with a hat (e.g. $\hat{\pi}_{\theta}$), indicates that the policy or set is implicitly defined through composition of the original policy (e.g. $\pi_{\theta}$) and the expectation defined in \eqref{equ:pi_hat}.  The implicit policy defines a posterior predictive density, marginalizing over the uncertainty over state.  We can then show that the solution to the AIL objective in \eqref{equ:background:asym_dagger_mixture} (for $\beta = 0$) is equivalent to the implicit policy:
\begin{theorem}[Asymmetric IL target]
\label{def:ail}
For any fully observing policy $\pi_{\theta}$ and fixed policy $\pi_{\eta}$, and assuming $\hat{\Pi}_{\Theta} \subseteq \Pi_{\Phi}$, then the implicit policy $\hat{\pi}_{\theta}^{\eta}$, defined in Definition \ref{def:implicit_policy}, minimizes the AIL objective:
\begin{align}
      &\hat{\pi}_{\theta}^{\eta} =
      \mathop{\argmin}_{\pi \in \Pi_{\Phi}} \mathbb{E}_{d^{\pi_{\eta}}(s,b)}  \left[ \mathbb{KL} \left[ \pi_{\theta}(a|s) || \pi(a|b) \right] \right]. \label{equ:prelim:ail_target}
\end{align}
\end{theorem}
\begin{proof}\renewcommand{\qedsymbol}{}
An extended proof is included in Appendix \ref{supp:thoery}.
\begin{align*}
    &\mathbb{E}_{d^{\pi_{\eta}}(s,b)}  \left[ \mathbb{KL} \left[ \pi_{\theta}(a|s) || \pi(a|b) \right] \right] \\
    &= -\mathbb{E}_{d^{\pi_{\eta}}(b)}  \left[ \mathbb{E}_{d^{\pi_{\eta}}(s)}  \left[ \mathbb{E}_{\pi_{\theta}(a|s)} \left[ \log \pi(a|b) \right] \right] \right] + K \\
    &= -\mathbb{E}_{d^{\pi_{\eta}}(b)}  \left[ \mathbb{E}_{\hat{\pi}_{\theta}^{\eta}(a|b)} \left[ \log \pi(a|b) \right] \right] + K \\
    &=
    \mathbb{E}_{d^{\pi_{\eta}}(b)}  \left[ \mathbb{KL} \left[ \hat{\pi}_{\theta}^{\eta}(a|b) || \pi(a|b) \right] \right] + K' 
\end{align*}
Since $\hat{\pi}_{\theta}^{\eta} \in \Pi_{\Phi}$, it follows that
\begin{align}
\hat{\pi}_{\theta}^{\eta} &=
\mathop{\argmin}_{\pi \in \Pi_{\Phi}}\ \mathop{\mathbb{E}}_{d^{\pi_{\eta}}(b)}  \left[ \mathbb{KL} \left[ \hat{\pi}_{\theta}^{\eta}(a | b) || \pi (a | b) \right] \right] \label{eq:variational} \\
&=
\mathop{\argmin}_{\pi \in \Pi_{\Phi}}\ \mathop{\mathbb{E}}_{d^{\pi_{\eta}}(s,b)}  \left[ \mathbb{KL} \left[ \pi_{\theta}(a|s) || \pi(a|b) \right] \right]. \quad \square 
\end{align}
\end{proof}
Theorem \ref{def:ail} shows that the implicit policy compactly defines the solution to the AIL objective.  This allows us to specify the dependence of the learned trainee through AIL on the expert policy.  We will in turn leverage this solution to derive the update applied to the expert parameters.  We note that this definition and theorem are closely related to a result also derived by \citet{weihs2020bridging}.  

However, drawing multiple state samples from a single conditional occupancy, $d^{\pi_{\eta}}(s \mid b)$, is not generally tractable without access to a model of $\mathcal{T}$ and $\mathcal{T}_0$. This is because sampling from $d^{\pi_{\eta}}(s \mid b)$ requires resampling multiple trajectories that include the specified belief state $b$, which cannot be done through direct environment interaction. Therefore, generating the samples required to integrate \eqref{equ:pi_hat} is not generally tractable.  We are, however, able to draw samples from the joint occupancy, $d^{\pi_{\eta}}(s,b)$, simply by rolling out under $\pi_{\eta}$.  Therefore, in practice, AIL instead learns a variational approximation to the implicit policy, $\pi_{\psi} \in \Pi_{\Psi}:\mathcal{B} \rightarrow \mathcal{A}$, by minimizing the following objective:
\begin{align}
    F(\psi) &= \mathop{\mathbb{E}}_{d^{\pi_{\eta}}(s,b)}  \left[ \mathbb{KL} \left[ \pi_{\theta}(a|s) || \pi_{\psi}(a|b) \right] \right], \label{equ:def:variational:obj} \\%
    \hspace{-0.2cm}\nabla_{\psi} F (\psi) &= \!  \mathop{\mathbb{-E}}_{d^{\pi_{\eta}}(s, b)} \bigg[ \mathop{\mathbb{E}}_{\pi_{\theta}(a | s)} \! \left[ \nabla_{\psi} \log \pi_{\psi} (a | b) \right] \bigg]. \label{equ:def:variational:gradient}
\end{align}
Crucially, this approach only requires samples from the \emph{joint} occupancy.  This avoids sampling from the \emph{conditional} occupancy, as required to directly solve \eqref{equ:pi_hat}.  If the variational family is sufficiently expressive, there exists a $\pi_{\psi}\in\Pi_{\Psi}$ for which the divergence between the implicit policy and variational approximation is zero.  In OIL, it is common to sample under the trainee policy by setting $\pi_{\eta} = \pi_{\psi}$, thereby defining a fixed point equation.  Under sufficient expressivity and exact updates, an iteration solving this fixed point equation converges to the implicit policy (see Appendix \ref{supp:thoery}).  In practice, this iterative scheme converges even in the presence of inexact updates and restricted policy classes. 

\begin{figure}[t!]
    \centering
    \begin{subfigure}[t]{0.23\textwidth}  
        \includegraphics[width=0.49\textwidth]{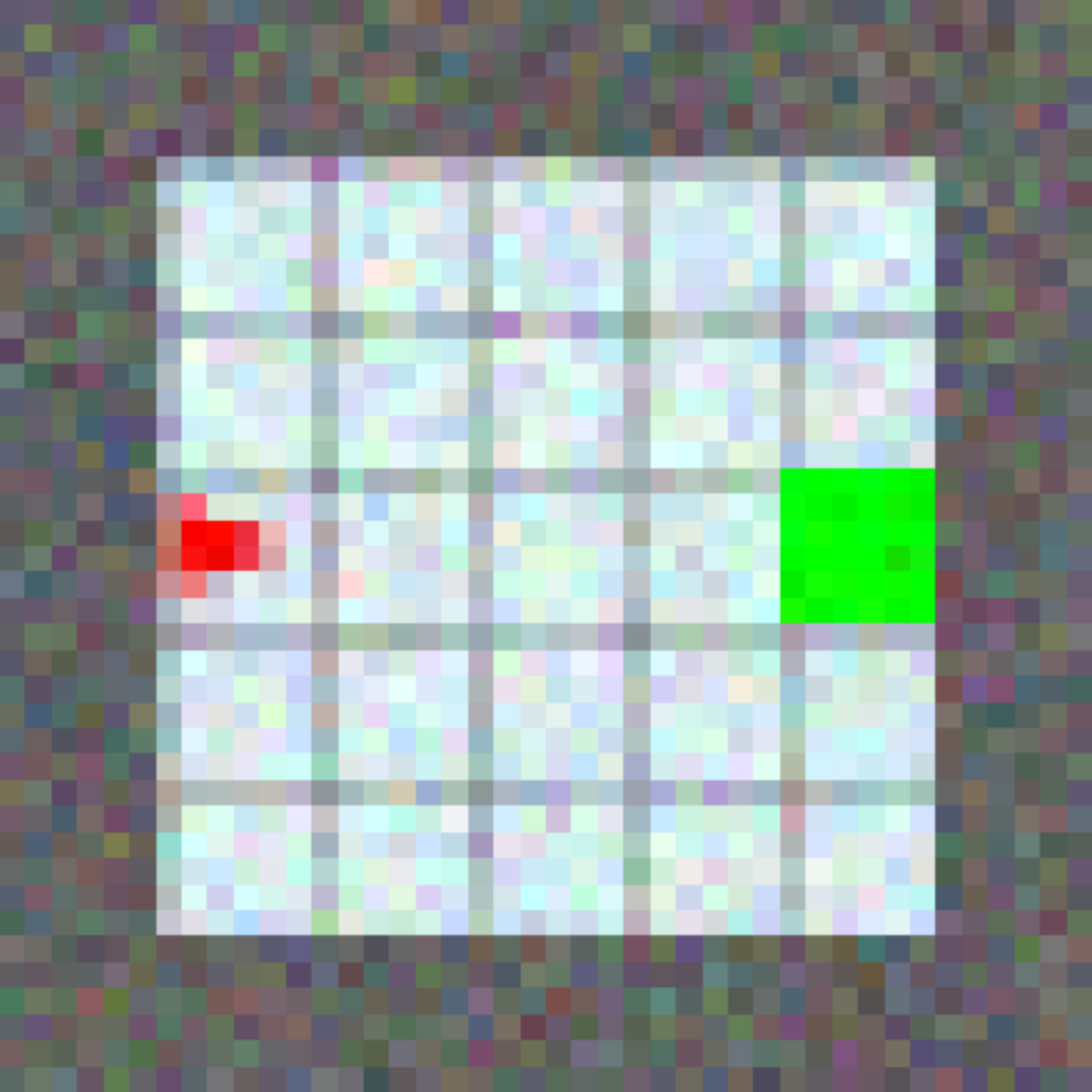}  
        \includegraphics[width=0.49\textwidth]{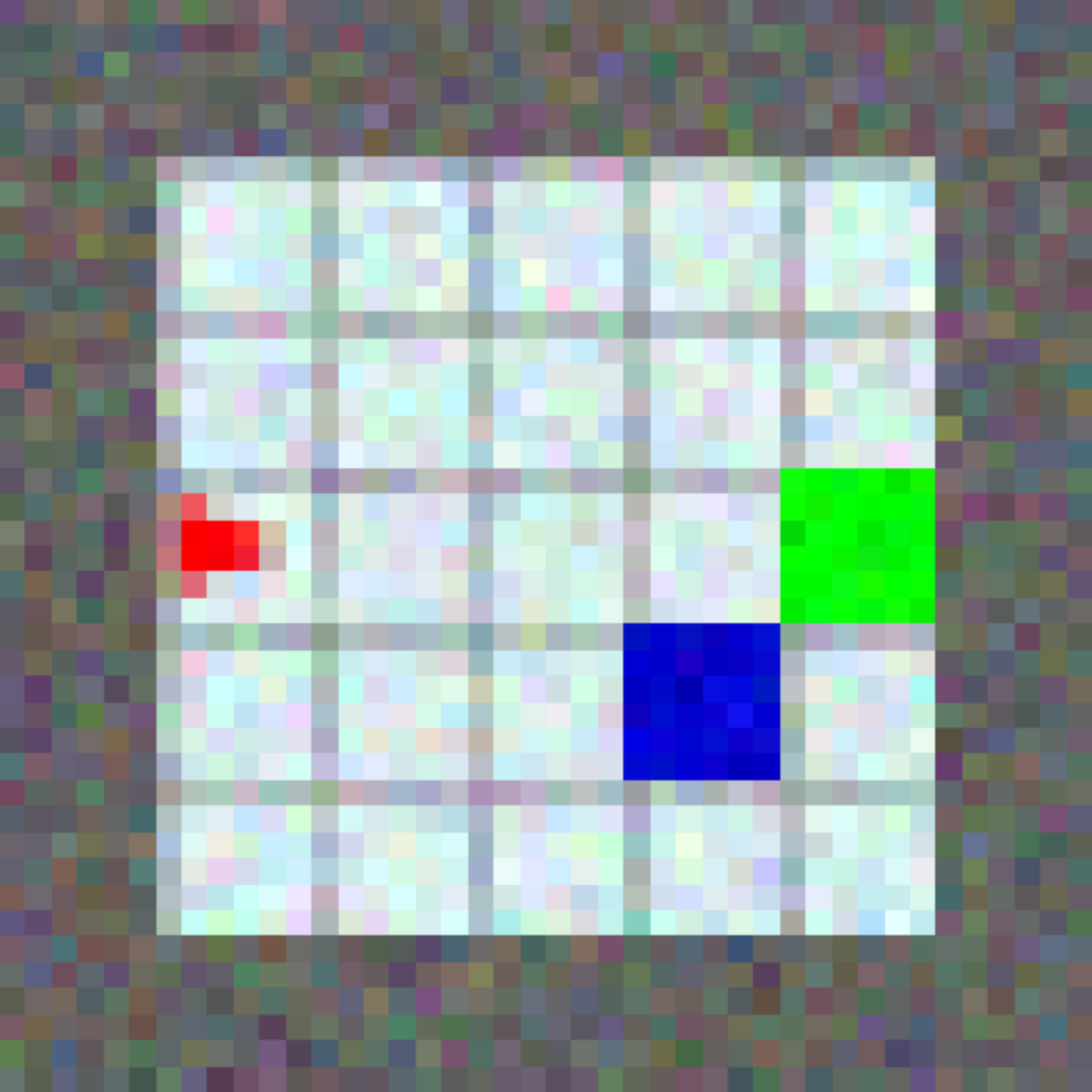} 
        \caption{Frozen Lake.}
        \label{fig:gridworld_asym:IceLake:im}
    \end{subfigure}\hfill%
    \begin{subfigure}[t]{0.23\textwidth}  
        \includegraphics[width=0.49\textwidth]{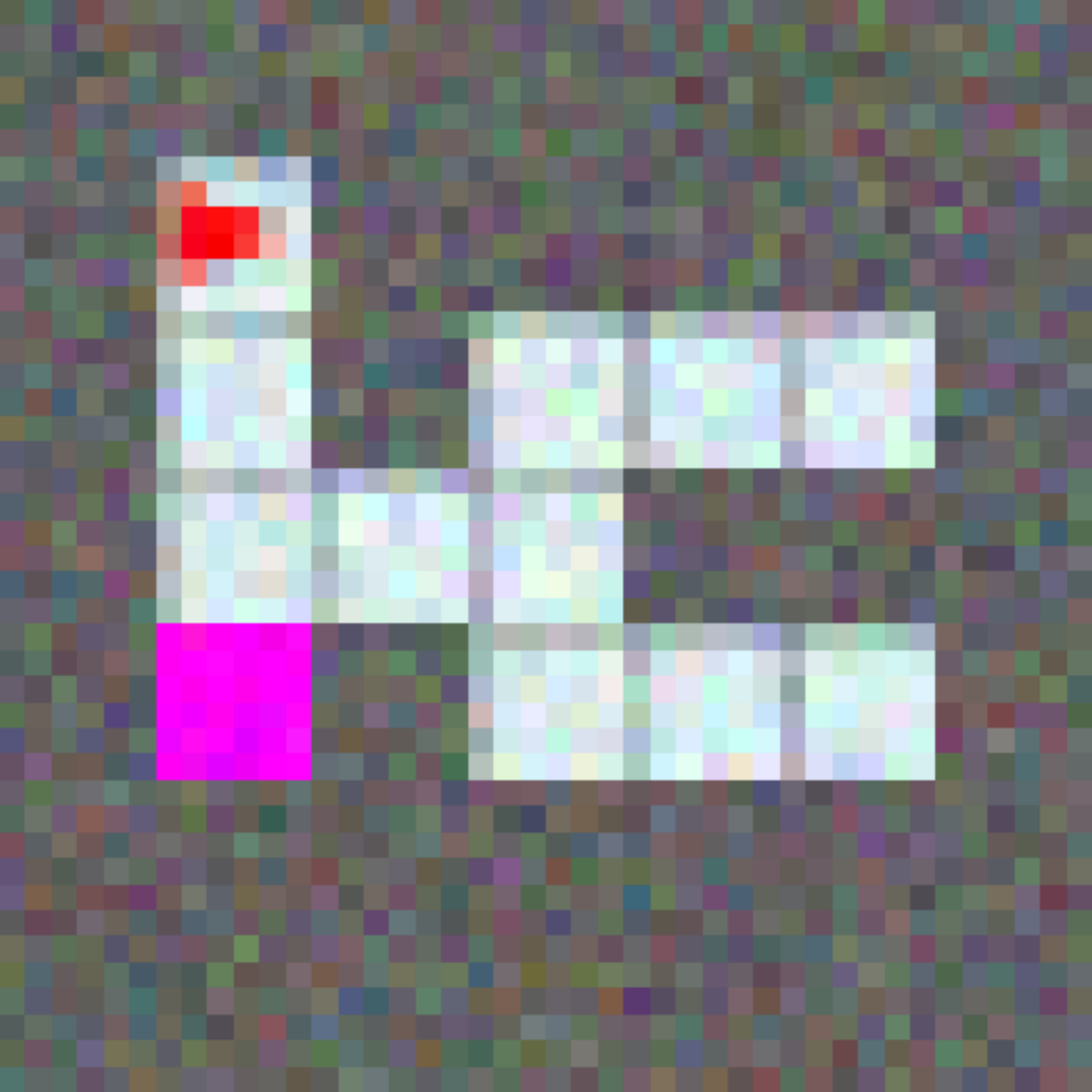}
        \includegraphics[width=0.49\textwidth]{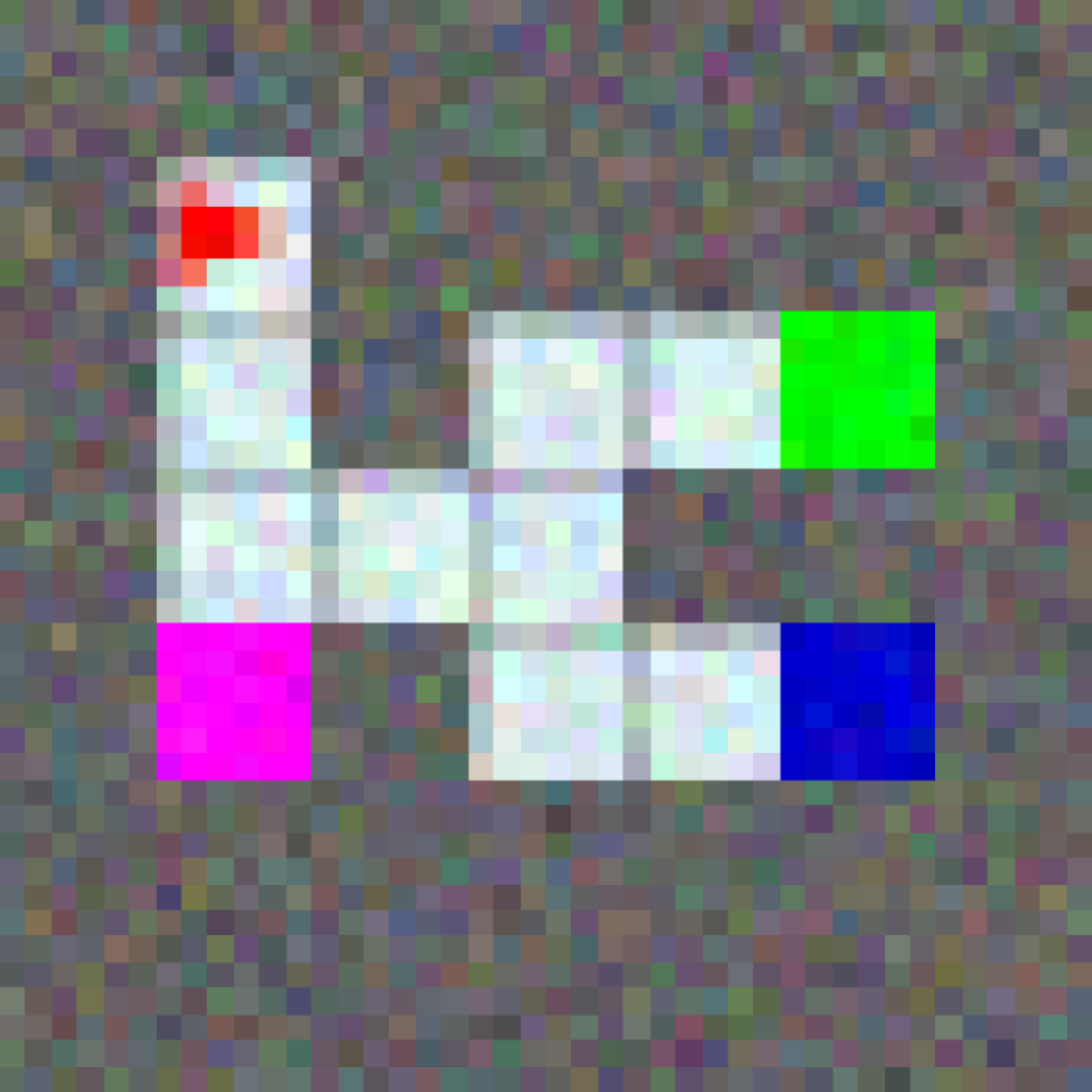} 
        \caption{Tiger Door.}
        \label{fig:gridworld_asym:TigerDoor:im}
    \end{subfigure}
    \caption{The two gridworlds we study.  An agent (red) must navigate to the goal (green) while avoiding the hazard (blue).  Shown are the raw, noisy $42\times 42$ pixel observations available to the agent.  The expert is conditioned on an omniscient compact state vector indicating the position of the goal and hazard.  In Frozen Lake, the trainee is conditioned on the left image and cannot see the hazard.  In Tiger Door, pushing the button (pink) illuminates the hazard.  } 
    \label{fig:gridworld_asym:grids}
\end{figure}

\section{Failure of Asymmetric Imitation Learning}
\label{sec:il-failure}
We now reason about the failure of AIL in terms of \emph{reward}.  The crucial insight is that to guarantee that the reward earned by the trainee policy is optimal, the divergence between expert and trainee must go to exactly zero.  The reward earned by policies with even a small (but finite) divergence may be arbitrarily low.  This condition, referred to as \emph{identifiability}, is formalized below.  We leverage this condition in Section \ref{sec:algorithm} to derive the update applied to the expert which guarantees the optimal partially observed policy is recovered under the assumptions specified by each theorem, and discussed in further detail in Appendix \ref{supp:thoery}.

\begin{figure*}[t!]
    \centering
    \begin{subfigure}[b]{0.36\textwidth}
        \centering
        \includegraphics[width=\textwidth]{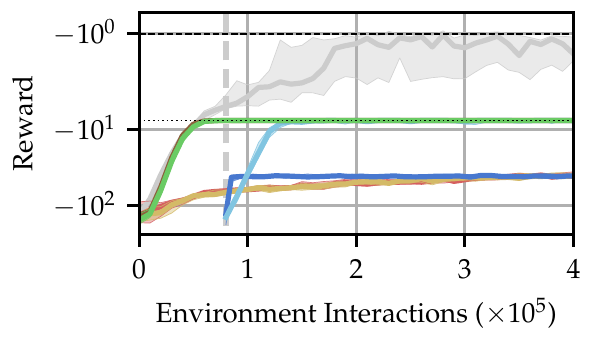}
        \caption{Frozen Lake.}
        \label{fig:gridworld_asym:IceLake}
    \end{subfigure}%
    \hfill
    \begin{subfigure}[b]{0.36\textwidth}
        \centering
        \includegraphics[width=\textwidth]{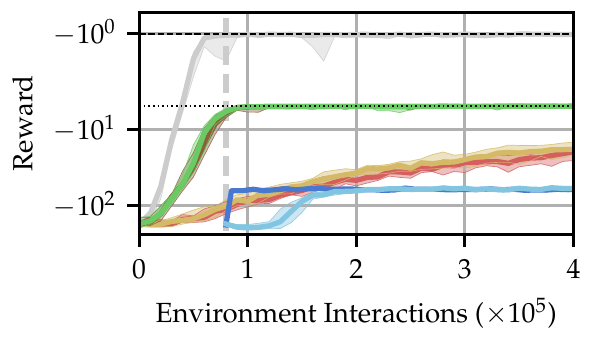}
        \caption{Tiger Door.}
        \label{fig:gridworld_asym:TigerDoor}
    \end{subfigure}%
    \hfill
    \begin{subfigure}[b]{0.18\textwidth}
        \centering
       \includegraphics[width=\textwidth]{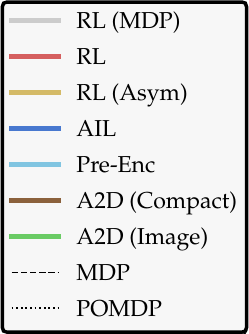}
    \end{subfigure}
    \caption{Results for the gridworld environments.  Median and quartiles across $20$ random seeds are shown.  TRPO~\citep{schulman2015trust} is used for RL methods.  Broken lines indicate the optimal reward, normalized so the optimal MDP reward is $-1$ (\emph{MDP}).  All agents and trainees are conditioned on a image-based input, except \emph{A2D (Compact)} which is conditioned on a partial compact state representation.  All experts, and \emph{RL (MDP)}, are conditioned on an omniscient compact state.  \emph{Pre-Enc} uses a fixed pretrained image encoder, trained on examples from the MDP.  \emph{AIL} and \emph{Pre-Enc} begin when the MDP has converged, as this is the required expenditure for training.  A2D is the only method that reliably and efficiently finds the optimal POMDP policy, and, in a sample budget comparable with \emph{RL (MDP)}.  The convergence of A2D is also similar for \emph{both} image-based (\emph{A2D (Image)}) and compact (\emph{A2D (Compact)}) representations, highlighting that we have effectively subsumed the image perception task.  Configurations, additional results and discussions are included in the appendix.}
    \label{fig:gridworld_asym}
\end{figure*}

However, to first motivate and explore this behavior, we introduce two pedagogical environments, referred to as ``Frozen Lake'' and ``Tiger Door''~\citep{littman1995pomdp,Spaan2012}, illustrated in Figure \ref{fig:gridworld_asym:grids}. Both require an agent to navigate to a goal while avoiding hazards.  The trainee is conditioned on an image of the environment where the hazard is not initially visible. The expert is conditioned on an omniscient compact state vector.  Taking actions, reaching the goal, and hitting the hazard incurs rewards of $-2$, $20$, and $-100$ respectively.  In Frozen Lake, the hazard (weak ice) is in a random location in the interior nine squares.  In Tiger Door, the agent can detour via a button, incurring additional negative reward, to reveal the goal location.

We show results for application of AIL, and comparable RL approaches, to these environments in Figure \ref{fig:gridworld_asym}.  These confirm our intuitions:   RL in the MDP (\emph{RL (MDP)}) is stable and efficient, and proceeds directly to the goal, earning maximum rewards of $10.66$ and $6$.  Direct RL in the POMDP (\emph{RL} and \emph{RL (Asym)}) does not converge to a performant policy in the allocated computational budget.  AIL (\emph{AIL}) converges almost immediately, but, to a trainee that averages over expert actions.  In Frozen Lake, this trainee averages the expert over the location of the weak patch, never circumnavigates the lake, and instead crosses directly, incurring an average reward of $-26.6$.  In Tiger Door, the trainee proceeds directly to a possible goal location without pressing the button, incurring an average reward of $-54$.  Both solutions represent catastrophic failures.  Instead, the trainee should circumnavigate the lake, or, push the button and then proceed to the goal, earning rewards of $4$ and $2$ respectively. 

These results, and insight from Theorem \ref{def:ail}, lead us to define two important properties which provide guarantees on the performance of AIL:
\begin{definition}[Identifiable Policies]
\label{def:consistency_pol} 
Given an MDP-POMDP pair $\left\lbrace \mathcal{M}_{\Theta}, \mathcal{M}_{\Phi} \right\rbrace$, an MDP policy $\pi_{\theta} \in \Pi_{\Theta}$, and POMDP policy $\pi_{\phi} \in \Pi_{\Phi}$, we describe $\{\pi_{\theta}, \pi_{\phi}\}$ as an \textbf{identifiable policy pair} if and only if $\mathbb{E}_{d^{\pi_{\phi}}(s,b)} \left[ \mathbb{KL} \left[ \pi_{\theta}(a | s) || \pi_{\phi}(a | b) \right] \right] = 0$.  
\end{definition}
\begin{definition}[Identifiable Processes]
\label{def:consistency_proc} 
If each optimal MDP policy, $\pi_{\theta^*} \in \Pi_{\Theta^*}$, and the corresponding implicit policy, $\hat{\pi}_{\theta^*} \in \hat{\Pi}_{\Theta^*}$, form an identifiable policy pair, then we define $\{\mathcal{M}_{\Theta}, \mathcal{M}_{\Phi}\}$ as an \textbf{identifiable process pair}.  
\end{definition}
Identifiable policy pairs enforce that the partially observing implicit policy, recovered through application of AIL, can \emph{exactly} reproduce the actions of the fully observing policy.  These policies are therefore guaranteed to incur the same reward.  Identifiable processes then extends this definition, requiring that such an identifiable policy pair exists for all optimal fully observing policies.  Using this definition, we can then show that performing AIL using any optimal fully observing policy on an identifiable process pair is guaranteed to recover an optimal partially observing policy: 
\begin{theorem}[Convergence of AIL]
\label{thm:consistency_conv}
For any identifiable process pair defined over sufficiently expressive policy classes, under exact intermediate updates, the iteration defined by:
\begin{equation}
    \hspace*{-0.1cm} \psi_{k+1} \!= \!\mathop{\argmin}_{\psi \in \Psi} \!\! \mathop{\mathbb{E}}_{d^{\pi_{\psi_k}}(s,b)} \! \left[  \mathbb{KL} \left[ \pi_{\theta^*}(a|s) || \pi_\psi(a|b) \right] \right],
\end{equation}
where $\pi_{\theta^*}$ is an optimal fully observed policy, converges to an optimal partially observed policy, $\pi_{\psi^*}(a|b)$, as $k \tends \infty$.
\end{theorem}
\begin{proof}
See Appendix \ref{supp:thoery}.
\end{proof}
Therefore, identifiability of processes defines a sufficient condition to guarantee that any optimal expert policy provides asymptotically unbiased supervision to the trainee.  If a process pair is identifiable, then AIL recovers the optimal partially observing policy, and garners a reward equal to the fully observing expert. When processes are not identifiable, the divergence between expert and trainee is non-zero, and the \emph{reward} garnered by the trainee can be arbitrarily sub-optimal (as in the gridworlds above).  Unfortunately, identifiability of two processes represents a strong assumption, unlikely to hold in practice. Therefore, we propose an extension that modifies the \emph{expert} on-line, such that the modified expert policy and corresponding implicit policy pair form an identifiable \emph{and} optimal policy pair under partial information. This modification, in turn, guarantees that the expert provides asymptotically correct AIL supervision.

% % \aw{even if the divergence has been minimized within the achievable policy class.}

% However, identifiability of two processes is a fixed property of the processes, and represents a strong condition that is unlikely to hold in general, and hence the performance of AIL cannot be guaranteed.  

% We therefore propose an extension that modifies the \emph{expert} on-line, such that the modified expert policy and corresponding implicit policy pair form an identifiable \emph{and} optimal policy pair (under partial information), thereby guaranteeing that the expert provides asymptotically correct AIL supervision.

% % huge final sentence needs to be broken up.

\section{Correcting AIL with Expert Refinement}
\label{sec:algorithm}

We now use the insight from Sections \ref{sec:prelim} and \ref{sec:il-failure} to construct an update, applied to the expert policy, which improves the expected reward ahead under the implicit policy.  Crucially, this update is designed such that, when interleaved with AIL, the optimal partially observed policy is recovered.  We refer to this iterative algorithm as adaptive asymmetric DAgger (A2D). To derive the update to the expert, $\pi_{\theta}$, we first consider the RL objective under the implicit policy, $\hat{\pi}_{\theta}$:
\begin{align}
     &J(\theta) = \mathop{\mathbb{E}}_{d^{{\hat{\pi}_{\theta}}}(b) \hat{\pi}_\theta(a|b) } \left[ Q^{{\hat{\pi}_{\theta}}}(a,b) \right], \where \\
    & Q^{{\hat{\pi}_{\theta}}}(a,b) =\!\!\!\! \mathop{\mathbb{E}}_{ p(b',s',s|a,b)} \!\bigg[ r(s,a,s') + \!\!\! \mathop{\gamma\mathbb{E}}_{\hat{\pi}_\theta(a'|b')} \! \left[ Q^{\hat{\pi}_\theta} (a', b') \right] \bigg] \!. \nonumber 
\end{align}
This objective defines the cumulative reward of the trainee in terms of the parameters of the expert policy. This means that maximizing $J(\theta)$ maximizes the reward obtained by the implicit policy, and ensures proper expert supervision:
\begin{theorem}[Convergence of Exact A2D]
\label{thm:conv_exact} 
Under exact intermediate updates, the following iteration converges to an optimal partially observed policy $\pi_{\psi^*}(a|b)\in\Pi_{\Psi}$, provided both  $\Pi_{\Phi^*} \subseteq \hat{\Pi}_{\Theta^*} \subseteq \Pi_{\Psi}$:
\begin{align}
    \!&\psi_{k+1} \! = \! \mathop{\argmin}_{\psi \in \Psi} \!\mathop{\mathbb{E}}_{d^{\pi_{\psi_k}}(s,b)} \!\! \left[ \mathbb{KL} \left[ \pi_{\hat{\theta}^*}(a|s) || \pi_\psi(a|b) \right] \right], \label{eq:implicit_ail_argmin}\\
    &\where  \hat{\theta}^* = \mathop{\argmax}_{\theta \in \Theta} \mathop{\mathbb{E}}_{\hat{\pi}_{\theta}(a | b) d^{{\pi_{\psi_k}}}(b)} \left[Q^{{\hat{\pi}_{\theta}}}(a,b) \right]. \label{eq:implicit_rl_argmin}
\end{align}
\end{theorem}
\vspace{-0.2cm}
\begin{proof}
\vspace{-0.2cm}
See Appendix \ref{supp:thoery}.
\end{proof}
\vspace{-0.2cm}
First, an inner optimization, defined by \eqref{eq:implicit_rl_argmin}, maximizes the expected reward of the implicit policy by updating the parameters of the \emph{expert} policy, under the current trainee policy.  The outer optimization, defined by \eqref{eq:implicit_ail_argmin}, then updates the trainee policy by projecting onto the updated implicit policy defined by the updated expert. This projection is performed by minimizing the divergence to the updated expert, as per Theorem \ref{def:ail}.  %\aw{halp}
% This algorithm can be considered as a projected optimization~\cite{bertsekas2014constrained}. 

Unfortunately, directly differentiating through $Q^{\hat{\pi}_\theta}$, or even sampling from $\hat{\pi}_\theta$, is intractable.  We therefore optimize a surrogate reward instead, denoted $J_{\psi}(\theta)$, that defines a lower bound on the objective function in \eqref{eq:implicit_rl_argmin}.  This surrogate is defined as the expected reward ahead under the variational trainee policy $Q^{{\pi_{\psi}}}$.  By maximizing this surrogate objective, we maximize a lower bound on the possible improvement to the implicit policy with respect to the parameters of the expert:
\begin{align}
    \max_{\theta \in \Theta} J_{\psi}(\theta) &= \max_{\theta \in \Theta} \mathop{\mathbb{E}}_{{\hat{\pi}_\theta(a|b) d^{{\pi_{\psi}}}(b)}} \left[ Q^{{\pi_{\psi}}}(a,b) \right] \label{equ:bound:1}\\
    \leq \max_{\theta \in \Theta} J(\theta) &= \max_{\theta \in \Theta}  \mathop{\mathbb{E}}_{{\hat{\pi}_\theta(a|b) d^{{\pi_{\psi}}}(b)}} \left[ Q^{{\hat{\pi}_{\theta}}}(a,b) \right] . \label{equ:bound:2}
\end{align}
To verify this inequality, first note that we assume that the implicit policy is capable of maximizing the expected reward ahead at every belief state (c.f. Theorem \ref{thm:conv_exact}).  Therefore, by definition, replacing the implicit policy, $\hat{\pi}_{\theta}$, with any \emph{behavioral policy}, here $\pi_{\psi}$, cannot yield \emph{larger} returns when maximized over $\theta$ (see Appendix \ref{supp:thoery}).  Replacement with a behavioral policy is a common analysis technique, especially in policy gradient~\citep{schulman2015trust,schulman2017proximal,sutton1992reinforcement} and policy search methods (see \S 4,5 of \citet{bertsekas2019reinforcement} and \S 2 of \citet{deisenroth2013survey}).  This surrogate objective permits the following REINFORCE gradient estimator, where we define $f_{\theta} = \log \pi_{\theta}(a \mid s)$:
\begin{align}
    \nabla_{\theta} & J_{\psi}(\theta) =  \nabla_{\theta} \mathop{\mathbb{E}}\nolimits_{{\hat{\pi}_\theta(a|b) d^{{\pi_{\psi}}}(b)}} \left[ Q^{{\pi_{\psi}}}(a,b) \right] \\
    &= \mathop{\mathbb{E}}\nolimits_{{d^{\pi_{\psi}}(b)}} \left[ \nabla_{\theta} \mathop{\mathbb{E}}\nolimits_{{ d^{{\pi_{\psi}}}(s | b)}} \left[ 
        \mathop{\mathbb{E}}\nolimits_{{\pi_\theta(a|s)}} \left[ Q^{{\pi_{\psi}}}(a,b) \right] \right] \right] \nonumber \\
    &= \mathop{\mathbb{E}}\nolimits_{{d^{{\pi_{\psi}}}(s, b)}} \left[ \mathop{\mathbb{E}}\nolimits_{\pi_{\theta}(a|s)} \left[ 
        Q^{{\pi_{\psi}}}(a,b) \nabla_\theta f_\theta \right] \right] \nonumber \\
    &= \mathop{\mathbb{E}}\nolimits_{d^{\pi_{\psi}}(s, b) \pi_{\psi}(a | b)} \left[ \frac{\pi_{\theta}(a|s)}{\pi_{\psi}(a | b)} Q^{{\pi_{\psi}}}(a,b) \nabla_\theta f_{\theta} \right]. \label{equ:expert-gradient}
\end{align}
Equation \eqref{equ:expert-gradient} defines an importance weighted policy gradient, evaluated using states sampled under the variational agent, which is equal to the gradient of the implicit policy reward with respect to the expert parameters.  For \eqref{equ:expert-gradient} to provide an unbiased gradient estimate we (unsurprisingly) require an unbiased estimate of $Q^{\pi_{\psi}}(a,b)$.  While, this estimate can theoretically be generated by directly learning the Q function using a universal function approximator, in practice, learning the Q function is often challenging.  Furthermore, the estimator in \eqref{equ:expert-gradient} is \emph{strongly} dependent on the quality of the approximation.  As a result, imperfect Q function approximations yield biased gradient estimates.

This strong dependency has led to the development of RL algorithms that use Monte Carlo estimates of the Q function instead.  This circumvents the cost, complexity and bias induced by approximating Q, by leveraging these rollouts to provide unbiased, although higher variance, estimates of the Q function.  Techniques such as generalized advantage estimation (GAE)~\citep{schulman2015high} allow bias and variance to be traded off. However, as a direct result of asymmetry, using Monte Carlo rollouts in A2D can bias the gradient estimator.  Full explanation of this is somewhat involved, and so we defer discussion to Appendix \ref{supp:exp}.  However, we note that for most \emph{environments} this bias is small and can be minimized through tuning the parameters of GAE.  The final gradient estimate used in A2D is therefore:
\begin{align}
    &\nabla_\theta J_{\psi}(\theta) = \mathop{\mathbb{E}}_{\substack{d^{\pi_{\beta}}(s_t, b_t) \\ \pi_{\beta}(a_t | s_t, b_t)}} \left[ \frac{\pi_{\theta}(a_t|s_t) }{\pi_{\beta}(a_t | s_t, b_t)} \hat{A}^{\pi_{\beta}} \nabla_\theta f_{\theta} \right] , \label{equ:a2d:a2d_update} \\
    & \mathrm{where} \quad \hat{A}^{\pi_{\beta}}(a_t,s_t,b_t) = \sum\nolimits_{t=0}^{\infty} (\gamma \lambda)^t \delta_t , \label{equ:a2d:gae_1}\\
    & \mathrm{and} \quad \delta_t = r_t + \gamma V^{\pi_{\beta}}(s_{t+1}, b_{t+1}) - V^{\pi_{\beta}}(s_t, b_t) , \label{equ:a2d:gae}
\end{align}
where \eqref{equ:a2d:gae_1} and \eqref{equ:a2d:gae} describe GAE~\citep{schulman2015high}.  Similar to DAgger, we also allow A2D to interact under a mixture policy, $\pi_{\beta}(a | s, b) = \beta \pi_{\theta}(a | s) + (1 - \beta) \pi_{\psi}(a | b)$, with Q and value functions defined as $Q^{\pi_\beta}(a, s, b)$ and $V^{\pi_\beta}(a, s, b)$ similarly.  However, as was also suggested by \cite{Ross2011}, we found that aggressively annealing $\beta$, or even setting $\beta = 0$ immediately, often provided the best results.  The full A2D algorithm, also shown in Algorithm \ref{alg:a2d}, is implemented by repeating three individual steps:
\begin{enumerate}[topsep=0pt]
    \item \textbf{Gather data} (Alg. \ref{alg:a2d}, Ln 8): Collect samples from $q_{\pi_{\beta}}(\tau)$ by rolling out under the mixture, defined in \eqref{equ:background:pomdp_dist}.
    \item \textbf{Refine Expert} (Alg. \ref{alg:a2d}, Ln 11):  Update expert policy parameters, $\theta$, with importance weighted policy gradient as estimated in \eqref{equ:a2d:a2d_update}.  This step also updates the trainee and expert value function parameters, $\nu_p$ and $\nu_m$.
    \item \textbf{Update Trainee} (Alg. \ref{alg:a2d}, Ln 12): Perform an AIL step to fit the (variational) trainee policy parameters, $\psi$, to the expert policy using \eqref{equ:def:variational:gradient}. 
\end{enumerate}
As the gradient used in A2D, defined in \eqref{equ:a2d:a2d_update}, is a REINFORCE-based gradient estimate, it is compatible with any REINFORCE-based policy gradient method, such as TRPO or PPO~\cite{schulman2015trust, schulman2017proximal}.  Furthermore, A2D does not require pretrained experts or example trajectories.  In the experiments we present, all expert and trainee policies are learned from scratch.  Although using A2D with pretrained expert policies is possible, such pipelined approaches are susceptible to suboptimal local minima.  
 
\begin{figure}[t]
\begin{minipage}{0.46\textwidth}
    \vspace{-0.25cm}
    \begin{algorithm}[H]
        \setstretch{1.15}
        \small %\small, \footnotesize, \scriptsize, or \tiny
        \caption{Adaptive Asymmetric DAgger (A2D)}
        \label{alg:a2d}
        \begin{algorithmic}[1]
          \STATE {\bfseries Input:} MDP $\mathcal{M}_{\Theta}$, POMDP $\mathcal{M}_{\Phi}$, Annealing schedule $\texttt{AnnealBeta}(n, \beta)$.
          \STATE {\bfseries Return:} Variational trainee parameters $\psi$.
          \STATE $\theta, \psi, \nu_m, \nu_p, \gets \texttt{InitNets} \left(\mathcal{M}_{\Theta}, \mathcal{M}_{\Phi} \right)$
          \STATE $\beta \gets 1,\ D \gets \emptyset$ 
          \FOR {$n = 0,\ \dots,\ N$}
            \STATE $\beta \gets \texttt{AnnealBeta}\left(n, \beta\right)$
            \STATE $\pi_{\beta} \gets \beta  \pi_{\theta}  + (1 - \beta) \pi_{\psi}$
            \STATE $\mathcal{T} = \{\tau_i\}_{i=1}^\mathcal{I} \sim q_{\pi_{\beta}} (\tau)$ \label{ln:alg:a2d:q}
            \STATE $D \gets \texttt{UpdateBuffer}\left(D, \mathcal{T} \right)$
            \STATE \textcolor{blue}{$V^{\pi_{\beta}} \gets \beta V^{\pi_{\theta}}_{\nu_m} + (1 - \beta) V^{\pi_{\psi}}_{\nu_p}$}  \label{ln:alg:a2d:rl_v}
            \STATE \textcolor{blue}{$\theta, \nu_m, \nu_p \gets \texttt{RLStep} \left( \mathcal{T}, V^{\pi_{\beta}}, \pi_{\beta} \right)$} \label{ln:alg:a2d:rl_p}
            \STATE $\psi \gets \texttt{AILStep}\left(D, \pi_{\theta}, \pi_{\psi} \right)$ \label{ln:alg:a2d:proj}
          \ENDFOR
        \end{algorithmic}
    \end{algorithm}
\end{minipage}
\vspace{-0.2cm}
\setcounter{algorithm}{0}
\captionof{algorithm}{Adaptive asymmetric DAgger (A2D) algorithm.  Additional steps we introduce beyond DAgger~\citep{Ross2011} are highlighted in blue, and implement the feedback loop in Figure \ref{fig:a2d}.  \texttt{RLStep} is a policy gradient step, updating the expert, using the gradient estimator in \eqref{equ:a2d:a2d_update}.  \texttt{AILStep} is an AIL variational policy update, as in \eqref{equ:def:variational:gradient}.}
\end{figure}

\begin{figure*}[!htb]

    \begin{subfigure}[t]{0.48\textwidth}
        \includegraphics[width=0.95\textwidth]{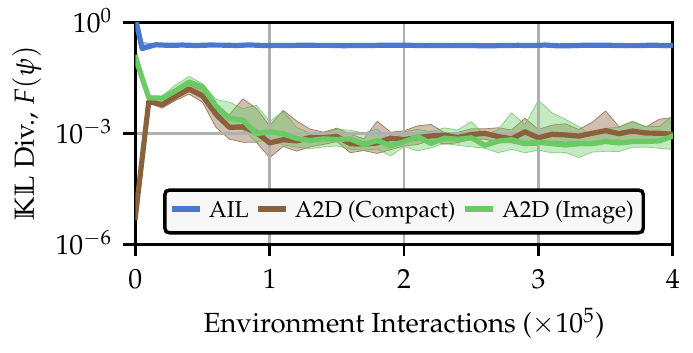}
        \caption{Frozen Lake.}
        \label{fig:grid:a2dplot:lg}
    \end{subfigure}%
    \hfill%
    \begin{subfigure}[t]{0.48\textwidth}
        \includegraphics[width=0.95\textwidth]{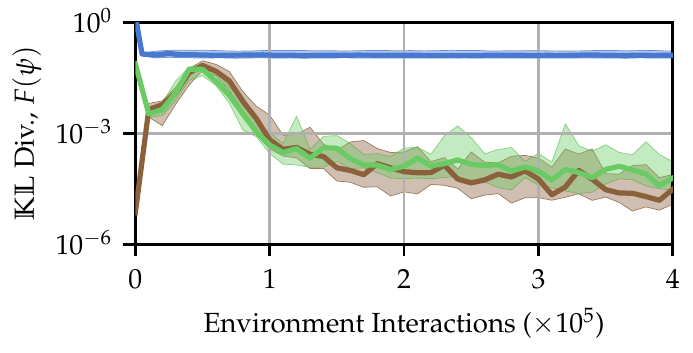}
        \caption{Tiger Door.}
        \label{fig:grid:a2dplot:td}
    \end{subfigure}%
    \vspace{-0.2cm}
    \caption{The evolution of the policy divergence, $F(\psi)$.  Shown are median and quartiles across $20$ random seeds.  \emph{AIL} converges to a high divergence, whereas A2D achieves a low divergence for both representations, indicating that the trainee recovered by A2D is faithfully imitating the expert (see Figure \ref{fig:gridworld_asym} for more information). 
    { }%
    }
    \label{fig:grid:a2dplot}
\end{figure*}

\section{Experiments}
\label{sec:results}

\subsection{Revisiting Frozen Lake \& Tiger Door}
\label{sec:results:gridworld} 
We evaluate A2D on the gridworlds introduced in Section \ref{sec:prelim}.  Results are shown in Figures \ref{fig:gridworld_asym} and \ref{fig:grid:a2dplot}.  Figure \ref{fig:gridworld_asym} shows that A2D converges to the optimal POMDP reward quickly, and, in a comparable number of interactions to the best-possible convergence of RL in the MDP when using similar hyperparameters to those used for A2D (\emph{RL (MDP)}).  Convergence rates are also similar for high-dimensional images (\emph{A2D (Image)}) and low-dimensional representations (\emph{A2D (Compact)}).  Other methods fail for one, or both, gridworlds.  A2D can also operate with hyperparameters broadly similar to those tuned specifically for RL in the MDP, where tuning is easy.  However, A2D did then benefit from increased batch size, entropy regularization, and reduced $\lambda$ (see Appendix \ref{supp:exp}).  The IL hyperparameters are largely independent of the RL hyperparameters, further simplifying tuning overall.  

\begin{figure*}[!htb]
    \centering
    \includegraphics[width=0.245\textwidth]{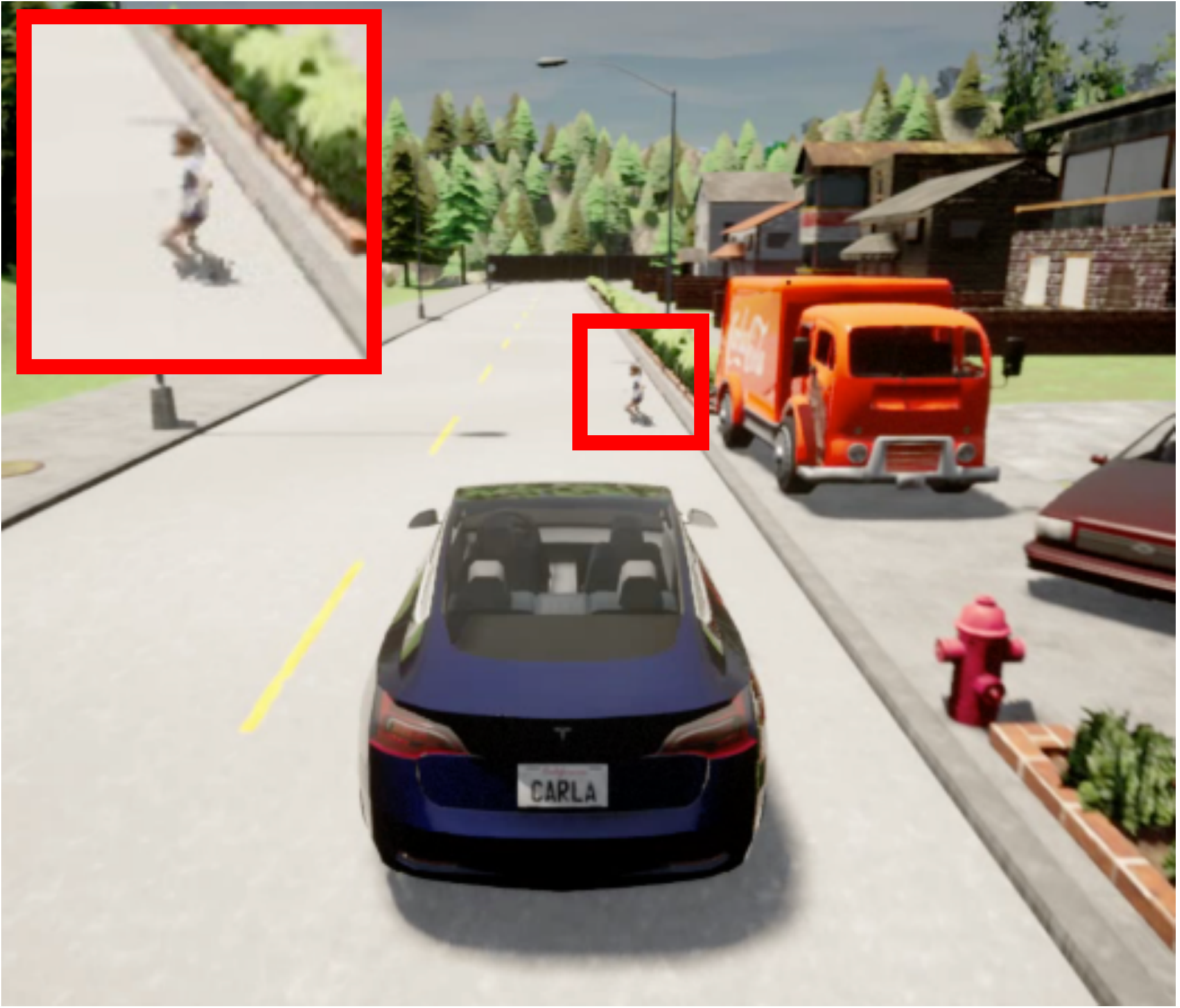} \hfill
    \includegraphics[width=0.2105\textwidth]{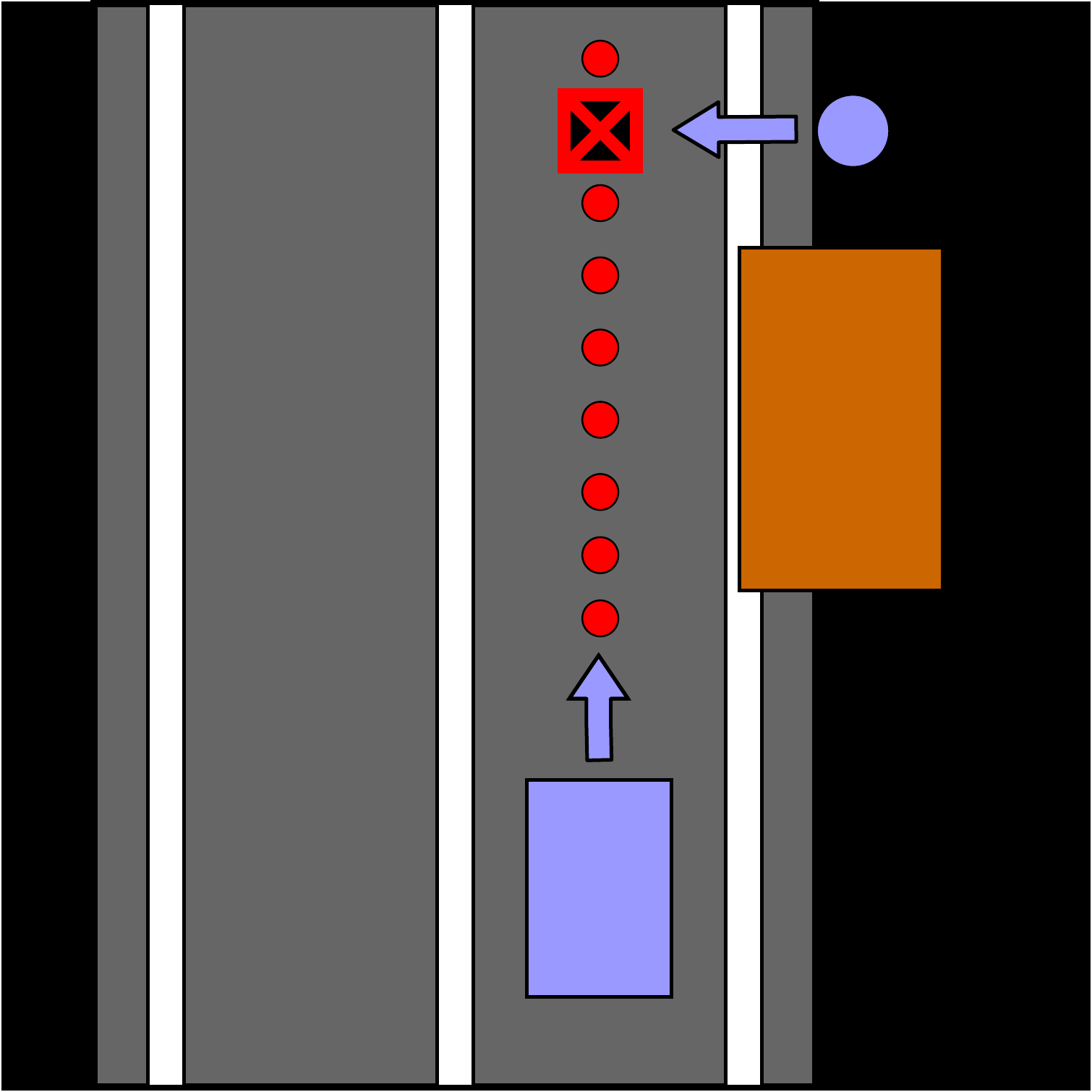} \hfill
    \includegraphics[width=0.48\textwidth]{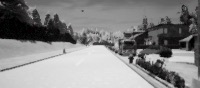}
    \caption{Visualizations of the AV scenario.  Left: third-person view showing the egovehicle and child running out.  Center: top-down schematic of the environment and asymmetric information.  Right: front-view camera input provided to the agent.}
    \vspace*{-0.2cm}
    \label{supp:fig:carla:a2dplot:inputs}
\end{figure*}

Figure \ref{fig:grid:a2dplot} shows the divergence between the expert and trainee policies during learning.  \emph{AIL} saturates to a high divergence, indicating that the trainee is unable to replicate the expert.  The divergence in A2D increases initially, as the expert learns using the full-state information.  This rise is due to the non-zero value of $\beta$, imperfect function approximation, slight bias in the gradient estimator, and the tendency of the expert to initially move towards a higher reward policy not representable under the agent.  As the learning develops, and $\beta \tends 0$, the expert is forced to optimize the reward of the trainee.  This, in turn, drives the divergence towards zero, producing a policy that can be represented by the agent.  A2D has therefore created an identifiable expert and implicit policy pair (Definition \ref{def:consistency_pol}), where the implicit policy is also optimal under partial information.

\subsection{Safe Autonomous Vehicle Learning}
\label{sec:results:carla}
Autonomous vehicle (AV) simulators~\citep{Dosovitskiy17, wymann_torcs_2014, kato_open_2015} allow safe virtual exploration of driving scenarios that would be unsafe to explore in real life.  The inherent complexity of training AV controllers makes exploiting efficient AIL an attractive opportunity~\citep{Chen2019}.  The expert can be provided with the exact state of other actors, such as other vehicles, occluded hazards and traffic lights.  The trainee is then provided with sensor measurements available in the real world, such as camera feeds, lidar and the egovehicle telemetry.  

%CARLA challenge traffic scenario 3 \& 4
The safety-critical aspects of asymmetry are highlighted in context of AVs. Consider a scenario where a child may dart into the road from behind a parked truck, illustrated in Figure \ref{supp:fig:carla:a2dplot:inputs}.  The expert, aware of the position and velocity of the child from asymmetric information, will only brake if there is a child, and will otherwise proceed at high speed.  However, the trainee is unable to distinguish between these scenarios, before the child emerges from, just the front-facing camera.  As the expected expert behavior is to accelerate, the implicit policy also accelerates.  The trainee only starts to brake once the child is visible, by which time it is too late to guarantee the child is not struck.  The expert should therefore proceed at a lower speed so it can slow down or evade the child once visible. This cannot be achieved by naive application of AIL.  

\begin{figure*}[!htb]
    \includegraphics[width=0.315\textwidth]{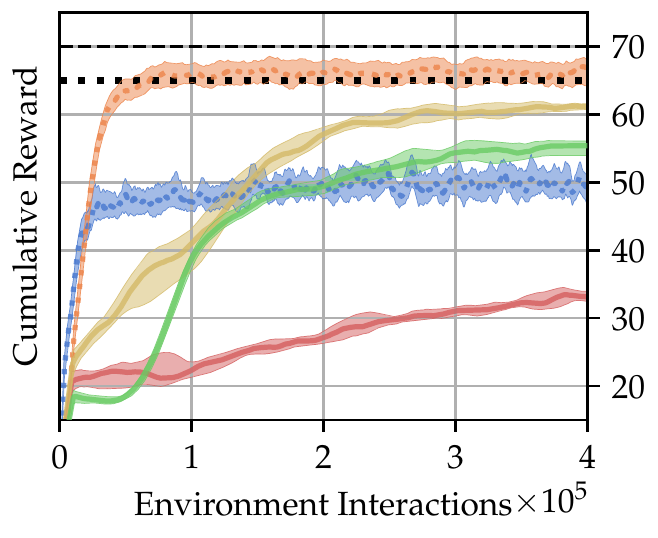}
    \includegraphics[width=0.335\textwidth]{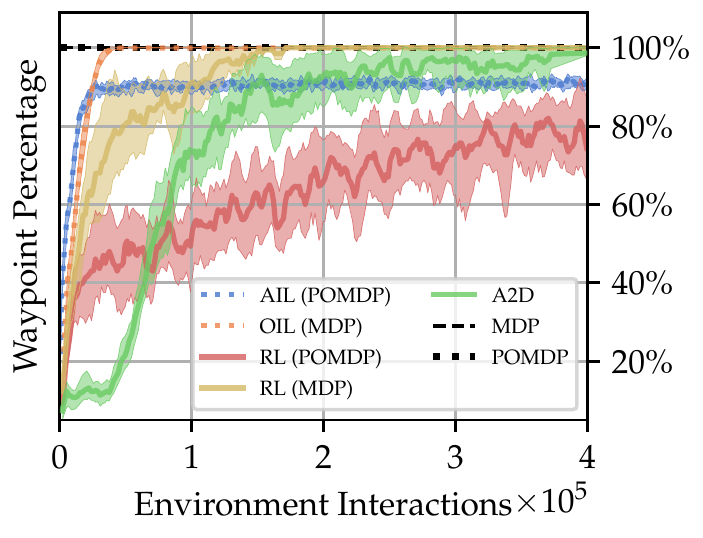}
    \includegraphics[width=0.33\textwidth]{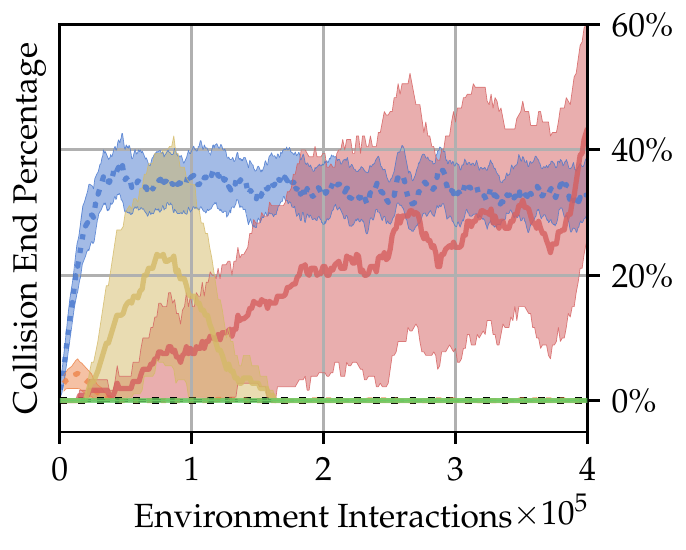}
    \caption{Performance metrics for the AV scenario, introduced in Section \ref{sec:results}.  We show median and quartiles across ten random seeds.  Left: average cumulative reward.  Center: average percentage of waypoints collected, measuring progress along route.  Right: percentage of trajectories ending in a child collision.  Optimal MDP and POMDP solutions are shown by dashed and dotted lines respectively.  In methods marked as MDP the agent uses an omniscient compact state, including the child's state.  AIL (\emph{AIL (MDP)}) and RL (\emph{RL (MDP)}) learn a performant (high reward and waypoint percentage, low collision percentage) policy quickly and reliably.  In methods marked as POMDP the agent uses the high-dimensional monocular camera view.  Therefore, \emph{AIL} leads to a high collision, and the perception task makes RL in the POMDP (\emph{RL (POMDP)}) slow and variable (low reward and waypoint percentage, high collision percentage).  \emph{A2D} solves the scenario (high reward and waypoint percentage, low collision percentage) in a budget commensurate with the best-case convergence of \emph{RL (MDP)}.}
    \label{supp:fig:grid:a2dplot:results}
\end{figure*}

We implement this scenario in the CARLA simulator~\citep{Dosovitskiy17}, which is visualized in Figure \ref{supp:fig:carla:a2dplot:inputs}.  A child is present in 50\% of trials, and, if present, emerges with variable velocity.  The action space consists of the steering angle and amount of throttle/brake.  As an approximation to the optimal policy under privileged information, we used a hand-coded expert that completes the scenario driving at the speed limit if the child is absent, and slows down when approaching the truck if the child is present.  The differentiable expert is a small neural network, operating on a six-dimensional state vector that fully describes the simulator state.  The agent is a convolutional neural network that operates on grayscale images from the front-view camera. 

Results comparing A2D to four baselines are shown in Figure \ref{supp:fig:grid:a2dplot:results}.  \emph{RL (MDP)} uses RL to learn a policy conditioned on the omniscient compact state, only available in simulation, and hence does not yield a usable agent policy.  This represents the absolute best-case convergence for an RL method, achieving good, although not optimal, performance quickly and reliably.  \emph{RL} learns an agent conditioned on the camera image, yielding poor, high-variance results within the experimental budget. \emph{AIL} uses asymmetric DAgger to imitate the hand-coded expert using the camera image, learning quickly, but converging to a sub-optimal solution.  We also include \emph{OIL (MDP)}, which learns a policy conditioned on the omniscient state by imitating a hand-coded expert, and converges quickly to the near-optimal solution (\emph{MDP}). As expected, \emph{A2D} learns more slowly than AIL, since RL is used to update to the expert, but achieves higher reward than \emph{AIL} and avoids collisions.  This scenario, as well as any future asymmetric baselines, are distributed in the repository.

\section{Discussion}
\label{sec:discussion}

In this work we have discussed learning policies in POMDPs.  Partial information and high-dimensional observations can make direct application of RL expensive and unreliable.  Asymmetric learning uses additional information to improve performance beyond comparable symmetric methods.  Asymmetric IL can efficiently learn a partially observing policy by imitating an omniscient expert.  However, this approach requires a pre-existing expert, and, critically, assumes that the expert can provide suitable supervision -- a condition we formalize as identifiability.  The learned trainee can perform arbitrarily poorly when this is not satisfied.  We therefore develop adaptive asymmetric DAgger (A2D), which adapts the expert policy such that AIL can efficiently recover the optimal partially observed policy.  A2D also allows the expert to be learned online with the agent, and hence does not require any pretrained artifacts.  

There are three notable extensions of A2D.  The first extension is investigating more conservative updates for the expert and trainee which take into consideration the limitations or approximate nature of each intermediate update.  The second extension is studying the behavior of A2D in environments where the expert is not omniscient, but observes a superset of the environment relative to the agent. The final extension is integrating A2D into differentiable planning methods, exploiting the low dimensional state vector to learn a latent dynamics model, or, improve sample efficiency in sparse reward environments.

We conclude by outlining under what conditions the methods discussed in this paper may be most applicable.  If a pretrained expert or example trajectories are available, AIL provides an efficient methodology that should be investigated first, but, that may fail catastrophically.  If the observed dimension is small, and no reliable expert is available, direct application of RL is likely to perform well.  If the observed dimension is large, and trajectories which adequately cover the state-space are available, then pretraining an image encoder can provide a competitive and flexible approach.  Finally, if a compact state representation is available alongside a high dimensional observation space, A2D offers an alternative that is robust and expedites training in high-dimensional and asymmetric environments.

\balance

\section{Acknowledgements}
We thank Frederik Kunstner for invaluable discussions and reviewing preliminary drafts; and the reviewers for their feedback and improvements to the paper.  AW is supported by the Shilston Scholarship, University of Oxford.  JWL is supported by Mitacs grant IT16342. We acknowledge the support of the Natural Sciences and Engineering Research Council of Canada (NSERC), the Canada CIFAR AI Chairs Program, and the Intel Parallel Computing Centers program.  This material is based upon work supported by the United States Air Force Research Laboratory (AFRL) under the Defense Advanced Research Projects Agency (DARPA) Data Driven Discovery Models (D3M) program (Contract No. FA8750-19-2-0222) and Learning with Less Labels (LwLL) program (Contract No.FA8750-19-C-0515). Additional support was provided by UBC's Composites Research Network (CRN), Data Science Institute (DSI) and Support for Teams to Advance Interdisciplinary Research (STAIR) Grants. This research was enabled in part by technical support and computational resources provided by WestGrid (https://www.westgrid.ca/) and Compute Canada (www.computecanada.ca).

% The acknowledgments must go inside the 9-page limit, and references come afterwards.
\clearpage
\balance
\bibliographystyle{icml2021}
\bibliography{main}

% Do not touch anything below this.  On pain of death.  It is _so_ sensitive.
\clearpage

\appendix

\twocolumn[
\icmltitle{Supplementary Materials: Robust Asymmetric Learning in POMDPs}
\icmlsetsymbol{equal}{*}
\begin{icmlauthorlist}
\icmlauthor{Andrew Warrington}{equal,oxf}
\icmlauthor{J. Wilder Lavington}{equal,ubc,iai}
\icmlauthor{Adam \'Scibior}{ubc,iai}
\icmlauthor{Mark Schmidt}{ubc,ami}
\icmlauthor{Frank Wood}{ubc,iai,mil}
\end{icmlauthorlist}
\vskip 0.3in
]

\label{sec:supp}

\allowdisplaybreaks
\numberwithin{equation}{section}
\numberwithin{table}{section}
\numberwithin{figure}{section}

\section{Table of Notation}

\renewcommand{\arraystretch}{1.5}
\begin{minipage}{\textwidth}
\centering
% \begin{figure}[t]
\begin{adjustwidth}{-.5in}{-.5in}  
\begin{center}
\tiny
\begin{tabular}{@{}p{2cm}p{2cm}p{2cm}p{3cm}p{6.5cm}@{}}
\toprule
Symbol                             & Name                         & Alternative Name(s)                                             & Type                                                                                                                              & Description                                                                                                                                                                   \\ \hline
$t$                                & Time                         & Discrete time step                                              & $\mathbb{Z}$                                                                                                                      & Discrete time step used in integration.  Indexes other values.                                                                                                                \\
$s_t$                              & State                        & Full state, compact state, omniscient state                     & $\mathcal{S} = \mathbb{R}^D$                                                                                                      & State space of the MDP.  Sufficient to fully define state of the environment.                                                                                                 \\
$o_t$                              & Observation                  & Partial observation                                             & $\mathcal{O} = \mathbb{R}^{A\times B \times \dots}$                                                                               & Observed value in POMDP, emitted conditional on state.  State is generally not identifiable from observation. Conditionally dependent only on state.                          \\
$a_t$                              & Action                       &                                                                 & $\mathcal{A} = \mathbb{R}^K$                                                                                                      & Interaction made with the environment at time $t$.                                                                                                                            \\
$r_t$                              & Reward                       &                                                                 & $\mathcal{R}$                                                                                                                     & Value received at time $t$ indicating performance.  Maximising sum of rewards is the objective.                                                                               \\
$b_t$                              & Belief state                 &                                                                 & $\mathcal{B}$                                                                                                                     &                                                                                                                                                                               \\
$q_{\pi}$                          & Trajectory distribution      &                                                                 & $\mathcal{Q} : \Pi \rightarrow (\mathcal{A} \times \mathcal{B} \times \mathcal{O} \times \mathcal{S}^2 \times \mathcal{R})^{t+1}$ & Process of sampling trajectories using the policy $\pi$.  If the process is fully observed $\mathcal{O} = \emptyset$.                                                         \\
$\tau_{0:t}$                       & Trajectory                   & Rollouts                                                        & $(\mathcal{A} \times \mathcal{B} \times \mathcal{O} \times \mathcal{S}^2 \times \mathcal{R})^{t+1}$                               & Sequence of tuples containing state, next state, observation, action and reward.                                                                                              \\
$\gamma$                           & Discount factor              &                                                                 & $\Gamma = \left[  0, 1 \right]$                                                                                                   & Factor attenuating future reward in favor of near reward.                                                                                                                    \\
$p(s_{t+1} | s_t, a_t)$            & Transition distribution      & Plant model, environment                                        & $\mathcal{T} : \mathcal{S} \times \mathcal{A} \rightarrow \mathcal{S}$                                                            & Defines how state evolves, conditional on the previous state and the action taken.                                                                                            \\
$p(o_t | s_t)$                     & Emission distribution        & Observation function                                            & $\mathcal{Y} : \mathcal{S} \rightarrow \mathcal{O}$                                                                               & Distribution over observed values conditioned on state.                                                                                                                       \\
$p(s_0)$                           & Initial state distribution   & State prior                                                     & $\mathcal{T}_0 : \rightarrow \mathcal{S}$                                                                                         & Distribution over state at $t=0$.                                                                                                                                             \\
$\pi_{\theta}(a_t | s_t)$          & MDP policy                   & Expert, omniscient policy, asymmetric expert, asymmetric policy & $\Pi_{\Theta} : \mathcal{S} \rightarrow \mathcal{A}$                                                                              & Distribution over actions conditioned on state.  Only used in MDP.                                                                                                            \\
$\theta$                           & MDP policy parameters        &                                                                 & $\Theta$                                                                                                                          & Parameters of MDP policy.  Cumulative reward is maximized over these parameters.                                                                                              \\
$\pi_{\phi}(a_t | b_t)$            & POMDP policy                 & Agent, partially observing policy                               & $\Pi_{\Phi} : \mathcal{B} \rightarrow \mathcal{A}$                                                                                & Distribution over actions conditioned on belief state.  Only used in POMDP.                                                                                                   \\
$\phi$                             & POMDP policy parameters      &                                                                 & $\Phi$                                                                                                                            & Parameters of MDP policy.  Cumulative reward is maximized over these parameters.                                                                                              \\
$\pi_{\psi}(a_t | b_t)$            & Variational trainee policy                 & Variational approximation                               & $\Pi_{\Psi} : \mathcal{B} \rightarrow \mathcal{A}$                                                                                & Variational approximation of the implicit policy.                                                                                                   \\
$\psi$                             & Variational trainee policy parameters      &                                                                 & $\Psi$                                                                                                                            & Parameters of the variational approximation of the implicit policy.                                                                                              \\
$\pi_{\beta}$                      & Mixture policy               &                                                                 & $\Pi_{\beta} : \mathcal{S} \times \mathcal{B} \rightarrow \mathcal{A}$                                                            & Mixture of MDP policy ($\pi_{\theta}$) and POMDP policy ($\pi_{\phi}$).                                                                                                       \\
$\beta$                            & Mixing coefficient           &                                                                 & $\left[ 0, 1 \right]$                                                                                                             & Fraction of MDP policy used in mixture policy.                                                                                                                                \\
$D$                                & Replay buffer                & Data buffer                                                     & $\mathcal{D} = \left\lbrace \tau_{0:T_n} \right\rbrace_{n \in 1:N}$                                                               & Store to access previous trajectories.  Facilitates data re-use.                                                                                                              \\
$\mathbb{KL}\left[ p || q \right]$ & Kullback–Leibler divergence  & KL divergence, forward KL, mass-covering KL                     &                                                                                                                                   & Particular divergence between two distributions.  Forward KL is mass covering.  Reverse KL ($\mathbb{KL} \left[ q || p \right]$) is mode seeking.                             \\
$Q^{\pi}(s_t, a_t)$                & Q-function                   & State Q-function                                                & $\mathcal{Q}_s : \mathcal{S} \times \mathcal{A} \rightarrow \mathbb{R}$                                                           & Expected sum of rewards ahead, garnered by taking action $a_t$ in state $s_t$ induced by policy $\pi$.                                                                        \\
$Q^{\pi}(b_t, a_t)$                & Belief state Q-function      &                                                                 & $\mathcal{Q}_b : \mathcal{B} \times \mathcal{A} \rightarrow \mathbb{R}$                                                           & Expected sum of rewards ahead, garnered by taking action $a_t$ in belief state $b_t$ induced by policy $\pi$.                                                                 \\
$\hat{\pi}_{\theta}(a_t | b_t)$    & Implicit policy &                                                                 & $\Pi_{\Phi} : \mathcal{B} \rightarrow \mathcal{A}$                                                                                & Agent policy obtained by marginalizing over state given belief state.   Closest approximation of $\pi_{\theta}$ under partial observability.  Approximated by $\pi_{\phi}$.   \\
% $\hat{Q}^{\pi_{\phi}}(s_t, a_t)$   & Belief-marginal Q-function   &                                                                 & $\mathcal{Q}_s : \mathcal{S} \times \mathcal{A} \rightarrow \mathbb{R}$                                                           & Q-function, conditioned on state, obtained by rolling out under belief-state conditioned policy $\pi_{\phi}$.                                                                 \\
$d^{\pi}(s_t, b_t)$                & Occupancy                    & Discounted state visitation distribution~\citep{pmlr-v125-agarwal20a}                                                                & $M : \mathcal{S} \times \mathcal{B} \rightarrow \mathbb{R}$                                                                     & Joint density of $s_t = s$ and $b_t = b$ given policy $\pi$. Marginal of $q_{\pi}$ over previous and future states, belief states, and all actions, observations and rewards. \\
$\pi_{\eta}$                & Fixed reference distribution                    &                                                                 & $\Pi$                                                                     & Fixed distribution that is rolled out under to generate samples that are used in gradient calculation. \\\hline
\end{tabular}
\end{center}
\end{adjustwidth}
\captionof{table}{Notation and definitions used throughout the main paper.}
\label{tab:notation}
% \end{figure}
\end{minipage}

\onecolumn 
\clearpage

\section{Additional Experimental Results}

\printAffiliationsAndNotice{\icmlEqualContribution} % otherwise use the standard text.

\label{supp:exp}

\subsection{Estimating the Q Function}
\label{supp:sub:sub:q}
In Section \ref{sec:algorithm} we briefly discussed the possibility of avoiding explicitly estimating the Q function.  All the terms in \eqref{equ:expert-gradient} can be computed directly, with the exception of the Q function.  One approach therefore is to train an additional function approximator targeting the Q function directly.  This can then be used to estimate the discounted sum of rewards ahead given a particular action and belief state (when $\beta = 0$) without directly using the Monte Carlo rollouts.  However, estimating the Q function increases the computational cost, increases the number of hyperparameters that need tuning, and can lead to instabilities and biased training by over reliance on imperfect function approximators, especially in high-dimensional environments.  Therefore, as in many on-policy RL algorithms, an alternative is to use Monte Carlo estimates of the Q function, computed directly from a sampled trajectory (c.f. \eqref{equ:a2d:a2d_update}-\eqref{equ:a2d:gae}).  

However, somewhat unexpectedly, this second approach can lead to the systemic failure of A2D in particular \emph{environments}.  This can be shown by expanding the definition of $Q^{{\pi_{\psi}}}(a,b)$:
\begin{align}
     Q^{{\pi_{\psi}}}(a,b) = \mathop{\mathbb{E}}_{p(s, s'|a,b)} \Bigg[ \mathop{\mathbb{E}}_{d^{{\pi_{\psi}}}(b'|s')} \Big[ r(s,a,s') + \mathop{\gamma\mathbb{E}}_{\pi_{\psi}(a'|b')} \left[ Q^{\pi_{\psi}} (a', b') \right] \Big] \Bigg], \label{supp:equ:q_loss}
\end{align}
where $s'$ and $b'$ are the state and belief state after taking action $a$ in state $s$ and belief state $b$.  Since sampling from $p(s, s'|a,b)$ and $d^{{\pi_{\psi}}}(b' | s')$ is intractable, directly using the trajectories is equivalent to using a singled sample value throughout this expression \emph{and} the gradient estimator in \eqref{equ:a2d:a2d_update}.  Re-using just a single value of $s$ inside and outside of this expectation biases the gradient estimator, as the estimate of $Q$ is \emph{not} conditionally independent of the current (unobserved) state given the belief state.  Intuitively, using Monte Carlo rollouts  essentially allows the expert to ``cheat'' by learning using exclusively the true state and reward signal over a \emph{single} time step of a trajectory.  

When the Q function is estimated directly, the expectation in \eqref{supp:equ:q_loss} is estimated directly by the learned Q function, thereby amortizing this inference by learning across many different sampled trajectories.  Therefore, from a theoretical perspective, estimating the Q function is important for A2D to be guaranteed to function.  However, we find that this bias is only significant in specific environments, and hence, in many environments, explicitly estimating the Q function can be avoided.  This reduces the computational cost of the algorithm, and reduces the number of hyperparameters and network architectures that need tuning.  Furthermore, and most importantly, this eliminates the direct dependence on faithfully approximating the Q function, which, in environments with high-dimensional observations and actions, can be prohibitively difficult.  

\begin{figure}[t]

    \begin{subfigure}[t]{0.3\textwidth}
        \includegraphics[width=\textwidth]{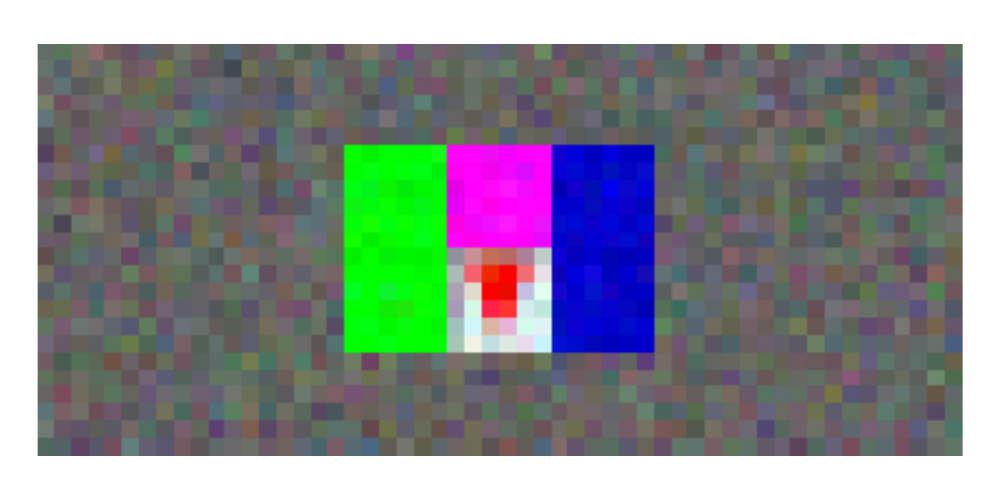}
        \includegraphics[width=\textwidth]{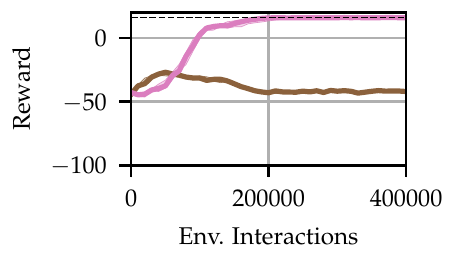}
        \includegraphics[width=\textwidth]{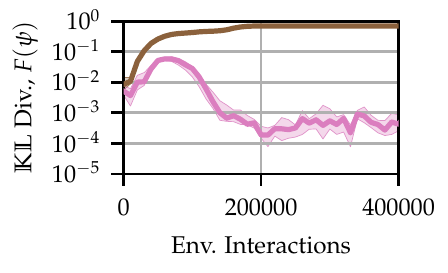}
        \caption{Tiger Door 1.}
        \label{supp:fig:grid:OSC:td1}
    \end{subfigure}%
    \hfill%
    \begin{subfigure}[t]{0.3\textwidth}
        \includegraphics[width=\textwidth]{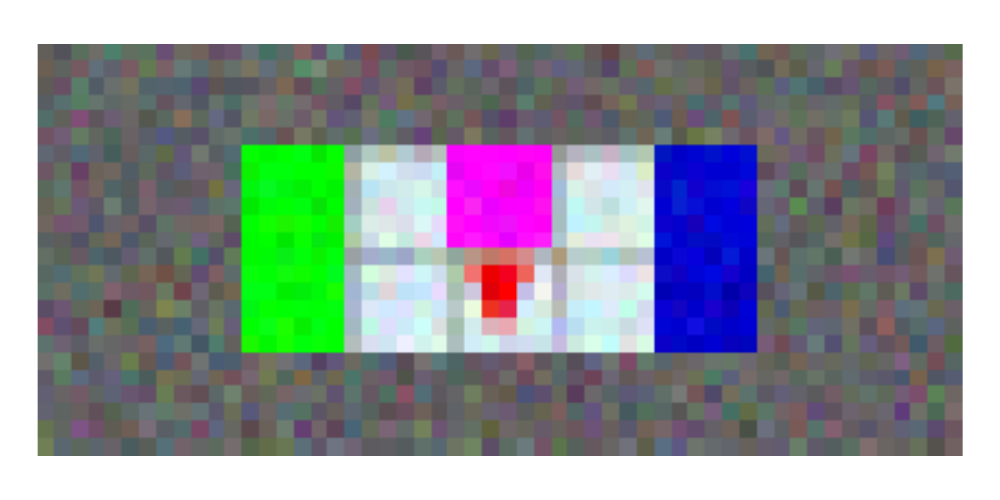}
        \includegraphics[width=\textwidth]{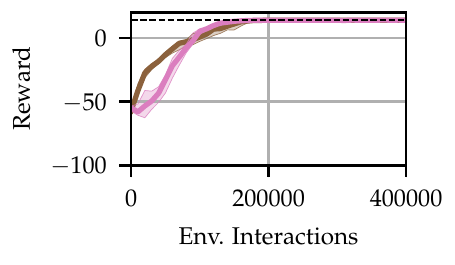}
        \includegraphics[width=\textwidth]{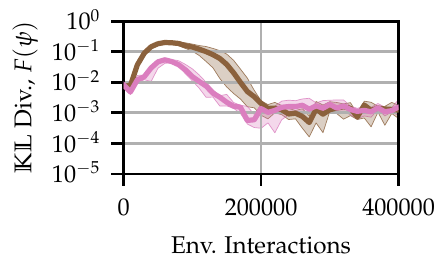}
        \caption{Tiger Door 2.}
        \label{supp:fig:grid:OSC:td2}
    \end{subfigure}%
    \hfill%
    \begin{subfigure}[t]{0.3\textwidth}
        \includegraphics[width=\textwidth]{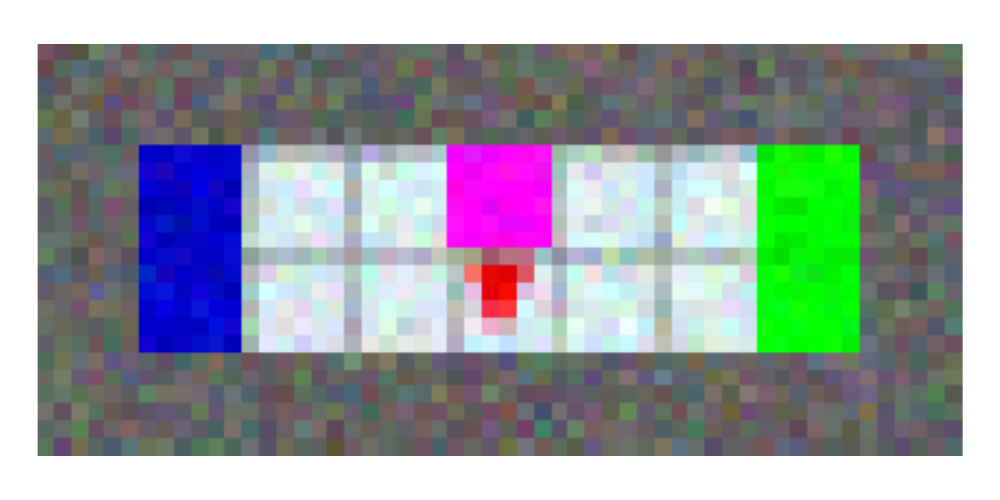}
        \includegraphics[width=\textwidth]{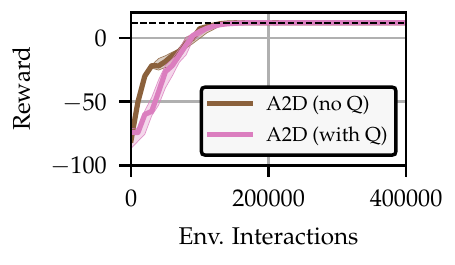}
        \includegraphics[width=\textwidth]{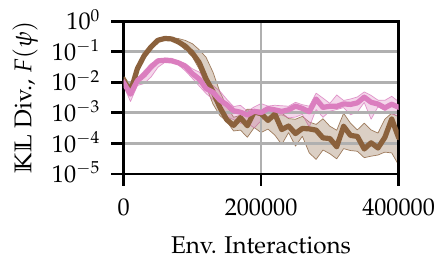}
        \caption{Tiger Door 3.}
        \label{supp:fig:grid:OSC:td3}
    \end{subfigure}%

    \caption{Results investigating requirement of directly estimating the Q function, as initially introduced in Section \ref{sec:algorithm} and discussed further in Section \ref{supp:sub:sub:q}.  Median and quartiles across $20$ random seeds are shown.  The Q function is learned targeting the expected discounted sum of rewards ahead conditioned on a particular (belief) state-action pair.  A value function is also learned in this way, and is used in conjunction with the Q function to directly estimate the advantage in \eqref{equ:a2d:a2d_update}.  Hence the A2D gradient is computed without direct use of Monte Carlo rollouts.  When no Q function is being used, the advantage is computed using GAE (c.f. Equations \eqref{equ:a2d:a2d_update}-\eqref{equ:a2d:gae}), with $\lambda = 0.5$.  We instantly anneal $\beta = 0$.  { }%  
    Figure \ref{supp:fig:grid:OSC:td1}: Training curves for Tiger Door 1~\citep{littman1995pomdp}.  As predicted by the discussion in Section \ref{supp:sub:sub:q}, A2D does not converge to the correct policy if a Q function is not simultaneously learned.  This deficiency is instrumented by the high $\mathbb{KL}$ divergence throughout training and a discrepancy between the expected reward of the expert and agent.  If a Q function is learned, the desired partially observed behavior is recovered.   { }%
    Figure \ref{supp:fig:grid:OSC:td2} and \ref{supp:fig:grid:OSC:td3}: By separating the goal by at least one square means the desired behavior is recovered regardless of whether a Q function is used.  This is because the bias has been reduced through the use of GAE and the introduction of additional random variables.  { }%
    }
    \label{supp:fig:grid:OSC}
\end{figure}

To explore this behavior, and verify this theoretical insight, we introduce three variants of the Tiger Door problem, shown in Figure \ref{supp:fig:grid:OSC}.  The first variant, ``Tiger Door 1,'' shown in Figure \ref{supp:fig:grid:OSC:td1}, actually corresponds to a gridworld embedding of the original Tiger Door problem~\citep{littman1995pomdp}.  ``Tiger Door 2'' \& ``Tiger Door 3,'' shown in Figures \ref{supp:fig:grid:OSC:td2} and \ref{supp:fig:grid:OSC:td3}, then separate the goal by one and two squares respectively.  

The analysis above predicts that A2D should not be able to solve Tiger Door 1 without direct estimation of the Q function.  This is because the expert can reach the goal with certainty in a single action, which ends the episode.  This means the expert can always maximize reward by proceeding directly to the goal, and as the episode ends, the gradient signal is dominated by the bias from the single step.  This causes the expert to put additional mass on directly proceeding to the goal, even though the goal is not visible to the agent.  We note that this is also the most extreme example of this bias, and we believe this environment to be somewhat of an unusual corner-case.  

However, in Tiger Doors 2 and 3, the episode does not end immediately after proceeding directly towards the goal.  Therefore, the value of proceeding directly towards the goal is diminished, as the marginalization over state provided by GAE and the value function reduces the estimated advantage value.  The gradient computed in these scenarios is therefore dramatically less biased, to the point where directly estimating the Q function not required.  

The predicted behavior is indeed observed when applying A2D to each Tiger Door variant, shown in Figure \ref{supp:fig:grid:OSC}.  We see that in Tiger Door 1, the correct policy is only recovered when the Q function is explicitly estimated.  When the Q function is not estimated, the expert directly optimizes just the reward under the MDP, earning itself a reward of $18$, but rendering an implicit policy that performs poorly, earning a reward of $-42$.  In Tiger Doors 2 \& 3, the correct trainee policy is recovered regardless of whether a Q function is explicitly learned.  Interestingly, we observe that the policy divergence, $F(\psi)$, is often lower during training when using the Q function.  This further reinforces that estimating the Q function more directly optimizes the reward of the trainee.  We note, however, that the \emph{final} divergence achieved by the Q function is often higher than that obtained without Q. This is likely due to the systemic bias introduced by using function approximation.  Note that for all of these experiments we use the compact representation.  

\begin{figure}[t]

    \begin{subfigure}[t]{0.48\textwidth}
        \includegraphics[width=0.95\textwidth]{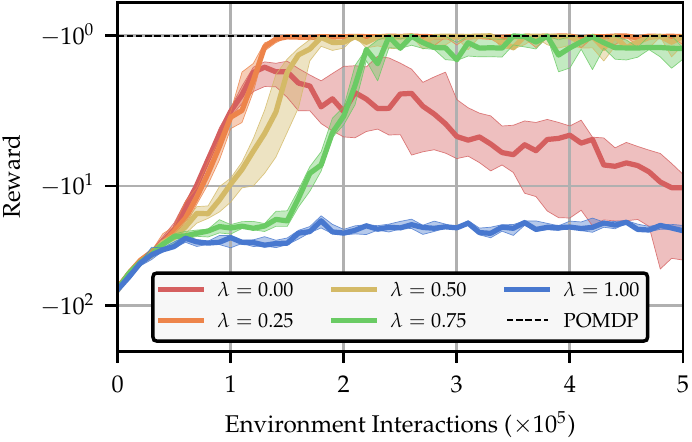}
        \caption{Reward.}
        \label{fig:osc:lambda:reward}
    \end{subfigure}%
    \hfill%
    \begin{subfigure}[t]{0.48\textwidth}
        \includegraphics[width=0.95\textwidth]{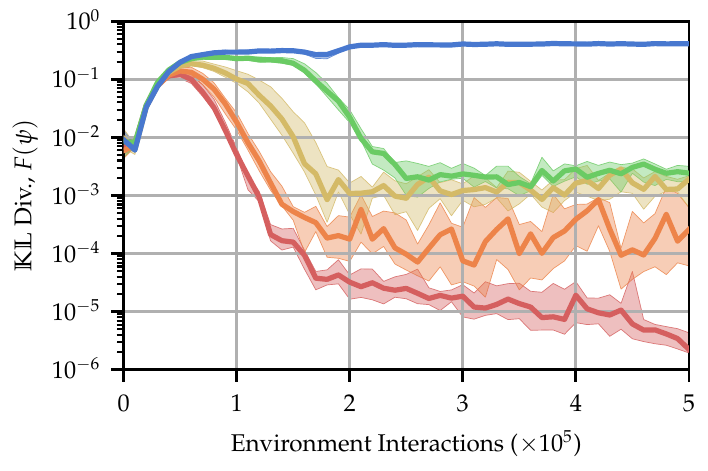}
        \caption{Divergence.}
        \label{fig:osc:lambda:divergence}
    \end{subfigure}%
    \vspace{-0.25cm}
    \caption{Results showing the affect of the GAE parameter $\lambda$ on A2D, applied to the  Tiger Door 2 environment.  The reward is normalized such that the optimal reward under the POMDP is $-10^{0}$.  As predicted, we see that lower $\lambda$ values yield faster convergence and monotonically lower policy divergences.  However, as this is equivalent to TD0, the RL is unstable (obscured in this plot are short, sharp drops in the reward and rises in the divergence).  Eventually, all traces begin to diverge from the optimal policy.  For any $\lambda$ value less than unity, convergence is stable (and the short, sharp drops do not exist).  Finally, and again as predicted, we see that learning does not converge when $\lambda = 1$, with reward remaining flat and low, and the divergence remaining high. { }%
    }
    \label{fig:osc:lambda}
\end{figure}

We also explore, in Figure \ref{fig:osc:lambda}, the affect that the GAE parameter~\citep{schulman2015high}, $\lambda$, has on A2D training.  Inspecting \eqref{equ:a2d:gae} indicates that GAE provides the ability to diminish the unmodelled dependence on $s_t$, and hence reduce the bias in the estimator by attenuating future reward from the Monte Carlo rollouts and replacing this reward with the correctly amortized value, integrating over the true state, estimated by the value function (which in the limit of $\beta = 0$ is only conditioned on $b_t$).   This suggests that $\lambda=0$, corresponding to the expected temporal difference reward, is as close to the theoretically ideal Q function based estimator in \eqref{equ:expert-gradient} as is possible.  The dependency on $s_t$ (as denoted in \eqref{equ:a2d:gae}) is maximally reduced, to the point where it only affects the gradient signal for a single step (further reinforcing why Tiger Door 1 fails, but Tiger Doors 2 \& 3 succeed).  In contrast, using $\lambda = 1$ maximizes the bias, by not attenuating any Monte Carlo reward signal. 

We observe this behavior in Figure \ref{fig:osc:lambda}.  We see that $\lambda=1$ does not converge to the optimal solution, as the bias term dominates, effectively halting learning.  Lower $\lambda$ values allow more state information to be integrated out, and hence the partially observed policy can improve.  This is seen by faster and more stable convergence for lower $\lambda$ values.  This can also be observed in Figure \ref{fig:osc:lambda:divergence}, where lower $\lambda$ values achieve a lower policy divergence.  This implies that the expert is less able to leverage state information to learn a policy that cannot be represented by the agent.  However, reducing $\lambda$ must unfortunately be balanced against the limiting behavior of GAE, which corresponds to a TD0 estimate of the return.  In this regime, bias introduced by function approximation can make the convergence of RL unreliable.  Indeed, even in idealized settings, it can be shown such estimators diverge without further modifications, such parameter averaging~\citep{maei2009convergent}.

Therefore, the hyperparameter $\lambda$ takes on additional importance when tuning A2D using the biased Monte Carlo gradient estimator.  If the coefficient $\lambda$ is too close to zero, then the effects of bootstrapping error can lead learning to stall, unstable solutions, or even divergence, as is often observed in RL, and may reduce the effectiveness of GAE and RL by overly relying on function approximators.  However, this lower $\lambda$ value reduces the bias in the estimator, and hence provides faster convergence, more stable convergence, and achieves a lower final policy divergence (c.f. $\lambda = 0.00$ in Figure \ref{fig:osc:lambda}).  If $\lambda$ is too close to unity, there may too much bias in the gradient estimate.  This bias may force, either, A2D to not converge outright if $\lambda = 1.00$, or, cause A2D to drift away slightly from the optimal solution after convergence as the expert aggregates this slight bias into the solution.  In practice, we find that this second failure mode only occurs once learning has already converged to the optimal solution, and so the optimal policy can simply be taken prior to any divergence.  Further analysis of this effect, both theoretically, such as defining and bounding the precise nature of this bias, and practically, such as adaptively adjusting $\lambda$ to control the bias-variance trade-off, are interesting directions of future work.  

We note that Frozen Lake subtly exhibits the bias in the Monte Carlo gradient estimator if the value of $\lambda$ is too high.  First, the trainee quickly converges to the optimal partially observing policy (and so the environment \emph{is} solved).   Then, after \emph{many} more optimization steps, the probability that the agent steps onto the ice can rise slightly.  This causes the divergence to rise slightly and the expected stochastic reward to fall slightly before stabilizing.  The rise is small enough that the deterministic policy evaluation remains unchanged.  However, as predicted by the analysis above, this behavior can be eradicated by reducing the value of $\lambda$.  However, in RL generally, lowering $\lambda$ can stall, or even halt, learning from the outset by overreliance on biased function approximators.  This can cause the optimization to become stuck in local minima.  As eluded to above, we find that we can control this behavior by slightly reducing the value of $\lambda$ from its initial value during the optimization.  This minimizes the dependency on function approximators early in training and retains the fast and reliable convergence to the optimal policy, and then attenuates any bias after learning has converged.  While we believe this behavior only presents in a small number of very specific environments and can be eradicated through tuning of hyperparameters, we propose that further investigation of adaptively controlling $\lambda$ during the optimization is a promising and practical future research directions.  More generally, investigating methods for quantifying and ameliorating this bias is an exciting topic of future research.  Note that we did not apply this annealing in the experiments presented in the main text.  When the Q function is estimated, this behavior is not observed, and the reward remains optimal and the divergence remains low.  

It is also paramount to highlight here that A2D, at its core, is underpinned by an RL step, and more specifically, a policy gradient step, and hence all the considerations when designing, applying and tuning a regular RL algorithm still apply in A2D. In fact, A2D can, in many respects, be considered a special class of projected policy gradient methods.  Specifically, we optimize \emph{through} the projection defined by AIL, guaranteeing the desired behavior through subsequent imitation. Therefore, in this respect, it is important to reinforce that A2D does not provide a ``free lunch,'' and hyperparameters are still important and can have direct affects on the performance of A2D -- even if, in practice, we find that A2D works well when using many of the same hyperparameters as RL in the underlying MDP.  

The strong dependency on the Q function leads us to recommend that ``default'' A2D algorithm is to not directly estimate the Q function, and instead estimate the advantage directly from the Monte Carlo trajectories, and the bias is tolerated, or $\lambda$ adjusted accordingly.  There are then two instruments available to diagnose if a Q function must be directly approximated: if there is consistently a performance gap between the expert policy and the agent policy, or, if there is a non-negligible $\mathbb{KL}$ divergence between the expert and agent policy (indicating the policies are not being forced to be identifiable).  If either of these behaviors are observed, then a Q function should be directly estimated.  

While the discussion and example presented in this example provide some explanation of this behavior, we were unable to provide a concrete definition, condition, or test identifying when direct estimation of the Q function is required, or, precise mathematical quantification of how $\lambda$ influences the bias.  Crucially, the core of this behavior is a function of the \emph{environment}, and hence there may be no readily available or easy-to-test condition for when a Q function is required.  Beyond this, this effect may manifest as a complication in any method for ameliorating the drawbacks of AIL, and hence further investigation of this is a challenging, interesting, and potentially pivotal theoretical topic for future research, studying the very nature of MDP and POMDPs.  Beyond this, building further intuition, understanding, and eventually defining, the relative influence of different hyperparameter settings in A2D, particularly between when estimating Q and not estimating Q, is a future research direction with great practical benefits.

\begin{figure*}[t]

    \begin{subfigure}[t]{0.42\textwidth}
        \includegraphics[width=0.95\textwidth]{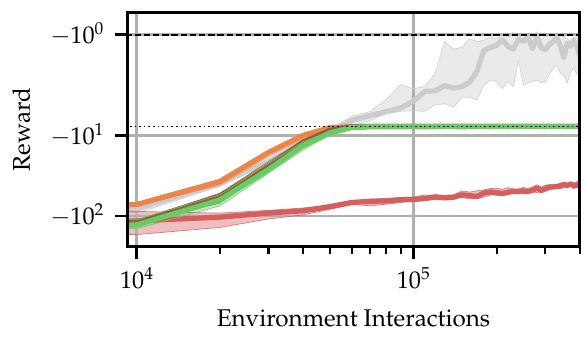}
        \caption{Frozen Lake.}
        \label{supp:fig:grid:a2dplot_2:lg}
    \end{subfigure}%
    \hfill%
    \begin{subfigure}[t]{0.42\textwidth}
        \includegraphics[width=0.95\textwidth]{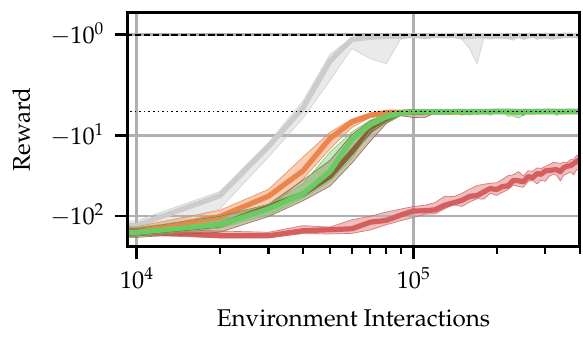}
        \caption{Tiger Door.}
        \label{supp:fig:grid:a2dplot_2:td}
    \end{subfigure}%%
    \hfill%
    \begin{subfigure}[t]{0.15\textwidth}
        \includegraphics[width=0.95\textwidth]{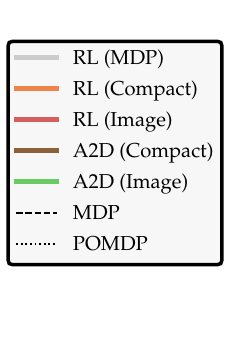}
    \end{subfigure}%

    \caption{Training curves comparing convergence of A2D and vanilla RL on the POMDP for compact (one-hot vector) representations and image-based representations.  We see that RL on the compact partial representation (orange) converges to the optimal POMDP reward (horizontal dotted line, $-9 \times 10^0$ and $-7 \times 10^0$) quickly, and in a sample complexity similar to the best-case convergence of RL in the MDP (gray), which converges to the optimal MDP reward (horizontal dashed line, $- 1 \times 10^0$), when both use hyperparameters comparable to as were used with A2D.  In contrast, RL on the images (red) converges slowly, and does not reach the optimal POMDP reward within the allocated computational budget.  A2D on the other hand converges to the optimal POMDP reward in a sample complexity commensurate with RL operating directly on the compact representation, for \emph{both} image-based and compact representations (brown and green respectively).  This confirms our hypothesis that A2D can reduce the complexity of operating in high-dimensional, partially observed environments to a complexity commensurate with the best-possible convergence rate obtained by performing RL directly on the most efficient encoding or complete state.  { }
    }
    \label{supp:fig:grid:a2dplot_2}
\end{figure*}

\subsection{Differences in Representation}
In Figure \ref{supp:fig:grid:a2dplot_2} we investigate A2D when the trainee uses different representations.  Specifically, we investigate using a compact-but-partial vector representation (labeled as \emph{Compact}), and the original image-based representation (labeled as \emph{Image}).  Both representations include the same partial information, but the compact representation is a much more efficient representation for RL.  The compact representation for Frozen Lake is a length $25$ one-hot vector representing the position of the agent.  For Tiger Door the compact representation is the concatenation of three one-vectors: a length $25$ one-hot vector encoding the position of the agent, a length two vector encoding the position of the goal, and a length two vector encoding the position of the hazard.  The goal and hazard vectors are all zeros until the button is pressed, at which time they become one-hot vectors.  This can be considered as the optimal encoding of the observation and action history.  We note that analytically recovering such an encoding is not always possible (in the AV example, for instance), and learning an encoding (c.f. \emph{Pre-Enc} in Figure \ref{fig:gridworld_asym}) is unreliable, and introduces a non-trivial amount of additional complexity and hyperparameter tuning.  

Results are shown in Figure \ref{supp:fig:grid:a2dplot_2}.  We see performing RL directly on the compact representation (\emph{RL (Compact)}) is fast and stable.  Direct RL in the image-based representation (\emph{RL (Image)}) is slow, and does not converge within the computational budget.  For both Frozen Lake and Tiger Door, A2D converges in a similar number of interactions for both image-based inputs (\emph{A2D (Image)}) and the compact representation (\emph{A2D (Compact)}), and that is commensurate with the convergence of an omniscient MDP expert and RL in the compact state, when using the A2D hyperparameters.  This shows that A2D has successfully abstracted the perception task into the efficient AIL step, and performs RL in the efficient and low-variance omniscient state representation in the best-case sample complexity for those hyperparameters.  This means that A2D is able to exploit the relative strengths of RL to offset the weaknesses of AIL, and vice versa, in an efficient, low-overhead and end-to-end manner.  Crucially, the expert is co-trained with the trainee, and hence there is no requirement for pre-specified expert policies or example trajectories from which to learn policies or static encoders.

\newpage

\clearpage

\section{Additional Proofs}
\label{supp:thoery}

\newcommand{\klbars}{\ ||\ }
\newcommand{\lspsqb }{\left[ \hspace*{0.1em} }
\newcommand{\rspsqb}{\hspace*{0.1em} \right]}

In this section we provide full proofs for the material presented in the main text.  These proofs describe more completely how the A2D estimator is constructed.  We briefly give an overview of how the following proofs and details are laid out.

We begin in Section \ref{supp:sec:occupancy} by discussing in more detail the occupancy $d^{\pi}(s,b)$.  This joint occupancy is a convenient term to define as it allows us to compactly denote the probability that an agent is in a particular state and belief state at any point in time.  We can then construct conditional and marginal occupancies by operating on this joint occupancy.  

In Section \ref{supp:sec:ail} we analyze the behavior of AIL.  We first detail a full proof of Theorem \ref{supp:theorem:1_ail_target}, stating that the implicit policy is the solution to the minimization of the most conveniently defined AIL objective, where the trainee simply imitates the expert at each state-belief state pair.  This allows us to compactly write and analyze the solution to AIL as the implicit policy.  However, the implicit policy is defined by an intractable inference over the conditional occupancy, $d^{\pi}(s \mid b)$, from which we cannot sample.  

We therefore show in Section \ref{supp:sec:vap} that we can define a variational approximation to the implicit policy, referred to as a \emph{trainee}, that is learned using the AIL objective.  We construct an estimator of the gradient of the trainee parameters to learn this trainee, under a fixed distribution over trajectories, directly targeting the result of the inference defined by the implicit policy.  Crucially, the trainee can be learned using samples from the joint occupancy, $d^{\pi}(s,b)$, from which we \emph{can} sample (instead of samples from the conditional $d^{\pi}( s \mid b )$ as per the implicit policy).  If the variational family is sufficiently expressive, this minimization can be performed exactly. 

We then show that an iterative AIL approach, that updates the fixed distribution over trajectories at each iteration, recovers the desired trainee.  We then show that the limiting behavior of this iterative algorithm is equivalent to learning under the occupancy of the implicit policy.  Finally, using these results, we prove Theorem \ref{supp:def:fixed_point_variational}, which shows that for an identifiable MDP-POMDP pair, the iterative AIL approach outlined above recovers an optimal partially observing policy.  

However, identifiability is a \emph{very} strong condition.  Therefore, mitigating unidentifiability in AIL is primary the motivation behind A2D.  In Section \ref{supp:sec:a2d} we provide a proof of the ``exact'' form of A2D.  We begin by providing additional detail on intermediate results, including a brief explanation of the policy bound stated in Equations \eqref{equ:bound:1}-\eqref{equ:bound:2}, a derivation of the Q-based A2D update in Equation \eqref{equ:expert-gradient}, and the advantage-based update in \eqref{equ:a2d:a2d_update}.  We then use the assumptions, intermediate lemmas, and theorems to prove exact A2D (using a similar strategy as we used to prove Theorem \ref{supp:def:fixed_point_variational}).  This verifies that, under exact updates, A2D converges to the optimal partially observing policy.  We then conclude by evaluating the requirements of this algorithm.

\subsection{Occupancy Measures}\label{supp:sec:occupancy}
Throughout this paper we use $q_{\pi}(\tau)$ as general notation for the trajectory generation process, indicating which policy is used to generate the trajectory as a subscript (c.f. \eqref{equ:background:mdp} and \eqref{equ:background:pomdp_dist}).  We define the \emph{joint occupancy}, $d^{\pi}(s, b)$, as the time-marginal of $q_{\pi}(\tau)$ over all variables in the trajectory other than $s$ and $b$:
\begin{align}
    d^{\pi}(s,b) &= (1-\gamma) \int_{\tau \in \mathcal{T}} \sum_{t=0}^{\infty} \gamma^t q_{\pi} (\tau) \delta (s_t - s) \delta (b_t - b) \d \tau , \where \gamma \in [0, 1), \\
    d^{\pi}(s) &= \int_{b' \in \mathcal{B}} d^{\pi}(s, b') \d b',  \quad d^{\pi}(s | b) = \int_{b' \in \mathcal{B}} d^{\pi}(s, b') \delta ( b' - b ) \d b' , \\
    d^{\pi}(b) &= \int_{s \in \mathcal{S}} d^{\pi}(s', b) \d s' , \quad d^{\pi}(b | s) = \int_{s' \in \mathcal{S}} d^{\pi}(s', b) \delta ( s' - s ) \d s'.
\end{align} 
We refer the reader to \S 3 of \citet{pmlr-v125-agarwal20a} for more discussion on the occupancy (described instead as a \emph{discounted state visitation distribution}).  Despite the complex form of these expressions, we can sample from the joint occupancy $d^{\pi}(s, b)$ by simply rolling out under the policy $\pi$ according to $q_{\pi}(\tau)$, and taking a random state-belief state pair from the trajectory.  We can then trivially obtain a sample from either marginal occupancy, $d^{\pi}(s)$ or $d^{\pi}(b)$, by simply dropping the other variable.  We can also recover a \emph{single} sample, for a \emph{sampled} $b$, from the conditional occupancy $d^{\pi}(s \mid b)$ by taking the associated $s$ (and vice-versa for conditioning on a sampled $s$).  However, and critically for this work, sampling multiple states or belief states from either conditional occupancy is intractable.  Therefore, much of the technical work presented is carefully constructing and manipulating the learning task such that we can use samples from the joint occupancy (from which we can sample), in-place of samples from the conditional occupancy (from which we cannot sample).

\newpage

\subsection{Analysis of AIL}
\label{supp:sec:ail}

We begin by analyzing the behavior of AIL.  This will allow us to subsequently define the behavior of A2D by building on these results.

\subsubsection{Proof of Theorem \ref{supp:theorem:1_ail_target}}
\label{supp:sec:t1}

We first verify the claim that the implicit policy minimizes the AIL objective.

\setcounter{sthe}{\value{theorem}}
\setcounter{theorem}{0}                   % CHANGE THE COUNTER VALUES IN HERE.
\begin{theorem}[Asymmetric IL Target, reproduced from Section \ref{sec:prelim}]\label{supp:theorem:1_ail_target}
For any fully observing policy $\pi_{\theta} \in \Pi_{\Theta}$ and fixed policy $\pi_{\eta}$, the implicit policy $\hat{\pi}_{\theta}^{\eta} \in \hat{\Pi}_{\Theta}$, defined in Definition 1, minimizes the following asymmetric IL objective:
\begin{equation}
      \hat{\pi}_{\theta}^{\eta}(a|b) = \mathop{\argmin}_{\pi \in \Pi_{\Phi}}\ \mathbb{E}_{d^{\pi_{\eta}}(s,b)}  \lspsqb{}  \mathbb{KL} \lspsqb{}  \pi_{\theta}(a|s) \klbars{} \pi(a|b) \rspsqb{}  \rspsqb{} .
\end{equation}
\end{theorem}
\setcounter{theorem}{\value{sthe}}
\begin{proof}
Considering first the optima of the right-hand side: 
\begin{equation}
    \pi^*(a|b) = \mathop{\argmin}_{\pi \in \Pi}\ \mathbb{E}_{d^{\pi_{\eta}}(s, b)}  \lspsqb{}  \mathbb{KL} \lspsqb{}  \pi_{\theta}(a|s) \klbars{} \pi(a|b) \rspsqb{}  \rspsqb{} ,
\end{equation}
and expanding the expectation and $\mathbb{KL}$ term:
\begin{align}
    \pi^*(a|b) &= \mathop{\argmin}_{\pi \in \Pi} \mathbb{E}_{d^{\pi_{\eta}}(b)}\lspsqb{}  \int_{s\in\mathcal{S}} \int_{a\in\mathcal{A}} \pi_{\theta}(a | s)  \log \left( \frac{\pi_{\theta}(a|s)}{\pi(a | b)} \right) \d a \ d^{\pi_{\theta}}(s | b) \d s \rspsqb{}  , \\%
    &= \mathop{\argmin}_{\pi \in \Pi} \mathbb{E}_{d^{\pi_{\eta}}(b)}\lspsqb{} \int_{s\in\mathcal{S}} \int_{a\in\mathcal{A}} \pi_{\theta}(a | s)  \log \pi_{\theta}(a|s) \d a \ d^{\pi_{\eta}}(s | b) \d s \rspsqb{}  - \\
    & \hspace*{1.8cm} \mathbb{E}_{d^{\pi_{\eta}}(b)}\lspsqb{}  \int_{s\in\mathcal{S}} \int_{a\in\mathcal{A}} \pi_{\theta}(a | s)  \log \pi(a|b) \d a \ d^{\pi_{\eta}}(s | b) \d s \rspsqb{} , \\%
    &= \mathop{\argmin}_{\pi \in \Pi} K - \mathbb{E}_{d^{\pi_{\eta}}(b)}\lspsqb{} \int_{s\in\mathcal{S}} \int_{a\in\mathcal{A}} \pi_{\theta}(a | s)  \log \pi(a|b) \d a \ d^{\pi_{\eta}}(s | b) \d s \rspsqb{}  ,
\end{align}
where $K$ is independent of $\pi$.  Manipulating the rightmost term:
\begin{align}
    \pi^*(a|b) &= \mathop{\argmin}_{\pi \in \Pi} K - \mathbb{E}_{d^{\pi_{\eta}}(b)}\lspsqb{} \int_{a\in\mathcal{A}} \int_{s\in\mathcal{S}} \pi_{\theta}(a | s) d^{\pi_{\eta}}(s | b) \d s \  \log \pi(a | b)  \d a \rspsqb{} , \\
    &= \mathop{\argmin}_{\pi \in \Pi} K - \mathbb{E}_{d^{\pi_{\eta}}(b)}\lspsqb{} \int_{a\in\mathcal{A}} \hat{\pi}_{\theta}^{\eta}(a | b) \log \pi(a | b)  \d a \rspsqb{}  , 
\end{align}
We are now free to set the value of $K$, which we denote as $K'$, so long as it remains independent of $\pi$, as this does not alter the minimizing argument:
\begin{align}
    K' &= \mathbb{E}_{d^{\pi_{\eta}}(b)}\lspsqb{}  \int_{a\in\mathcal{A}} \hat{\pi}_{\theta}^{\eta}(a | b) \log \hat{\pi}_{\theta}^{\eta}(a | b)  \d a \rspsqb{} , \\
    \pi^*(a|b) &= \mathop{\argmin}_{\pi \in \Pi} K' - \mathbb{E}_{d^{\pi_{\eta}}(b)}\lspsqb{}  \int_{a\in\mathcal{A}} \hat{\pi}_{\theta}^{\eta}(a | b) \log \pi(a | b)  \d a \rspsqb{} , \\
    &= \mathop{\argmin}_{\pi \in \Pi} \mathbb{E}_{d^{\pi_{\eta}}(b)}\lspsqb{}  \int_{a\in\mathcal{A}} \hat{\pi}_{\theta}^{\eta}(a | b) \log \hat{\pi}_{\theta}^{\eta}(a | b)  \d a - \int_{a\in\mathcal{A}} \hat{\pi}_{\theta}^{\eta}(a | b) \log \pi(a | b) \d a \rspsqb{}  .
\end{align}
Combining the logarithms:
\begin{align}
    \pi^*(a|b) &= \mathop{\argmin}_{\pi \in \Pi} \mathbb{E}_{d^{\pi_{\eta}}(b)}\lspsqb{}  \int_{a\in\mathcal{A}} \hat{\pi}_{\theta}^{\eta}(a | b) \log \left( \frac{\hat{\pi}_{\theta}^{\eta}(a | b)}{\pi(a | b)} \right) \d a \rspsqb{} ,\\
    &= \mathop{\argmin}_{\pi \in \Pi} \mathbb{E}_{d^{\pi_{\eta}}(b)}\lspsqb{}  \mathbb{KL} \lspsqb{}  \hat{\pi}_{\theta}^{\eta}(a | b) \klbars{} \pi(a | b) \rspsqb{}  \rspsqb{}  .
\end{align}
Assuming that the policy class $\Pi$ is sufficiently expressive, this $\mathbb{KL}$ can be exactly minimized, and hence we arrive at the desired result:
\begin{align}
    \pi^*(a | b) = \hat{\pi}_{\theta}^{\eta}(a | b), \quad \forall\ a \in \mathcal{A},\ b \in \left\lbrace b' \in \mathcal{B} \mid d^{\pi_{\eta}}(b') > 0 \right\rbrace.
\end{align}
\end{proof}
This proof shows that learning the trainee policy ($\pi$ here, $\pi_{\psi}$ later) using $\mathbb{KL}$ minimization imitation learning (as in \eqref{equ:prelim:ail_target}) recovers the policy defined as the implicit policy (as defined in Definition \ref{def:implicit_policy}), and hence our definition of the implicit policy is well founded. 

\subsubsection{Variational Implicit Policy}\label{supp:sec:vap}
However, the implicit policy is defined as an intractable inference problem, marginalizing the conditional occupancy, $d^{\pi}(s \mid b)$, from which we cannot sample.  Therefore, we can further define a variational policy, $\pi_{\psi} \in \Pi_{\Psi}$, to approximate this policy, from which evaluating densities and sampling is more tractable.  This policy can be learned using gradient descent:
\begin{lemma}[Variational Implicit Policy Update, c.f. Section \ref{sec:prelim}, Equation \eqref{equ:def:variational:gradient}]
For an MDP $\mathcal{M}_{\Theta}$, POMDP $\mathcal{M}_{\Phi}$, and implicit policy $\hat{\pi}_{\theta}$ (Definition 1), if we define a variational approximation to $\hat{\pi}_{\theta}$, parameterized by $\psi$, denoted $\pi_{\psi} \in \Pi_{\Psi}$, such that the following divergence is minimized:
\begin{equation}
    \psi^* = \mathop{\argmin}_{\psi \in \Psi} F(\psi) = \mathop{\argmin}_{\psi \in \Psi} \mathbb{E}_{d^{\hat{\pi}_{\theta}}(b)} \lspsqb{}  \mathbb{KL} \lspsqb{}  {\hat{\pi}_{\theta}}(a|b) \klbars{}  \pi_{\psi}(a|b) \rspsqb{}  \rspsqb{} ,\label{supp:equ:l1:1}
\end{equation}
then an unbiased estimator for the gradient of this objective is given by the following expression:
\begin{align}
    \nabla_{\psi} F (\psi) &= - \mathbb{E}_{d^{\hat{\pi}_{\theta}}(s,b)} \lspsqb{}  \mathbb{E}_{\pi_{\theta}(a | s)} \lspsqb{}  \nabla_{\psi} \log \pi_{\psi} (a | b) \rspsqb{}  \rspsqb{} .
\end{align}
\end{lemma}
\begin{proof}
Note the objective in \eqref{supp:equ:l1:1} is corresponds to the original AIL objective via Theorem \ref{supp:theorem:1_ail_target}.  By manipulating the $\mathbb{KL}$ term, pulling out terms that are constant with respect to $\psi$, and rearranging the expectations we obtain:
\begin{align}
    F(\psi) &= \mathbb{E}_{d^{\hat{\pi}_{\theta}}(b)} \lspsqb{}  \mathbb{KL} \lspsqb{}  {\hat{\pi}_{\theta}}(a|b) \klbars{}  \pi_{\psi}(a|b) \rspsqb{}  \rspsqb{} , \\
    &= \mathbb{E}_{d^{\hat{\pi}_{\theta}}(b)} \lspsqb{}  \int_{a \in \mathcal{A}} \log \left(\frac{{\hat{\pi}_{\theta}}(a|b)}{{\pi_{\psi}}(a|b)}\right) {\hat{\pi}_{\theta}}(a|b) \d a \rspsqb{} , \\
    &= \int_{b \in \mathcal{B}} \int_{a \in \mathcal{A}} - \log {\pi_{\psi}}(a|b) {\hat{\pi}_{\theta}}(a|b) \d a \ d^{\hat{\pi}_{\theta}}(b) \d b + C,\\
    &= - \int_{b \in \mathcal{B}} \int_{a \in \mathcal{A}} \log {\pi_{\psi}}(a|b) \int_{s \in \mathcal{S}} \pi_{\theta}(a | s) d^{\hat{\pi}_{\theta}}(s | b) \d s \ \d a \ d^{\hat{\pi}_{\theta}}(b) \d b + C,\\
    &= - \int_{b \in \mathcal{B}} \int_{a \in \mathcal{A}}  \int_{s \in \mathcal{S}} \log {\pi_{\psi}}(a|b)\pi_{\theta}(a | s) d^{\hat{\pi}_{\theta}}(s , b) \d s \ \d a \ \d b + C,\\
    &= - \mathbb{E}_{d^{\hat{\pi}_{\theta}}(s, b)} \lspsqb{}  \int_{a \in \mathcal{A}} \log \pi_{\psi} (a | b) \pi_{\theta}(a | s) \d a \rspsqb{}  + C,  \\
    &= - \mathbb{E}_{d^{\hat{\pi}_{\theta}}(s, b)} \lspsqb{}  \mathbb{E}_{\pi_{\theta}(a | s)} \lspsqb{}  \log \pi_{\psi} (a | b) \rspsqb{}  \rspsqb{}  + C . \label{OLD:supp:equ:a2d:expectation}
\end{align}
As neither distribution in the expectation is a function of $\psi$, we can pass the derivative with respect to $\psi$ through this objective to obtain the gradient:
\begin{align}
    \nabla_{\psi} F (\psi) &= - \mathbb{E}_{d^{\hat{\pi}_{\theta}}(s, b)} \lspsqb{}  \mathbb{E}_{\pi_{\theta}(a | s)} \lspsqb{}  \nabla_{\psi} \log \pi_{\psi} (a | b) \rspsqb{}  \rspsqb{} . \label{supp:equ:l1:grad}
\end{align}
\end{proof}
Note here that in AIL $\theta$ is held constant.  In A2D we extend this by also updating the $\theta$, discussed later.  Importantly, the gradient estimator in \eqref{supp:equ:l1:grad} circumvents a critical issue in the initial definition of the implicit policy: we are unable to sample from the conditional occupancy, $d^{\hat{\pi}_{\theta}}(s \mid b)$.  However, and as is common in variational methods, the learning the variational policy only requires samples from the joint occupancy, $d^{\hat{\pi}_{\theta}}(s,b)$.  We can therefore train an approximator directly targeting the result of an intractable inference under the conditional density, and recover a variational policy that provides us with a convenient method of drawing (approximate) samples from the otherwise intractable implicit policy.  Under the relatively weak assumption that the variational family is sufficiently expressive, $\Pi_{\Psi} \supseteq \hat{\Pi}_{\Theta}$, this $\mathbb{KL}$ divergence can be exactly minimized, and exact samples from the implicit policy are recovered.  However, even if the expert policy is optimal under the MDP, and the divergence is minimized in the feasible set, this does not guarantee that the implicit policy (and hence the variational policy) is optimal under the partial information \emph{in terms of reward}, if the value of the divergence is not \emph{exactly} zero.  We therefore first build intermediate results by considering an identifiable process pair, where we show that we recover a sequence of updates which converges to the optimal partially observing policy, $\hat{\pi}_{\theta^*}(a \mid b)$, or its variational equivalent, $\pi_{\psi^*}(a \mid b)$.  In Section \ref{supp:sec:a2d} we then relax the identifiability requirement, and leverage these intermediate results to derive the A2D update in Theorem \ref{supp:def:exact_a2d}.

\subsubsection{Convergence of Iterative AIL}\label{supp:sec:convergence_of_occ}
We first verify the convergence of AIL for identifiable processes.  We will also introduce an assumption and two lemmas which provide important intermediate results and intuition, and will make the subsequent presentation of both Theorem \ref{supp:def:fixed_point_variational} and Theorem \ref{supp:def:exact_a2d} more compact. The assumption simply states that the variational family is sufficiently expressive such that the implicit policy can be replicated, and that the implicit policy is sufficiently expressive such that the optimal partially observing policy, $\pi_{\phi^*} \in \Pi_{\Phi^*} \subseteq \Pi_{\Phi}$, can actually be found.  

The first lemma shows that the solution to an iterative procedure, optimizing the trainee under the occupancy from the trainee policy at the previous iteration, actually converges to the solution of a single equivalent ``static'' optimization problem, directly optimizing over the trainee policy and the corresponding occupancy.  This will allow us to solve the challenging optimization over the trainee policy using a simple iterative procedure.  The second lemma shows that solving this static optimization is equivalent to an optimization under the occupancy induced by the implicit policy.  This will allow us to substitute the distribution under which we take expectations and will allow us to prove more complex relationships.  The assumption and both lemmas are then used in Theorem \ref{supp:def:fixed_point_variational} to show that iterative AIL will converge as required.

\begin{assumption}[Sufficiency of Policy Representations]\label{supp:assump:suff_var}
    We assume that for any behavioral policy, $\pi_{\eta} \in \Pi_{\Psi}$, the variational family is sufficiently expressive such that any implicit policy, $\hat{\pi}_{\theta} \in \hat{\Pi}_{\Theta}$, is exactly recovered in the regions of space where the occupancy under the occupancy under the behavioral policy places mass: 
    \begin{equation}
        \mathop{\min}_{\psi \in \Psi} \ \mathop{\mathbb{E}}_{d^{\pi_{\eta}}(b)} \lspsqb{}  \mathbb{KL} \lspsqb{}  {\hat{\pi}_{\theta}}(a|b) \klbars{}  \pi_{\psi}(a|b) \rspsqb{}  \rspsqb{}  = 0. \label{supp:equ:a1:1}
    \end{equation}
    We also assume that there exists an implicit policy, $\hat{\pi}_{\theta}$, such that an optimal POMDP policy, $\pi_{\phi^*} \in \Pi_{\Phi^*} \subseteq \Pi_{\Phi}$ can be represented:
    \begin{equation}
        \mathop{\min}_{\theta \in \Theta} \ \mathop{\mathbb{E}}_{d^{\pi_{\eta}}(b)} \lspsqb{}  \mathbb{KL} \lspsqb{}  \pi_{\phi^*}(a|b) \klbars{} {\hat{\pi}_{\theta}}(a|b) \rspsqb{}  \rspsqb{}  = 0,\label{supp:equ:a1:2}
    \end{equation}
    and hence there is a variational policy that can represent the optimal POMDP policy in states visited under $\pi_{\eta}$.
\end{assumption}

The condition in Equation \eqref{supp:equ:a1:1} (and similarly the condition in Equation \eqref{supp:equ:a1:2}) can also be written as:
\begin{equation}
    \exists \psi \in \Psi \quad \text{such\ that} \quad \hat{\pi}_{\theta}(a|b) = \pi_{\psi}(a|b), \quad \forall\ a \in \mathcal{A},\ b \in \left\lbrace b' \in \mathcal{B} \mid d^{\pi_{\eta}}(b') > 0 \right\rbrace.
\end{equation}
These conditions are weaker than simply requiring $\hat{\pi}_{\theta} \in \hat{\Pi}_{\Theta} \subseteq \Pi_{\Psi}$, as this only requires that the policies are equal where the occupancy places mass.  These assumptions are often made implicitly by AIL methods.  We will use this assumption throughout.  Note that by definition if the divergence in \eqref{supp:equ:a1:2} is equal to zero at all $\hat{\pi}_{\theta^*}$, then the processes are identifiable.  

\begin{lemma}[Convergence of Iterative Procedure]\label{supp:lemma:2}
    For an MDP $\mathcal{M}_{\Theta}$ and POMDP $\mathcal{M}_{\Phi}$, and implicit policy $\hat{\pi}_{\theta}$ (Definition \ref{def:implicit_policy}), if we define a variational approximation to $\hat{\pi}_{\theta}$, parameterized by $\psi$, denoted $\pi_{\psi} \in \Pi_{\Psi}$, then under Assumption \ref{supp:assump:suff_var}, and for the following AIL objective:
    \begin{equation}
       \quad \psi^* = \mathop{\argmin}_{\psi \in \Psi} \mathop{\mathbb{E}}_{d^{\pi_{\psi}}(b)} \lspsqb{}  \mathbb{KL} \lspsqb{}  \hat{\pi}_{\theta}(a|b) \klbars{} \pi_\psi(a|b) \rspsqb{}  \rspsqb{} ,\label{supp:equ:t2:condition2}
    \end{equation}
    the iterative scheme:
    \begin{align}
        \label{supp:equ:t2:condition1}
       \psi_{k+1} &= \mathop{\argmin}_{\psi \in \Psi}  \mathop{\mathbb{E}}_{d^{\pi_{\psi_k}}(b)}  \lspsqb{}  \mathbb{KL} \lspsqb{}  \hat{\pi}_{\theta}(a|b) \klbars{} \pi_\psi(a|b) \rspsqb{}  \rspsqb{} ,  \quad \text{with} \quad \psi_{\infty} = \lim_{k \rightarrow \infty} \psi_k,
    \end{align}
    converges to the solution to the optimization problem in Equation \eqref{supp:equ:t2:condition2} such that:
    \begin{align}
        \mathop{\mathbb{E}}_{d^{\pi_{\psi^*}}(b)}  \lspsqb{}  \mathbb{KL} \lspsqb{}  \pi_{\psi^*}(a|b) \klbars{} \pi_{\psi_{\infty}}(a|b) \rspsqb{}  \rspsqb{}  = 0 \label{supp:equ:l2:result}
    \end{align}
\end{lemma}
\begin{proof}
    We show this convergence by showing that the total variation between $d^{_{\psi^*}}(b)$ and $d^{_{\psi_k}}(b)$, over the set of belief states visited in Equation \eqref{supp:equ:l2:result}, denoted $b \in \hat{\mathcal{B}} = \left\lbrace b' \in \mathcal{B} \mid d^{\pi_{\psi^*}}(b') > 0 \right\rbrace$, converges to zero as $k \tends \infty$.  We begin by expressing the total variation at the $k^{\mathrm{th}}$ iteration:
    \begin{align}
        \sup_{b \in \hat{\mathcal{B}}} \left|d^{\pi_{\psi_k}}(b)-d^{\pi_{\psi^*}}(b)\right| &= \sup_{b \in \hat{\mathcal{B}}} \left|(1-\gamma) \sum_{t=0}^{\infty} \gamma^t q_{\pi_{\psi_k}}(b_t)-(1-\gamma) \sum_{t=0}^{\infty} \gamma^t q_{\pi_{\psi^*}}(b_t)\right|, \\
        &= (1-\gamma) \sup_{b \in \hat{\mathcal{B}}} \left| \sum_{t=0}^{\infty} \gamma^t q_{\pi_{\psi_k}}(b_t)- \sum_{t=0}^{\infty} \gamma^t q_{\pi_{\psi^*}}(b_t)\right|, \\
        &= (1-\gamma) \sup_{b \in \hat{\mathcal{B}}} \left| \sum_{t=0}^{k} \gamma^t q_{\pi_{\psi_k}}(b_t) + \sum_{t=k+1}^{\infty} \gamma^t q_{\pi_{\psi_k}}(b_t) - \sum_{t=0}^{k} \gamma^t q_{\pi_{\psi^*}}(b_t) - \sum_{t=k+1}^{\infty} \gamma^t q_{\pi_{\psi^*}}(b_t)\right| .
    \end{align}
    where $\gamma \in [0,1)$, and where we use the notational shorthand by defining $b_0, b_1, b_2, \ldots = b$.  
    
    We can then note that at the $k^{\mathrm{th}}$ iteration, the distribution over the first $k$ state-belief state pairs must be identical: $q_{\pi_{\psi_k}}(\tau_{0:k-1}) = q_{\pi_{\psi^*}}(\tau_{0:k-1})$ (recalling that $\tau$ contains both belief state and actions).  To verify this, consider the following inductive argument: If after a single iteration ($k=1$) we have exactly minimized the $\mathbb{KL}$ divergence between $\hat{\pi}_{\theta}$ and $\pi_{\psi_1}$ (and hence the divergence between $\pi_{\psi_1}$ and $\pi_{\psi^*}$) for all $b_0 \in \left\lbrace b_0 \in \mathcal{B} \mid q_{\pi_{\psi_k}}(b_0)>0 \right\rbrace$., then at time step zero the following equality must hold $q_{\pi_{\psi_1}}(\tau_0) = q_{\pi_{\psi^*}}(\tau_0)$, because the distribution over actions and the underlying dynamics are the same at the initial state and belief state. Therefore, because both the distribution over the initial state and belief state, as well as the action distributions must also be the same for $q_{\pi_{\psi^*}}$ and $q_{\pi_{\psi_1}}$ (i.e. $q_{\pi_{\psi_1}}(a_0,b_{0}) = q_{\pi_{\psi^*}}(a_{0},b_{0})$) then necessarily we have that $q_{\pi_{\psi_1}}(b_{1}) = q_{\pi_{\psi^*}}(b_{1})$.
    
    Next, using the inductive hypothesis $q_{\pi_{\psi_k}}(b_{k-1}) = q_{\pi_{\psi^*}}(b_{k-1})$, we can see that provided \eqref{supp:equ:a1:1} is exactly minimized, then $\pi_{\psi_{k-1}}(a_{k-1}|b_{k-1}) = \pi_{\psi^*}(a_{k-1}|b_{k-1})$. This then means that again we have $q_{\pi_{\psi_k}}(a_{k-1},b_{k-1}) = q_{\pi_{\psi^*}}(a_{k-1},b_{k-1})$, which by definition gives $q_{\pi_{\psi_k}}(b_{k}) = q_{\pi_{\psi^*}}(b_{k})$, which concludes our inductive proof. This allows us to make the following substitution and simplification:
    \begin{align}
        \sup_{b \in \hat{\mathcal{B}}} \left|d^{\pi_{\psi_k}}(b)-d^{\pi_{\psi^*}}(b)\right| &= (1-\gamma) \sup_{b \in \hat{\mathcal{B}}} \left| \sum_{t=0}^{k} \gamma^t q_{\pi_{\psi^*}}(b_t) + \sum_{t=k+1}^{\infty} \gamma^t q_{\pi_{\psi_k}}(b_t) - \sum_{t=0}^{k} \gamma^t q_{\pi_{\psi^*}}(b_t) - \sum_{t=k+1}^{\infty} \gamma^t q_{\pi_{\psi^*}}(b_t)\right|, \\
        &= (1-\gamma) \sup_{b \in \hat{\mathcal{B}}} \left| \sum_{t=k+1}^{\infty} \gamma^t q_{\pi_{\psi_k}}(b_t) - \sum_{t=k+1}^{\infty} \gamma^t q_{\pi_{\psi^*}}(b_t)\right| ,\\
        &= (1-\gamma) \sup_{b \in \hat{\mathcal{B}}} \left| \sum_{t=k+1}^{\infty} \gamma^t (q_{\pi_{\psi_k}}(b_t) - q_{\pi_{\psi^*}}(b_t)) \right|, \\
        &\leq (1-\gamma) \sup_{b \in \hat{\mathcal{B}}} \left| \sum_{t=k+1}^{\infty} \gamma^t C \right|
        = C (1-\gamma) \sum_{t=k+1}^{\infty} \gamma^t = C (1-\gamma) \left(\frac{1}{1-\gamma} - \frac{1-\gamma^{k+1}}{1-\gamma} \right)
        \\
        &= C(1 - 1 + \gamma^{k+1}) = C\gamma^{k+1} = O(\gamma^{k}),
    \end{align}
    where we assume that the maximum variation between the densities is bounded by $C \in \mathbb{R}_+$.  Hence, as $\gamma \in [0, 1)$, as $k\tends\infty$ the occupancy induced by the trainee learned through the iterative procedure, $d^{\pi_{\psi_{\infty}}}$, converges to the occupancy induced by the optimal policy recovered through direct, static optimization, $d^{\pi_{\psi^*}}$.  As a result of this, and the expressivity assumption in \eqref{supp:equ:a1:1}, we can state that the iterative procedure must recover a perfect variational approximation to the implicit policy $\hat{\pi}_{\theta}$, in belief states with finite mass under $d^{\hat{\pi}_{\theta}}$. 
\end{proof}

This lemma verifies that we can solve for a variational approximation to a particular implicit policy, defined by the static-but-difficult optimization defined in \eqref{supp:equ:t2:condition2}, by using the tractable iterative procedure defined in \eqref{supp:equ:t2:condition1}. However, the distribution under which we take the expectation is the trainee policy.  We therefore show now that this can be replaced with the occupancy under the implicit policy, which will allow us to utilize the identifiability condition defined in the main text.

\begin{lemma}[Equivalence of Objectives]\label{supp:lemma:3}
    For an MDP $\mathcal{M}_{\Theta}$, POMDP $\mathcal{M}_{\Phi}$, and implicit policy $\hat{\pi}_{\theta}$ (Definition \ref{def:implicit_policy}), if we define a variational approximation to $\hat{\pi}_{\theta}$, parameterized by $\psi$, denoted $\pi_{\psi} \in \Pi_{\Psi}$, and define: 
    \begin{align}
       \psi^*_1 &= \mathop{\argmin}_{\psi \in \Psi}  \mathop{\mathbb{E}}_{d^{\pi_{\psi}}(b)}  \lspsqb{}  \mathbb{KL} \lspsqb{}  \hat{\pi}_{\theta}(a|b) \klbars{} \pi_\psi(a|b) \rspsqb{}  \rspsqb{} ,  \label{supp:equ:make_it_stop_before}\\
       \psi^*_2 &= \mathop{\argmin}_{\psi \in \Psi}  \mathop{\mathbb{E}}_{d^{\hat{\pi}_{\theta}}(b)}  \lspsqb{}  \mathbb{KL} \lspsqb{}  \hat{\pi}_{\theta}(a|b)\klbars{} \pi_\psi(a|b) \rspsqb{}  \rspsqb{} , \label{supp:equ:make_it_stop}
    \end{align}
    then, under Assumption \ref{supp:assump:suff_var}, we are able to show that:
    \begin{align}
        \mathop{\mathbb{E}}_{d^{\pi_{\psi^*_2}}(b)}  \lspsqb{}  \mathbb{KL} \lspsqb{}  \pi_{\psi^*_2}(a|b) \klbars{} \pi_{\psi^*_1}(a|b) \rspsqb{}  \rspsqb{} = 0
    \end{align}
\end{lemma}
\begin{proof}
    We show this result by way of contradiction.  First assume that there exists some $t \in \mathbb{N}$ such that $q_{\hat{\pi}_{\theta}}(b_t) \neq q_{\pi_{\psi_k}}(b_t)$.  As a result of Assumption \ref{supp:assump:suff_var} we can state that:
    \begin{equation}
            \min_{\psi \in \Psi} \ \mathbb{E}_{d^{\pi_{\psi}}(b)} \lspsqb{}  \mathbb{KL} \lspsqb{}  {\hat{\pi}_{\theta}}(a|b) \klbars{}  \pi_{\psi}(a|b) \rspsqb{}  \rspsqb{}  = 0.
    \end{equation}
    We now use a similar approach to the one used in Lemma \ref{supp:lemma:2}, and consider initially the first time step.  We note that $q_{\hat{\pi}_{\theta}}(b_0) = q_{\pi_{\psi}}(b_0)$ because the initial state distribution is independent of the policy.  Because both \eqref{supp:equ:make_it_stop_before} and \eqref{supp:equ:make_it_stop} target the same density, by Assumption \ref{supp:assump:suff_var}, after the first iteration we again have that $q_{\hat{\pi}_{\theta}}(b_0)\hat{\pi}_{\theta}(a_0|b_0) = q_{\pi_{\psi}}(b_0)\pi_{\psi}(a_0|b_0)$. Because the dynamics are the same for both $q_{\pi_{\psi}}$ and $q_{\hat{\pi}_{\theta}}$, this result directly implies that $q_{\hat{\pi}_{\theta}}(b_{1}) = q_{\pi_{\psi}}(b_{1})$.
    
    Inductively extending this to $t-1$, we have that $q_{\hat{\pi}_{\theta}}(b_{t-1}) = q_{\pi_{\psi}}(b_{t-1})$, and further, that our action distribution again satisfies  $\pi_{\psi_t}(a_{t-1}|b_{t-1}) = \hat{\pi}_{\theta}(a_{t-1}|b_{t-1})$ due to Assumption \ref{supp:assump:suff_var}. Here we again have that $\pi_{\psi_t}(a_{t-1}|b_{t-1})q_{\pi_{\psi}}(b_{t-1}) = \hat{\pi}_{\theta}(a_{t-1}|b_{t-1})q_{\hat{\pi}_{\theta}}(b_{t-1})$, which directly implies that $q_{\hat{\pi}_{\theta}}(b_{t}) = q_{\pi_{\psi}}(b_{t})$ must also hold. However this contradicts our assumption that $\exists t \in \mathbb{N}$ such that  $q_{\hat{\pi}_{\theta}}(b_t) \neq q_{\pi_{\psi_k}}(b_t)$. Thus under the assumptions stated above, $q_{\hat{\pi}_{\theta}}(b_t) = q_{\pi_{\psi}}(b_t)$ for all $t$, and by extension, $d^{\hat{\pi}_{\theta}}(b) = d^{\pi_{\psi^*}}(b)$, where $\pi_{\psi^*}$ represents a solution to the right hand side of Equation \eqref{supp:equ:make_it_stop}. 
\end{proof}

This lemma allows us to exchange the distribution under which we take expectations.  We can now use Assumption \ref{supp:assump:suff_var}, Lemma \ref{supp:lemma:2} and Lemma \ref{supp:lemma:3} to show that for an identifiable process pair an iterative AIL procedure converges to the correct POMDP policy as desired.

% \todo{this needs cleaning with the $t$'s removed}
\setcounter{sthe}{\value{theorem}}
\setcounter{theorem}{1}                   % CHANGE THE COUNTER VALUES IN 
\begin{theorem}[Convergence of AIL, expanded from Section \ref{sec:il-failure}]\label{supp:def:fixed_point_variational}
Consider an \emph{identifiable} MDP-POMDP process pair ($\mathcal{M}_{\Theta}$, $\mathcal{M}_{\Phi}$), with optimal expert policy, $\pi_{\theta^*}$, and optimal partially observing policy $\pi_{\phi^*} \in \Pi_{\Phi^*} \subseteq \Pi_{\Phi}$.  For a variational policy $\pi_{\psi} \in \Pi_{\Psi}$, and assuming Assumption \ref{supp:assump:suff_var} holds, the following iterative procedure:
\begin{equation}
    \psi_{k+1} = \mathop{\argmin}_{\psi \in \Psi}  \mathop{\mathbb{E}}_{d^{\pi_{\psi_k}}(s,b)}  \lspsqb{}  \mathbb{KL} \lspsqb{}  \pi_{\theta^*}(a|s) \klbars{} \pi_\psi(a|b) \rspsqb{}  \rspsqb{} , \label{supp:equ:t1:1}
\end{equation}
converges to parameters $\psi^* = \lim_{k\tends\infty} \psi_{k}$ that define a policy equal to an optimal partially observing policy in visited regions of state-space: 
\begin{equation}
    \mathop{\mathbb{E}}_{d^{\pi_{\phi^*}}(b)}  \lspsqb{}  \mathbb{KL} \lspsqb{}  \pi_{\phi^*}(a|b) \klbars{} \pi_{\psi^*}(a|b) \rspsqb{}  \rspsqb{}  = 0 \label{supp:equ:please_make_it_stop}
\end{equation}
\end{theorem}
\setcounter{theorem}{\value{sthe}}
\begin{proof}
For brevity, we present this proof for the case that there is a unique optimal parameter value, $\psi^*$.  However, this is not a \emph{requirement}, and can easily be relaxed to consider a set of equivalent parameters, $\psi^*_{1:N}$, that yield the same policy over the relevant occupancy distribution, i.e. $\pi_{\psi^*_1}(a | b) = \ldots = \pi_{\psi^*_N}(a | b) \ \forall b \in \hat{B}$.  In this case, we would instead require that the $\mathbb{KL}$ divergence between the resulting policies is zero (analogous to \eqref{supp:equ:please_make_it_stop}), as opposed to requiring that the parameters recovered are \emph{equal} to $\psi^*$.  However, including this dramatically complicates the exposition and hence we do not include such a proof here. We begin by considering the limiting behavior of \eqref{supp:equ:t1:1} as $k \tends \infty$:
\begin{align}
    \psi^* = \lim_{k \rightarrow \infty} \mathop{\argmin}_{\psi \in \Psi}  \mathop{\mathbb{E}}_{d^{\pi_{\psi_k}}(s,b)}  \lspsqb{}  \mathbb{KL} \lspsqb{}  \pi_{\theta^*}(a|s) \klbars{} \pi_\psi(a|b) \rspsqb{}  \rspsqb{} .
\end{align}
Application of Theorem \ref{supp:theorem:1_ail_target} to replace the expert policy with the implicit policy yields:
\begin{align}
    = \lim_{k \rightarrow \infty} \mathop{\argmin}_{\psi \in \Psi}  \mathop{\mathbb{E}}_{d^{\pi_{\psi_k}}(b)}  \lspsqb{}  \mathbb{KL} \lspsqb{}  \hat{\pi}_{\theta^*}(a|b) \klbars{} \pi_\psi(a|b) \rspsqb{}  \rspsqb{} .  \label{supp:equ:iterative_conv_1}
\end{align}
Application of Lemma \ref{supp:lemma:2} to \eqref{supp:equ:iterative_conv_1} then recovers the limiting behavior as $k\tends \infty$:
\begin{align}
    = \mathop{\argmin}_{\psi \in \Psi}  \mathop{\mathbb{E}}_{d^{\pi_{\psi}}(b)}  \lspsqb{}  \mathbb{KL} \lspsqb{}  \hat{\pi}_{\theta^*}(a|b) \klbars{} \pi_\psi(a|b) \rspsqb{}  \rspsqb{} .
\end{align}
Application of Lemma \ref{supp:lemma:3} to change the distribution under which the expectation is taken yields: 
\begin{align}
    = \mathop{\argmin}_{\psi \in \Psi}  \mathop{\mathbb{E}}_{d^{\hat{\pi}_{\theta^*}}(b)}  \lspsqb{}  \mathbb{KL} \lspsqb{}  \hat{\pi}_{\theta^*}(a|b)\klbars{} \pi_\psi(a|b) \rspsqb{}  \rspsqb{} , \label{supp:equ:murh}
\end{align}
Identifiability then directly implies that the implicit policy defined by the optimal expert policy \emph{is} an optimal partially observing policy:
\begin{align}
    \mathop{\mathbb{E}}_{d^{\pi_{\phi^*}}(b)}  \lspsqb{}  \mathbb{KL} \lspsqb{}  \pi_{\phi^*}(a|b) \klbars{} \hat{\pi}_{\theta^*}(a|b)  \rspsqb{}  \rspsqb{}  = 0,  
\end{align}
and therefore we can replace $\hat{\pi}_{\theta^*}$ with $\pi_{\phi^*}$ in \eqref{supp:equ:murh} to yield:
\begin{align}
    \psi^* = \mathop{\argmin}_{\psi \in \Psi}  \mathop{\mathbb{E}}_{d^{\pi_{\phi^*}}(b)}  \lspsqb{}  \mathbb{KL} \lspsqb{}  \pi_{\phi^*}(a|b) \klbars{} \pi_{\psi}(a|b) \rspsqb{}  \rspsqb{} , 
\end{align}
Finally, under Assumption \ref{supp:assump:suff_var}, the expected $\mathbb{KL}$ divergence in \eqref{supp:equ:murh} can be exactly minimized, such that:
\begin{equation}
    \mathop{\mathbb{E}}_{d^{\pi_{\phi^*}}(b)}  \lspsqb{}  \mathbb{KL} \lspsqb{}  \pi_{\phi^*}(a|b) \klbars{} \pi_{\psi^*}(a|b) \rspsqb{}  \rspsqb{}  = 0
\end{equation}
\end{proof}

This proof shows that, if Assumption \ref{supp:assump:suff_var} holds and for an identifiable MDP-POMDP pair, we can use a convenient iterative scheme defined in \eqref{supp:equ:t1:1} to recover an optimal trainee (variational) policy that is exactly equivalent to an optimal partially observing policy.  This iterative process is more tractable than the directly solving the equivalent static optimization; instead gathering trajectories under the current trainee policy, regressing the trainee onto the expert policy at each state, and then rolling out under the new trainee policy until convergence.  However, assuming that processes are identifiable is a \emph{very} restrictive assumption.  This fact motivates our A2D algorithm, which exploits AIL to recover an optimal partially observing policy for any process pair by adaptively modifying the expert that is imitated by the trainee.

\subsection{A2D Proofs}\label{supp:sec:a2d}
In this section we provide the proofs, building on the results given above, that underpin our A2D method and facilitate robust exploitation of AIL in non-identifiable process pairs.  We begin this section by giving a proof of the bound described in \eqref{equ:bound:1}-\eqref{equ:bound:2}.  We then give proofs of the A2D gradient estimator given in \eqref{equ:expert-gradient}.  We then conclude with a proof of Theorem \ref{supp:def:exact_a2d}, which closely follows the proof for Theorem \ref{supp:def:fixed_point_variational}, and provides the theoretical underpinning of the A2D algorithm.  We conclude by discussing briefly the practical repercussions of this result, as well as some additional assumptions that can be made to simplify the analysis. 

\subsubsection{Objectives and Gradients Estimators}

We begin by expanding on the policy gradient bound given in \eqref{equ:bound:1}-\eqref{equ:bound:2}.
\begin{lemma}[Policy gradients bound, c.f. Section \ref{sec:algorithm}, Equations \eqref{equ:bound:1}-\eqref{equ:bound:2}]
Consider an expert policy, $\pi_{\theta}$, and a trainee policy learned through $\mathbb{KL}$-minimization, $\pi_{\psi}$, targeting the implicit policy, $\hat{\pi}_{\theta}$. If \eqref{supp:equ:a1:2} in Assumption \ref{supp:assump:suff_var} holds, the following bound holds:
\begin{align}
    \max_{\theta \in \Theta} J_{\psi}(\theta) = \max_{\theta \in \Theta} \mathop{\mathbb{E}}_{{\hat{\pi}_\theta(a|b) d^{{\pi_{\psi}}}(b)}} \lspsqb{}  Q^{{\pi_{\psi}}}(a,b) \rspsqb{}  
    \leq \max_{\theta \in \Theta}  \mathop{\mathbb{E}}_{\hat{\pi}_\theta(a|b) d^{{\pi_{\psi}}}(b)} \lspsqb{}  Q^{{\hat{\pi}_{\theta}}}(a,b) \rspsqb{}  = \max_{\theta \in \Theta} J(\theta) . \label{supp:equ:bound}
\end{align}
\end{lemma}
\begin{proof}
For a more extensive discussion on this form of policy improvement we refer the reader to \citet{pmlr-v125-agarwal20a,bertsekas1991analysis,bertsekas2011approximate}. Assumption \ref{supp:assump:suff_var} states that the optimal partially observing policy (or policies) is representable by an implicit policy for any occupancy distribution.  We denote the optimal value function as $V^*(b)$, where this value function is realizable by the implicit policy.  Considering the right hand side of \eqref{supp:equ:bound}, we can write, by definition, the following equality: 
\begin{align}
    \max_{\theta \in \Theta} \mathop{\mathbb{E}}_{\hat{\pi}_\theta(a|b) d^{\pi_{\psi}}(b)} \lspsqb{}  Q^{{\hat{\pi}_{\theta}}}(a,b) \rspsqb{} 
    &= 
    \max_{\theta \in \Theta} \mathop{\mathbb{E}}_{d^{\pi_{\psi}}(b)} \lspsqb{}  
    \mathop{\mathbb{E}}_{\hat{\pi}_\theta(a|b)}\lspsqb{}  
    \mathop{\mathbb{E}}_{p(b'|a,b)}[r(b,a,b')] + \gamma  \mathop{\mathbb{E}}_{p(b'|a,b)}[V^{\hat{\pi}_{\theta}}(b')] \rspsqb{} \rspsqb{}  
    \\
    &=
    \max_{\theta \in \Theta} \mathop{\mathbb{E}}_{d^{\pi_{\psi}}(b)} \lspsqb{}  
    \mathop{\mathbb{E}}_{\hat{\pi}_\theta(a|b)}\lspsqb{}  
    \mathop{\mathbb{E}}_{p(b'|a,b)}[r(b,a,b')] + \gamma  \mathop{\mathbb{E}}_{p(b'|a,b)}[V^{*}(b')] \rspsqb{} \rspsqb{} 
\end{align}
We then repeat this for the expression on the left side of \eqref{supp:equ:bound}, noting that instead of equality there is an inequality, as by definition the value function induced by $\pi_{\psi}(a|b)$, denoted $V^{\pi_{\psi}}(b)$, cannot be \emph{greater} than $V^*(b)$:
\begin{align}
     \quad V^{\pi_{\psi}}(b) &\leq  V^{*}(b) \quad \forall \ b \in \left\lbrace \tilde{b} \in \mathcal{B} \mid d^{\pi_{\psi}}(\tilde{b}) > 0 \right\rbrace, \\
    \max_{\theta \in \Theta} \mathop{\mathbb{E}}_{\hat{\pi}_\theta(a|b) d^{\pi_{\psi}}(b)} \lspsqb{}  Q^{{\pi_\psi}}(a,b) \rspsqb{} 
    &= 
    \max_{\theta \in \Theta} \mathop{\mathbb{E}}_{d^{\pi_{\psi}}(b)} \lspsqb{}  
    \mathop{\mathbb{E}}_{\hat{\pi}_\theta(a|b)}\lspsqb{}  
    \mathop{\mathbb{E}}_{p(b'|a,b)}[r(b,a,b')] + \gamma \mathop{\mathbb{E}}_{p(b'|a,b)} \lspsqb{}  V^{\pi_{\psi}}(b')\rspsqb{}  \rspsqb{} \rspsqb{}  \\
    &\leq \max_{\theta \in \Theta} \mathop{\mathbb{E}}_{d^{\pi_{\psi}}(b)} \lspsqb{}  
    \mathop{\mathbb{E}}_{\hat{\pi}_\theta(a|b)}\lspsqb{}  
    \mathop{\mathbb{E}}_{p(b'|a,b)}[r(b,a,b')] + \gamma  \mathop{\mathbb{E}}_{p(b'|a,b)}[V^{*}(b')] \rspsqb{} \rspsqb{} ,
\end{align}
and hence the inequality originally stated in \eqref{supp:equ:bound} must hold.
\end{proof}

This form of improvement over a behavioral policy is well studied in the approximate dynamic programming literature~\cite{bertsekas2019reinforcement}, and is a useful tool in analyzing classical methods such as approximate policy iteration. As was discussed in Section \ref{sec:algorithm}, it is also implicitly used in many policy gradient algorithms to avoid differentiating through the Q function, especially when a differentiable Q function is not available. In these cases (i.e. \citet{schulman2017proximal,schulman2015trust,schulman2015high,williams1992simple}) the behavioral policy is defined as the policy under which samples are gathered for Q function estimation. Then, as in the classical policy gradient theorem~\citep{bertsekas2019reinforcement,sutton1992reinforcement,williams1992simple}, the discounted sum of rewards ahead does not need to be differentiated through.  We can then exploit this lower bound to construct an estimator of the gradient of the expert parameters with respect to the reward garnered by the implicit policy. 

\begin{lemma}[A2D Q-based gradient estimator, c.f. Section \ref{sec:algorithm}, Equation \eqref{equ:expert-gradient}] 
For an expert policy, $\pi_{\theta}$, and a trainee policy learned through $\mathbb{KL}$-minimization, $\pi_{\psi}$, targeting the implicit policy, $\hat{\pi}_{\theta}$, we can transform the following policy gradient update applied directly to the trainee policy lower bound in \eqref{supp:equ:bound}:
\begin{align}
    \nabla_{\theta} J_{\psi}(\theta) =  \nabla_{\theta} \mathop{\mathbb{E}}_{{\hat{\pi}_\theta(a|b) d^{{\pi_{\psi}}}(b)}} \left[ Q^{{\pi_{\psi}}}(a,b) \right], \label{supp:equ:a2d_q_1}
\end{align}
into a policy gradient update applied to the expert:
\begin{align}
    \nabla_{\theta} J_{\psi}(\theta) &= \mathop{\mathbb{E}}_{d^{\pi_{\psi}}(s, b)} \lspsqb{}  \mathop{\mathbb{E}}_{\pi_{\theta}(a | s)} \lspsqb{}  Q^{\pi_{\psi}}(a,b)  \nabla_{\theta} \log \pi_{\theta}(a|s) \rspsqb{}  \rspsqb{} , 
\end{align}
\end{lemma}
\begin{proof}
To prove this we simply expand and rearrange \eqref{supp:equ:a2d_q_1}:
\begin{align}
    \nabla_{\theta} J_{\psi}(\theta) &= \nabla_{\theta} \mathop{\mathbb{E}}_{{\hat{\pi}_{\theta}(a|b) d^{{\pi_{\psi}}}(b)}} \lspsqb{}  Q^{\pi_{\psi}}(a,b) \rspsqb{} , \\%
    &= \nabla_{\theta} \int_{b\in\mathcal{B}} \int_{a\in\mathcal{A}} Q^{\pi_{\psi}}(a,b) \hat{\pi}_{\theta}(a | b) \d a \ d^{\pi_{\psi}}(b) \d b, \\%
    &= \nabla_{\theta} \int_{b\in\mathcal{B}} \int_{a\in\mathcal{A}} Q^{\pi_{\psi}}(a,b) \int_{s\in\mathcal{S}} \pi_{\theta}(a | s) d^{\pi_{\psi}}(s | b) \d s \ \d a \  d^{\pi_{\psi}}(b) \d b, \\%
    &= \nabla_{\theta} \int_{s\in\mathcal{S}} \int_{b\in\mathcal{B}} \int_{a\in\mathcal{A}} Q^{\pi_{\psi}}(a,b) \pi_{\theta}(a | s) d^{\pi_{\psi}}(s, b) \d a \ \d s \  \d b , \\%
    &= \mathop{\mathbb{E}}_{d^{\pi_{\psi}}(s, b)} \lspsqb{}  \nabla_{\theta} \int_{a\in\mathcal{A}} Q^{\pi_{\psi}}(a,b) \pi_{\theta}(a | s) \d a \rspsqb{} , \\ %
    &= \mathop{\mathbb{E}}_{d^{\pi_{\psi}}(s, b)} \lspsqb{}  \nabla_{\theta} \mathop{\mathbb{E}}_{\pi_{\theta}(a | s)} \lspsqb{}  Q^{\pi_{\psi}}(a,b) \rspsqb{}  \rspsqb{} , \\ %
    &= \mathop{\mathbb{E}}_{d^{\pi_{\psi}}(s, b)} \lspsqb{}  \mathop{\mathbb{E}}_{\pi_{\theta}(a | s)} \lspsqb{}  Q^{\pi_{\psi}}(a,b)  \nabla_{\theta} \log \pi_{\theta}(a|s) \rspsqb{}  \rspsqb{} , \label{supp:equ:aaaarrhhh}%
\end{align}
\end{proof}
The A2D gradient estimator given in \eqref{equ:expert-gradient} then adds an importance weight to the inner expectation, as we rollout under $\pi_{\psi}$.  This allows us to instead weight actions sampled under the current trainee policy, $\pi_{\psi}$, without biasing the gradient estimator.  We can then cast this estimator in terms of advantage, where the Q function with the value function subtracted as a baseline to reduce the variance of the estimator.

\begin{lemma}[A2D Advantage-based gradient estimator, c.f. Section \ref{sec:algorithm}, Equation \eqref{equ:a2d:a2d_update}] 
We can construct a gradient estimator from \eqref{supp:equ:aaaarrhhh} that uses the advantage by subtracting the value function as a \emph{baseline}~\cite{bertsekas2019reinforcement, sutton1992reinforcement,williams1992simple}:
\begin{align}
    \nabla_{\theta} J_{\psi}(\theta) &= \mathop{\mathbb{E}}_{d^{\pi_{\psi}}(s, b)} \lspsqb{}  \mathop{\mathbb{E}}_{\pi_{\theta}(a | s)} \lspsqb{}  Q^{\pi_{\psi}}(a,b)  \nabla_{\theta} \log \pi_{\theta}(a|s) \rspsqb{}  \rspsqb{}, \\
    &= \mathop{\mathbb{E}}_{d^{\pi_{\psi}}(s, b)} \lspsqb{}  \mathbb{E}_{\pi_{\theta}(a | s)} \lspsqb{}  (Q^{\pi_{\psi}}(a,b)-V^{\pi_{\psi}}(b))  \nabla_{\theta} \log \pi_{\theta}(a|s) \rspsqb{}  \rspsqb{} .
\end{align}
\end{lemma}
\begin{proof}
It is sufficient to show that:
\begin{equation}
    \mathop{\mathbb{E}}_{d^{\pi_{\psi}}(s, b)} \lspsqb{}  \mathop{\mathbb{E}}_{\pi_{\theta}(a | s)} \lspsqb{}  V^{\hat{\pi}_{\theta}}(b) \nabla_{\theta} \log \pi_{\theta}(a|s) \rspsqb{}  \rspsqb{}  = 0,
\end{equation}
which can be shown easily as:
\begin{align}
    \mathbb{E}_{d^{\pi_{\psi}}(s, b)} \lspsqb{}  \mathbb{E}_{\pi_{\theta}(a | s)} \lspsqb{}  V^{\hat{\pi}_{\theta}}(b) \nabla_{\theta} \log \pi_{\theta}(a|s) \rspsqb{}  \rspsqb{} &= \mathbb{E}_{d^{\pi_{\psi}}(s, b)} \lspsqb{}  V^{\hat{\pi}_{\theta}}(b) \mathbb{E}_{\pi_{\theta}(a | s)} \lspsqb{}   \nabla_{\theta} \log \pi_{\theta}(a|s) \rspsqb{}  \rspsqb{}  \\
    &= \mathbb{E}_{d^{\pi_{\psi}}(s, b)} \lspsqb{}  V^{\hat{\pi}_{\theta}}(b) \int_{a \in \mathcal{A}} \nabla_{\theta} \pi_{\theta}(a|s) \d a \rspsqb{} ,\\
    &= \mathbb{E}_{d^{\pi_{\psi}}(s, b)} \lspsqb{}  V^{\hat{\pi}_{\theta}}(b) \nabla_{\theta} \int_{a \in \mathcal{A}} \pi_{\theta}(a|s) \d a \rspsqb{},  \\
    &= \mathbb{E}_{d^{\pi_{\psi}}(s, b)} \lspsqb{}  V^{\hat{\pi}_{\theta}}(b) \nabla_{\theta} 1 \rspsqb{}  = 0,
\end{align}
Noting that this is an example of the \emph{baseline} trick used throughout RL~\cite{bertsekas2019reinforcement, sutton1992reinforcement, williams1992simple}.
\end{proof}
This allows us to construct a gradient estimator using the advantage, which in conventional RL, is observed to reduce the variance of the gradient estimator compared to directly using the Q values.  

We are now able to prove an exact form of the A2D update.  This proof is similar to Theorem \ref{supp:def:fixed_point_variational}, however, no longer assumes identifiability of the POMDP-MDP process pair by instead updating the expert at each iteration.  

\subsubsection{Theorem 3}

\setcounter{theorem}{2}  
\begin{theorem}[Convergence of Exact A2D, reproduced from Section \ref{sec:algorithm}]
\label{supp:def:exact_a2d}
Under exact intermediate updates to the expert policy (see \eqref{supp:equ:i_want_to_cry}), the following iteration converges to an optimal partially observed policy $\pi_{\psi^*}(a|b)\in\Pi_{\phi}$, provided Assumption \ref{supp:assump:suff_var} holds:
\begin{align}
    \psi_{k+1}  &= \mathop{\argmin}_{\psi \in \Psi} \mathop{\mathbb{E}}_{d^{\pi_{\psi_k}}(s,b)} \lspsqb{}  \mathbb{KL} \lspsqb{}  \pi_{\theta^*_k}(a|s) \klbars{} \pi_\psi(a|b) \rspsqb{} \rspsqb{} , \label{supp:equ:why_wont_it_stop}\\
    \where  \hat{\theta}^*_k &= \mathop{\argmax}_{\theta \in \Theta} \mathop{\mathbb{E}}_{d^{{\pi_{\psi_k}}}(b) \hat{\pi}_{\theta}(a | b)} \lspsqb{} Q^{{\hat{\pi}_{\theta}}}(a,b) \rspsqb{} . \label{supp:equ:i_want_to_cry}
\end{align}
\end{theorem}
\setcounter{theorem}{\value{sthe}}
\begin{proof}
We will again, for ease of exposition assume that a unique optimal policy exists, as in Theorem \ref{supp:def:fixed_point_variational}. We again reinforce that this is not a \emph{requirement}.  Extending this proof to include multiple optimal partially observable policies only requires that we reason about the $\mathbb{KL}$ divergence between $\pi_{\psi_k}$ and $\pi_{\phi^*}$ at each step in the proof, instead of showing that the optimal parameters are equal.  This alteration is technically simple, but is algebraically and notationally onerous.   Similar to Theorem \ref{supp:def:fixed_point_variational}, we begin by examining the limiting behavior of \eqref{supp:equ:why_wont_it_stop} as $k \tends \infty$, and apply Theorem \ref{supp:theorem:1_ail_target} to replace the expert policy with the implicit policy:
\begin{align}
    \psi^* &= \lim_{k \rightarrow \infty} \mathop{\argmin}_{\psi \in \Psi}  \mathop{\mathbb{E}}_{d^{\pi_{\psi_k}}(s,b)}  \lspsqb{}  \mathbb{KL} \lspsqb{}  \pi_{\hat{\theta}^*_k}(a|s) \klbars{} \pi_\psi(a|b) \rspsqb{}  \rspsqb{}, \\
    &= \lim_{k \rightarrow \infty} \mathop{\argmin}_{\psi \in \Psi}  \mathop{\mathbb{E}}_{d^{\pi_{\psi_k}}(b)}  \lspsqb{}  \mathbb{KL} \lspsqb{}  \hat{\pi}_{\hat{\theta}^*_k}(a|b) \klbars{} \pi_\psi(a|b) \rspsqb{}  \rspsqb{}
\end{align}
We can then apply a direct extension of Lemma \ref{supp:lemma:2}, where the parameters of the expert policy are also updated in each iteration of the $\mathbb{KL}$ minimization, now denoted $\hat{\theta}^*(\psi)$.  The induction in Lemma \ref{supp:lemma:2} then proceeds as before.  Application of this extended Lemma \ref{supp:lemma:2} yields:
\begin{align}
    \psi^* &= \lim_{k \rightarrow \infty} \mathop{\argmin}_{\psi \in \Psi}  \mathop{\mathbb{E}}_{d^{\pi_{\psi_k}}(b)}  \lspsqb{}  \mathbb{KL} \lspsqb{}  \hat{\pi}_{\hat{\theta}^*_k}(a|b) \klbars{} \pi_\psi(a|b) \rspsqb{}  \rspsqb{} ,  \\
    &= \mathop{\argmin}_{\psi \in \Psi}  \mathop{\mathbb{E}}_{d^{\pi_{\psi}}(b)}  \lspsqb{}  \mathbb{KL} \lspsqb{}  \hat{\pi}_{\hat{\theta}^*(\psi)}(a|b) \klbars{} \pi_\psi(a|b) \rspsqb{}  \rspsqb{} ,\ \ \ \where  \hat{\theta}^*(\psi) = \mathop{\argmax}_{\theta \in \Theta} \mathop{\mathbb{E}}_{d^{{\pi_{\psi}}}(b) \hat{\pi}_{\theta}(a | b)} \lspsqb{} Q^{{\hat{\pi}_{\theta}}}(a,b) \rspsqb{} \label{supp:equ:mahh}
\end{align}
We can then apply a similarly extended a version of Lemma \ref{supp:lemma:3}, by using the same logic to allow the parameters of the expert policy to be updated as a function of $\psi$ in the $\mathbb{KL}$ minimization.  Now $\hat{\theta}^*(\psi)$ is defined as the expectation under the optimal POMDP policy.  To clarify, this update is, of course, intractable; however, here we are deriving what the equivalent and tractable iterative scheme outlined in \eqref{supp:equ:why_wont_it_stop} converges to, and hence we never actually need to evaluate $\hat{\theta}^*(\psi)$ as it is defined in Equation \eqref{supp:equ:mahh}.  Application of this extended lemma yields:
\begin{align}
    \psi^* &= \mathop{\argmin}_{\psi \in \Psi}  \mathop{\mathbb{E}}_{d^{\pi_{\psi}}(b)}  \lspsqb{}  \mathbb{KL} \lspsqb{}  \hat{\pi}_{\hat{\theta}^*(\psi)}(a|b) \klbars{} \pi_\psi(a|b) \rspsqb{}  \rspsqb{} , 
    \\
    &= \mathop{\argmin}_{\psi \in \Psi}  \mathop{\mathbb{E}}_{d^{\pi_{\phi^*}}(b)}  \lspsqb{}  \mathbb{KL} \lspsqb{}  \hat{\pi}_{\hat{\theta}^*(\psi)}(a|b)\klbars{} \pi_\psi(a|b) \rspsqb{}  \rspsqb{} , \ \ \  \where  \hat{\theta}^*(\psi) = \mathop{\argmax}_{\theta \in \Theta} \mathop{\mathbb{E}}_{d^{{\pi_{\phi^*}}}(b) \hat{\pi}_{\theta}(a | b)} \lspsqb{} Q^{{\hat{\pi}_{\theta}}}(a,b) \rspsqb{}
\end{align}
Lastly, Assumption \ref{supp:assump:suff_var} states that $\pi_{\phi^*} \in \hat{\Pi}_{\hat{\theta}^*}$, and so we can replace $\hat{\pi}_{\hat{\theta}^*}$ with the optimal partially observing policy ${\pi_{\phi^*}}$.  As a result, we have shown that we are implicitly solving a symmetric imitation learning problem, imitating the optimal partially observing policy:
\begin{align}
    \psi^* &= \mathop{\argmin}_{\psi \in \Psi}  \mathop{\mathbb{E}}_{d^{\pi_{\phi^*}}(b)}  \lspsqb{}  \mathbb{KL} \lspsqb{}  \hat{\pi}_{\hat{\theta}^*(\psi)}(a|b) \klbars{} \pi_{\psi}(a|b) \rspsqb{}  \rspsqb{} , 
    \\
    &= \mathop{\argmin}_{\psi \in \Psi}  \mathop{\mathbb{E}}_{d^{\phi^*}(b)}  \lspsqb{}  \mathbb{KL} \lspsqb{}  \pi_{\phi^*}(a|b)\klbars{} \pi_\psi(a|b) \rspsqb{} \rspsqb{} ,
\end{align}
where this optima can be achieved by our variational policy, yielding the initially stated result:
\begin{align}
    \psi^* &= \lim_{k \rightarrow \infty} \mathop{\argmin}_{\psi \in \Psi}  \mathop{\mathbb{E}}_{d^{\pi_{\psi_k}}(s,b)}  \lspsqb{}  \mathbb{KL} \lspsqb{}  \pi_{\hat{\theta}_k^*}(a|s) \klbars{} \pi_\psi(a|b) \rspsqb{}  \rspsqb{}
    = \mathop{\argmin}_{\psi \in \Psi} \mathop{\mathbb{E}}_{d^{\pi_{\phi^*}}(b)}  \lspsqb{}  \mathbb{KL} \lspsqb{}  \pi_{\phi^*}(a|b) \klbars{} \pi_{\psi}(a|b)  \rspsqb{}  \rspsqb{} \label{supp:equ:last_one}
\end{align}
which can be exactly minimized, as per Assumption \ref{supp:assump:suff_var}.  Directly performing the imitation in the right hand side of Equation \eqref{supp:equ:last_one}, although practically intractable, is guaranteed to recover a performant trainee.  We have therefore shown that the iterative procedure outlined in Equations \eqref{supp:equ:why_wont_it_stop} and \eqref{supp:equ:i_want_to_cry} recovers a trainee that is equivalent to an optimal partially observing policy as desired.  
\end{proof}

We conclude by noting that if we assume that $d^{\pi_{\psi}} > 0$ for all $\pi_{\psi} \in \Pi_{\psi}$, then each of the steps given in Theorems \ref{supp:def:fixed_point_variational} and \ref{supp:def:exact_a2d} can be shown trivially. If we assume at each iteration we successfully minimize the $\mathbb{KL}$ divergence, we obtain a variational policy which perfectly matches the updated expert everywhere.  In Theorem \ref{supp:def:fixed_point_variational} this directly implies the result, and by definition the algorithm must have converged after just a single iteration. In Theorem \ref{supp:def:exact_a2d}, we need only note that the $\argmax$ that produces the updated expert policy parameters must itself by definition match the optimal partially observed policy everywhere, and thus Theorem \ref{supp:def:exact_a2d} collapses to the same logic from Theorem \ref{supp:def:fixed_point_variational}.

\subsubsection{Discussion}
In this section we presented a derivation of exact A2D, where the expert is defined through the exact internal maximization step defined in \eqref{supp:equ:i_want_to_cry}.  We include these derivations to show the fundamental limitations of imitation learning and thus A2D under ideal settings.  Exactly performing this maximization is difficult unto itself, and therefore the A2D algorithm presented in Algorithm \ref{alg:a2d} simply assumes that this maximization is performed sufficiently accurately to produce meaningful progress in policy space.  Although we note that empirically A2D is robust to inexact updates, we defer the challenging task of formally and precisely quantifying the convergence properties of A2D under inexact internal updates to future work.

\clearpage

\section{Experimental Configurations}
\subsection{Gridworld}

We implemented both gridworld environments by adapting the \emph{MiniGrid} environment provided by \citet{gym_minigrid}.  For both gridworld experiments, the image is rendered as a $42 \times 42$ RGB image.  The agent has four actions available, moving in each of the compass directions.  Each movement incurs a reward of $-2$, hitting the weak patch of ice or tiger incurs a reward of $-100$, and reaching the goal incurs a reward of $20$.  Pushing the button in Tiger Door is free, but effectively costs $4$ due to its position, or $2$ in the Q function experiments.  Policy performance is evaluated every five steps by sampling $2000$ interactions under the stochastic policy.  A discount factor of $\gamma=0.995$ was used and an upper limit of $T=200$ is placed on the time horizon.

For experts and agents/trainees that use the compact representation, the policy is a two layer MLP that accepts a compact vector as input, with $64$ hidden units in each layer, outputting the log-probabilities of each action.  The value function uses the same architecture and input, but outputs a number representing the reward ahead.   The value function is learned by minimizing the mean squared error in the discounted reward-to-go.  $32$ batches are constructed from the rollout and are used to update the value function using ADAM~\citep{kingma2014adam} with a learning rate of $7 \times 10^{-4}$, for $25$ epochs.  Q functions are only used here with compact representations, and so we can simply append a one-hot encoding of the action to the (flat) input vector.  The Q function is then learned in the same way as the value function, except for with a slightly lower learning rate of $3\times 10^{-4}$.  Policies and value functions conditioned on images use a two layer convolutional encoder, each with $32$ filters, and a single output layer mapping to a flat hidden state with $50$ hidden units.  Image-based policies and value functions learn separate image encoders in the gridworld examples, whereas in the CARLA examples, a shared encoder is used.  This output is then used as input into a two layer MLP, each with $64$ hidden units, outputting the log-probabilities of each action or the expected discounted reward ahead.   L2 regularization is applied to all networks, with a coefficient of $0.001$.  

We use TRPO with batch sizes of $2,000$ and a trust region of $0.01$.  An entropy regularizer is applied directly to the advantages computed, with coefficient $1$.  We set $\lambda=0.95$ in the GAE calculation~\citep{schulman2015high}.  This trust region, regularization and $\lambda$ value are used throughout, unless otherwise stated.  Reinforcement learning in the POMDP (\emph{RL}) uses separate policies and value functions conditioned on the most recent image.  In asymmetric reinforcement learning (\emph{RL (Asym)}) the policy is conditioned on the image, but the value function takes the compact and omniscient state representation ($s_t$) as input.  Policies and value functions are then learned using the same process as before.

The policy learned by \emph{RL (MDP)} is then used as the expert in AIL (\emph{AIL}), where $2,000$ samples are collected at each iteration and appended to a rolling buffer of $5,000$ samples.  The KL-divergence between the expert and trainee action distributions is minimized by performing stochastic gradient descent, using ADAM with a learning rate of $3\times 10^{-4}$, using a batch size of $64$ for two whole epochs per iteration.  We find that the MDP converges within approximately $80,000$ environment interactions, and so we begin the AIL line at this value.  $\beta$ is annealed to zero after the first time step (as recommended by \citet{Ross2011}).  

For experiments using a pretrained encoder (\emph{Pre-Enc}), we roll out for $10,000$ environment interactions under a trained MDP expert from \emph{RL (MDP)} to generate the data buffer.  The encoder, that takes images as input and targets the true state vector, is learned by regressing the predictions on to the true state.  We perform $100$ training epochs with a learning rate of $3 \times 10^{-4}$.  We start this curve at the $80,000$ interactions required to train the expert from which the encoder is learned.  We use an asymmetric value function conditioned on the true state.  The encoder is then frozen and a two-layer, $64$ hidden unit MLP policy head is learned using TRPO.  We found that a lower trust region size of $0.005$ was required for Tiger Door to stably converge.  We confirmed separately for both pretrained encoders and AIL that the encoder class can represent and learn the required policies and transforms, and both converge to the solution of the MDP when conditioned on \emph{omniscient} image-based input.  

For A2D, expert and trainee policies are initialized from scratch, and are learned using the broadly the same settings as \emph{RL (MDP)} and \emph{AIL}.  In A2D, we decay $\beta$ with coefficient $0.8$ at each iteration, although faster $\beta$ decays did not hurt performance.  Slower $\beta$ decays can lead to higher and longer divergences during training, and can lead to the agent becoming trapped in local optima.  We use a higher entropy regularization coefficient, equal to $10$, finding that this increased regularization helped A2D avoid falling into local minima, although this can be further ameliorated by setting $\beta = 0$ throughout, as we do in the CARLA experiments.  We found for Frozen Lake that a lower $\lambda = 0.9$ value of yielded more stable convergence and a lower final policy divergence (we refer the reader to Section \ref{supp:sub:sub:q} for more information).  Value and Q functions are learned by individually targeting the sum of rewards ahead (i.e. is not back-propagated through any mixture).  We note that choosing to parameterize the mixture value function as the weighted sum of individual value functions is an assumption.  However, we note that we require $\beta \tends 0$ for the gradient to be unbiased.  In this limit the mixture is equal to just the value function of the agent.  Therefore, explicitly parameterizing the value function in this way \emph{ensures} that state information is removed from the estimation.  Exploring different ways of parameterizing the value function is a potential topic for future research.

In Section \ref{supp:sub:sub:q} we use a $\lambda$ value of $0.5$ in GAE~\citep{schulman2015high} (when not sweeping over $\lambda$).  We used an entropy regularizer of $0.02$ is applied directly to the surrogate loss.  We also use TRPO with a trust region KL-divergence of $0.001$.

\subsection{CARLA Experiments}
We implemented our autonomous vehicle experiment using CARLA~\citep{Dosovitskiy17}. This scenario represents a car driving forward at the speed limit, while avoiding a pedestrian which may run out from behind a vehicle $50\%$ of the time, at a variable speed.  There are a total of $10$ waypoints, indicating the path the vehicle should take as produced by an external path planner.  We enforce that the scenario will end prematurely if one of the following occurs: a time limit of $90$ time-steps is reached, a collision with a static object, a lane invasion occurs, if a waypoint is not gathered within $35$ time-steps, or, the car's path is not within a certain distance of the nearest waypoint.  We found that inclusion of these premature endings was crucial for efficient learning.  The reward surface for this problem is generated using a PID controller which is computed using an example nominal trajectory.  The reward at any given time-step is defined as the product of the absolute difference between the agents actions and the optimal actions by a low-level PID controller to guide the vehicle to the next waypoint, and is bounded to lie in $[0,1]$.  

For the expert policy used both in AIL and A2D, we use a two layer MLP with $64$ hidden units and ReLU activations. The agent and trainee policies use a shared image encoder~\citep{laskin2020reinforcement,laskin_srinivas2020curl,yarats2021image}, followed by the same MLP architecture as the expert policy to generate actions.  The RL algorithm used in both the expert and agent RL updates is PPO~\citep{schulman2017proximal} with generalized advantage estimation (GAE)~\citep{schulman2015high}.  We detach the encoder during the policy update and learn the encoder during the value function update~\citep{laskin2020reinforcement,laskin_srinivas2020curl,yarats2021image}. In A2D we use the MLP defined above for the expert policy.  The trainee policy and value functions use a common encoder, updated during the trainees value update and frozen during the policy update, and the MLP defined above as the policy head or value head network.  For all algorithms we used a batch size of $64$ in both the PPO policy update, value function update, and the imitation learning update.  As in the previous experiments, in the imitation learning step, we iterate through all data seen and stored in the replay buffer. We found that starting the $\beta$ parameter at zero produced faster convergence.

We performed a coarse-grained hyperparameter search using the Bayesian optimization routine provided by the experimental control and logging software \emph{Weights \& Biases}~\citep{wandb}.  This allows us to automate hyperparameter search and distribute experimental results for more complex experiments in a reproducible manner.  Each method was provided approximately the same amount of search time, evaluating at least $60$ different hyperparameter settings.  The optimal settings were then improved manually over the course of approximately $5$ further tests.  We score each method and hyperparameter setting using a running average of the reward over the previous $25$ evaluation steps, and used early stopping if a run consistently performed poorly.  

Each algorithm uses different learning rates and combinations of environment steps between updates. For example, we found that all AIL algorithms performed best when taking $10$ steps between updates, RL in the expert tended to work better by taking more steps in between updates ($\approx 400$) with a larger step-size $\approx 4\times 10^{-4}$, where the agents RL updates favored fewer steps $\approx 75$ with smaller steps $7 \times 10^{-5}$. For all algorithms $4$ parallel environments were run concurrently, as this was observed to improve performance across all algorithms. This was especially the case for the RL methods, which relied on more samples to accurately compute the advantage.  

We note that there is a point of diminishing returns for PPO specifically~\cite{Engstrom2020Implementation}, where policy learning degrades as the number of examples per update increases. Even though the advantage becomes progressively more accurate with increasing sample size, the mini-batch gradient decent procedure in PPO eventually leads to off-policy behavior that can be detrimental to learning.  We also found that pre-generating a number of trajectories and pretraining the value function tended to improve performance for both A2D, as well as the compact expert RL algorithm.  For A2D specifically, this ensured that the replay buffer for imitation learning was reasonably large prior to learning in the expert. This ensures that for any given update, the agent tends to be close to the expert policy, ensuring that the "off-policy" RL update is not too severely destabilized through importance weighting. To further improve this, we also introduced delayed policy updates, which further reduced the divergence between expert and the agent in A2D. In both A2D and the RL setups, this also helped ensure that the value function is always converging faster than the policy, ensuring that the error in the resulting advantage estimates are low.

\clearpage

\section{Additional Related Work}
We now present a comprehensive review of existing literature not already covered.  Exploiting asymmetric learning to accelerate learning has been explored in numerous previous work under a number of different frameworks, application domains, and levels of theoretical analysis.

The notion of using fully observed states unavailable at deployment time is often referred to as exploiting ``privileged information''~\citep{vapnik2009new, lambert2018deep}.  For clarity, we refer to the expert as having access to privileged information, and the agent as only having access to a partial observation.  We note that the use of the term ``expert'' does not imply that this policy is necessarily optimal under the MDP.  Indeed, in A2D, the expert is co-trained with the agent, such that the expert is approximately a uniform random distribution at the start of the learning procedure.  The term privileged information is more general than simply providing the world state, and may include additional loss terms or non-trivial transforms of the world state that expedite learning the agent.  In this work, we exclusively consider the most general scenario where the privileged information is the full world state.  However, there is nothing precluding defining an extended state space to include hand-designed features extracted from the state, or, using additional, hand crafted reward shaping terms when learning (or adapting) the expert. 

\subsection{Encodings}
The first use-case we examine is probably the simplest, and the most widely studied.  Asymmetric information is used to learn an encoding of the observation that reduces the dimensionality while retaining information.  Standard reinforcement learning approaches are then employed freezing this encoding.  Two slight variations on this theme exist.  In the first approach, an MDP policy is learned to generate rollouts conditioned on omniscient information, and an encoder is learned on state-observation pairs visited during these rollouts~\citep{Finn2016, Levine2016}.  Either the encoder acts to directly recover the underlying states, or simply learns a lower-dimensional embedding where performing reinforcement learning is more straightforward.

\citet{Andrychowicz2020} explore learning to manipulate objects using a mechanical hand using \emph{both} state information from the robot (joint poses, fingertip positions etc) and RGB images.  This particular application is an interesting hybrid approach dictated by the domain.  State information pertaining to the manipulator is easily obtained, but state information about the pose of the object being manipulated is unavailable and must be recovered using the images.  A controller is learned in simulation (MDP), while simultaneously (and separately from the MDP) a separate perception network is learned that maps the image to the pose of the object being manipulated.  State information and pose encoding are then concatenated and used as the state vector on which the policy acts.  While the pose of the object is unobserved, it is readily recoverable from a single frame (or stack of frames), and hence the partial observation is predominantly a high-dimensional and bijective embedding of the true state.  If the true position of the hand was not available, this would be less certain as the object and other parts of the manipulator obfuscates much of the manipulator from any of the three viewpoints (more viewpoints would of course reverse this to being a bijection).  The use of a recurrent policy further improves the recovery of state as only the innovation in state needs to be recovered. 

\subsection{Asymmetric values}
Another well-explored use-case is to instead exploit asymmetric information for to improve learning a value or Q- function~\citep{kononen2004asymmetric, pinto2017asymmetric, Andrychowicz2020}.  This is achieved by conditioning either the value function or Q-function on different information than the policy that is either more informative, or lower dimensional representations, and can help guide learning~\cite{kononen2004asymmetric, pinto2017asymmetric}.  Learning the value or Q function in a lower-dimensional setting enables this function to be learned more stably and with fewer samples, and hence can track the current policy more effectively.  Since the value and Q-function are not used at test time, there is no requirement for privileged information to be available when deployed.  \citet{pinto2017asymmetric} introduce this in a robotics context, using an asymmetric value function, conditioned on the true underlying state of a robotic manipulator, to learn a partially observing agent conditioned only on a third-person monocular view of the arm.  Similar ideas were explored previously by \citet{kononen2004asymmetric} in relation to semi-centralized multi-agent systems, where each agent only partially observes the world state, but a central controller is able to observe the whole state.  The state used by the central controller is used to evaluate the value of a particular world state, whilest each agent only acts on partial information.

\subsection{Behavioral Cloning \& Imitation Learning}
Behavioral cloning and imitation learning~\citep{pmlr-v80-kang18a, Ross2011}, introduced in Main Section 2.3, is, in our opinion, an under-explored avenue for expediting learning in noisy and high-dimensional partially observed processes.  The main observation is that this process separates learning to act and learning to perceive~\citep{Chen2019}.  The fully observing expert learns to act, without the presence of extraneous patterns or noise.  The agent then learns to perceive such that it can replicate the actions of the expert.  A major benefit of cloning approaches is that perception is reduced to a supervised learning task, with lower variance than the underlying RL task.  

\citet{pinto2017asymmetric} briefly assess using asymmetric DAgger as a baseline.  It is observed that the agent learns quickly, but actually converges to a worse solution than the asymmetric actor-critic solution.  This difference is attributed to the experts access to (zero variance) state information otherwise unavailable to the partially observing agent.  Our work builds on this observation, seeking to mitigate such weaknesses.  Surprisingly, and to the best of our knowledge, no work (including \citet{pinto2017asymmetric}) has provided and in-depth analysis of this method, or directly built off this idea.  

\citet{Chen2019} showed that large performance gains can be found in an autonomous vehicles scenario by using IL through the use of an asymmetric expert, specifically for learning to drive in the autonomous vehicle simulator CARLA~\citep{Dosovitskiy17}.  \citet{Chen2019} train an expert from trajectories, created by human drivers, using behavioral cloning conditioned on an encoded aerial rendering of the environment including privileged information unavailable to the agent at deployment time.  The aerial rendering facilitates extensive data augmentation schemes that would otherwise be difficult, or impossible, to implement in a symmetric setting.  The agent is then learned using DAgger-based imitation learning.  However, this general approach implicitly makes assumptions about the performance of the expert, as well as the underlying identifiability (as we define in Section \ref{sec:il-failure}) between the underlying fully and partially observed Markov decision processes.

Other works combine RL and IL to gain performance beyond that of the expert by considering that the expert is sub-optimal~\citep{choudhury2018data, Sun2018, weihs2020bridging}, where the performance differential is either as a result of asymmetry, or, the expert simply not being optimal.  These works, most often, train a policy that exploits knowledge of the performance differential between the expert and agent, or, the difference in policies.  The weight applied to the sample in IL is increased for policies that are similar, or, where the performance gap is small.  The example is then down-weighted when it is believed that the expert provides poor supervision in that state.  However, these works do not consider updating the expert, and instead focus on ameliorating the drawbacks of AIL using derived statistics.  In our work, we seek to define a method for updating an expert directly.

\subsection{Co-learning Expert and Agent}
Work that is maybe thematically most similar to ours investigates co-training of the agent and expert.  This builds on the AIL approach, where instead of assuming an optimal expert exists, the expert and agent policies are learned simultaneously, where either an additional training phase as added to ``align'' the expert and agent~\citep{salter2019attentionprivileged, Song2019}, architectural modification~\citep{pierrealex2020privileged}, or both~\citep{Schwab2019}.  An alternative method for deriving such an aligning gradient is to introduce an auxiliary loss regularizing the representation used by the agent to be predictive of the the underlying state, or, a best-possible belief representation~\citep{nguyen2020belief}.

\citet{salter2019attentionprivileged} trains separate policies for agent and expert using spatial attention, where the expert is conditioned on the state of the system, and the agent is conditioned on a monocular viewpoint.  By inspecting the attention map of expert and agent, it is simple to establish what parts of the state or image the policy is using to act.  An auxiliary (negative) reward term is added to the reward function that penalizes differences in the attention maps, such that the agent and expert are regularized to use the same underlying features.  This auxiliary loss term transfers information from the MDP to the POMDP.  The main drawbacks of this approach however are its inherent reliance on an attention mechanism, and tuning the hyperparameters dictating the weight of having a performant agent, expert and the level of alignment between the attention mechanisms.  Further, using a attention as the transfer mechanism between the agent and expert somewhat introduces an additional layer of complexity and obfuscation of the actual underlying mechanism of information transfer.

\citet{Song2019} present an algorithm, CoPiEr, that co-trains two policies, conditioned on different information (any combination of fully or partially observing).  CoPiEr rolls out under both policies separately, and then selects the rollouts from the policy that performs the best.  These samples are then used in either an RL or IL (or hybrid of the two) style update.  In this sense, the better performing policy (with ostensibly ``better'' rollouts) provides high-quality supervision to the policy with lower quality rollouts.  MDP to POMDP transfer or privileged information is not considered.  Most significantly, imitation learning is proposed as a method of transferring from one policy to another, or, RL augmented with an IL loss to provide better supervision while retaining RLs capability to explore policy space. 

\citet{Schwab2019} on the other hand extend \citet{pinto2017asymmetric} by introducing multitask reinforcement learning themes.  A ``task'' is uniquely described by the set of variables that the policy is conditioned on, such as images from different view points, true state information and proprioceptive information.  An input-specific encoder encodes each observation before mixing the encoded input features and passing these to a head network which outputs the actions.  Instead of aligning attention mechanisms (as per \citet{salter2019attentionprivileged}), \citet{Schwab2019} the head network is shared between tasks providing alignment between the single-input policies.  At test time, only those observations that are available need to be supplied to the policy, respecting the partial observability requirement at test time.  This approach does not explicitly use an expert, instead using a greater range of more informative information channels to efficiently learn the policy head, while simultaneously co-training the channel-specific encoders.

Finally, the work of \citet{pierrealex2020privileged} present privileged information dropout (PI-D).  The general approach of information dropout~\citep{Achille_2018} is to lean a model while randomly perturbing the internal state of the model, effectively destroying some information.  The hypothesis is that this forces the model to learn more robust and redundant features that can survive this corruption.  \citet{pierrealex2020privileged} use this theme by embedding both partial observation and state, where the state embedding is then used to corrupt (through multiplicative dropout) the internal state of the agent.  The PI expert is then able to mask uninformative patterns in the observations (using the auxiliary state information), facilitating more efficient learning.  The PI can then be easily marginalized out by not applying the dropout term.  Importantly however, reinforcement learning is still performed in the partially observing agent, a characteristic we wish to avoid due to the high-variance nature of this learning.

% \bibliography{main}

\end{document}